%% file: sempipes.tex
\documentclass{article}

\usepackage{microtype}
\usepackage{graphicx}
\usepackage{subfigure}
\usepackage{booktabs} 
\usepackage{multirow} 
\PassOptionsToPackage{hyphens}{url}
\usepackage{hyperref}

\usepackage{etoc}

\makeatletter
\let\icml@addcontentsline\addcontentsline
\makeatother

\usepackage[accepted]{icml2025}

\usepackage{amsmath}
\usepackage{amssymb}
\usepackage{mathtools}
\usepackage{amsthm}
\usepackage[capitalize,noabbrev,nameinlink]{cleveref}

\theoremstyle{plain}

\theoremstyle{definition}

\theoremstyle{remark}

\usepackage[textsize=tiny]{todonotes}

\usepackage{balance}

\input{meta}

\icmltitlerunning{\textsc{SemPipes} -- Optimizable Semantic Data Operators for~Tabular~Machine~Learning~Pipelines}

\begin{document}

\twocolumn[
\icmltitle{\textsc{SemPipes} -- Optimizable Semantic Data Operators for~Tabular~Machine~Learning~Pipelines}




\begin{icmlauthorlist}
\icmlauthor{Olga Ovcharenko}{bt}
\icmlauthor{Matthias Boehm}{bt}
\icmlauthor{Sebastian Schelter}{bt}
\end{icmlauthorlist}

\icmlaffiliation{bt}{BIFOLD \& TU Berlin}

\icmlcorrespondingauthor{Olga Ovcharenko}{ovcharenko@tu-berlin.de}
\icmlcorrespondingauthor{Sebastian Schelter}{schelter@tu-berlin.de}

\icmlkeywords{TBD}

\vskip 0.3in
]



\printAffiliationsAndNotice{} 

\input{sections/00-abstract}

\input{sections/01-introduction}
\input{sections/02-related}
\input{sections/03-operators}
\input{sections/04-optimization}

\input{sections/05-evaluation}

\input{sections/06-conclusion}

\input{sections/impact-statement}


\bibliography{sempipes}
\bibliographystyle{icml2025}

\newpage
\appendix
\onecolumn

\input{sections/07-appendix}

\end{document}

%% file: meta.tex
\usepackage[utf8]{inputenc}
\usepackage[T1]{fontenc}
\usepackage[LGR, T1]{fontenc}
\usepackage[greek, english]{babel}
\usepackage{CJKutf8}
\usepackage{cuted}
\usepackage[normalem]{ulem}

\definecolor{lightgray}{RGB}{246,248,250}
\usepackage{inconsolata}

\usepackage{enumitem}
\usepackage{dirtytalk}
\usepackage[compatibility=false]{caption}

\usepackage{algorithm}
\usepackage{algorithmicx}
\usepackage[indLines=false]{algpseudocodex}
\makeatletter
\renewcommand{\algorithmicdo}{} 
\renewcommand{\algorithmicthen}{} 

\makeatother

\usepackage{amsmath}
\DeclareMathOperator*{\argmax}{argmax}
\usepackage{bm}

\def\ojoin{\setbox0=\hbox{$\bowtie$}%
  \rule[-.02ex]{.25em}{.4pt}\llap{\rule[\ht0]{.25em}{.4pt}}}
\def\leftjoin{\mathbin{\ojoin\mkern-5.8mu\bowtie}}

\usepackage{multirow}
\usepackage{graphicx} 

\usepackage{listings}
\usepackage{booktabs}

\usepackage{amssymb}
\usepackage{pifont}
\newcommand{\cmark}{\ding{51}}%

\usepackage{tikz}
\newcommand*\circled[1]{\protect\tikz[baseline=(char.base)]{\protect\node[shape=circle,fill=black,draw,inner sep=0.8pt] (char) {\textcolor{white}{\tiny {#1}}};}}

\usepackage{array}
\usepackage{ragged2e}
\usepackage[dvipsnames]{xcolor}

\hypersetup{
    colorlinks=true,
    linkcolor=blue,
    citecolor=blue,
    filecolor=blue,
    urlcolor=blue,
}

\definecolor{codegray}{rgb}{0.5,0.5,0.5}
\definecolor{backcolour}{rgb}{0.95,0.95,0.92}
\definecolor{strdspy}{HTML}{3A5F9B}
\definecolor{kwdspy}{HTML}{CC4444}
\definecolor{dspy}{HTML}{9B6EF3}
\definecolor{commentdspy}{HTML}{00AA00}

\lstdefinestyle{mypython}{
    language=Python,
    backgroundcolor=\color{backcolour},
    commentstyle=\color{commentdspy}\ttfamily,
    keywordstyle=\color{black}\bfseries,
    numberstyle=\tiny\color{codegray},
    stringstyle=\color{strdspy},
    basicstyle=\ttfamily\footnotesize,
    breaklines=true,
    captionpos=b,
    keepspaces=true,
    numbers=left,
    numbersep=10pt,
    showspaces=false,
    showstringspaces=false,
    showtabs=false,
    tabsize=2,
    xleftmargin=1em,
    xrightmargin=1em,
    framexleftmargin=1em,
    framexrightmargin=1em,
    deletekeywords={apply},
    postbreak=\mbox{\hspace{.2em}}
}

\newcommand{\header}[1]{\vspace{1mm}\noindent\textbf{#1}.}
\newcommand{\headerl}[1]{\vspace{1mm}\noindent\textit{#1}.}
\newcommand{\headerlu}[1]{\vspace{1mm}\noindent\textit{\uline{#1}}.}

\newcommand{\commithash}[0]{v1}
\newcommand{\repo}[1]{\url{https://github.com/deem-data/sempipes#1}}

\newcommand{\repofull}[1]{\url{https://github.com/deem-data/sempipes/tree/\commithash#1}}

\newcommand{\pmS}[1]{
    {\fontsize{7}{7}\selectfont $\pm$ #1}
    \vspace{0.007cm}
}

\newcommand{\money}[1]{
    {\fontsize{7}{7}\selectfont\textit{\$#1}}
}


%% file: sections/00-abstract.tex
\begin{abstract}
Real-world machine learning on tabular data relies on complex data preparation pipelines for prediction, data integration, augmentation, and debugging. Designing these pipelines requires substantial domain expertise and engineering effort, motivating the question of how large language models (LLMs) can support tabular ML through code synthesis.
We introduce \textsc{SemPipes}, a novel declarative programming model that integrates LLM-powered semantic data operators into tabular ML pipelines. Semantic operators specify data transformations in natural language while delegating execution to a runtime system. During training, \textsc{SemPipes} synthesizes custom operator implementations based on data characteristics, operator instructions, and pipeline context. This design enables the automatic optimization of data operations in a pipeline via LLM-based code synthesis guided by evolutionary search.
We evaluate \textsc{SemPipes} across diverse tabular ML tasks and show that semantic operators substantially improve end-to-end predictive performance for both expert-designed and agent-generated pipelines, while reducing pipeline complexity. We implement \textsc{SemPipes} in Python and release it at \url{https://github.com/deem-data/sempipes/tree/v1}.
\end{abstract}

%% file: sections/01-introduction.tex

\section{Introduction}

Large language models (LLMs) are rapidly reshaping data-centric software development, driven by their growing capabilities in code synthesis and program understanding. Recent approaches span from AI-assisted programming environments~\cite{github-copilot2025,CursorAI2025} and interactive code agents~\cite{yang2024swe,anthropic2025claudecode} to autonomous systems that synthesize new programs~\cite{chan2024mle,toledo2025ai,aygun2025ai,nam2025dsstardatascienceagent,fang2025mlzero,nam2025mlestar}. Despite their promise, AI-assisted methods raise concerns around safety, control, and unintended behavior~\cite{nolan2025replit,reddit:senior_ml_api_calls}, and may in some cases even reduce developer productivity~\cite{becker2025measuringimpactearly2025ai}. A key reason is that chat-based interaction only offers coarse, ambiguous control over program behavior~\cite{karpathy2025software}, and that the generated code is difficult to optimize reliably~\cite{khattab2023dspy,lee2025compound, shi2024efficient}.

\header{The Gap for Tabular ML Pipelines} Tabular data remains a central focus of machine learning research and practice~\cite{van2024tabular,hollmann2025accurate,qu2025tabicl,feykumorfm,tschalzev2024a}. In real-world settings, tabular data typically originates from heterogeneous sources such as data lakes, data warehouses, log files, or REST APIs, and must be joined, cleaned, augmented~\cite{hollmann2023large}, and encoded through \emph{tabular ML pipelines}~\cite{polyzotis2017data}. These pipelines support diverse tasks, including prediction, data integration, debugging, and postprocessing~\cite{polyzotis2019data,karlavs2022data, cappuzzo2025retrieve}, combine complex code across many libraries~\cite{psallidas2022data}, and are costly to execute in production~\cite{xin2021production}. As a result, developing data pipelines for tabular ML is tedious and error-prone, requiring substantial domain expertise and engineering effort. Industry surveys consistently identify data preparation as the most time-consuming and least enjoyable aspect of data science practice~\cite{anaconda2022stateofdatascience}. 
Recent work has explored fully automated, end-to-end pipeline generation for prediction tasks~\cite{toledo2025ai,aygun2025ai,nam2025dsstardatascienceagent,fang2025mlzero,yang2025r,nam2025mlestar, küken2025largelanguagemodelsengineer}, reporting promising results on constrained, Kaggle-derived benchmarks~\cite{chan2024mle}. However, these approaches leave several issues unresolved, see~\Cref{app:agents-shortcomings} for details on shortcomings. Common problems include a high fraction of initially erroneous non-runnable code~\cite{trirat2025automlagent}, susceptibility to data leakage, and unclear applicability beyond tabular prediction, e.g., for data integration. Several recent systems neither release generated code nor agent trajectories, and their designs are tightly coupled to specific benchmarks, limiting broader applicability.

\header{Declarative Programming} Declarative programming is emerging as a promising paradigm for integrating LLM-based operations into software systems in a controlled, modular manner. Instead of specifying \emph{how to compute} results, declarative programs  describe \emph{what to compute} using natural language instructions, while a runtime system determines execution strategies and orchestrates LLM calls. This separation of intent from execution mitigates common issues such as brittle, ad-hoc code that is hard to reuse. Importantly, declarative approaches expose automated optimization opportunities to the runtime, e.g., to choose or synthesize specialized operator implementations. Declarative approaches have gained traction in language model programming and data management: (1)~DSPy~\cite{khattab2023dspy} enables the declarative specification of LLM modules that are compiled into executable programs and automatically optimized with respect to data and model choices~\cite{opsahl2024optimizing,agrawal2025gepareflectivepromptevolution}. (2)~In databases, semantic query processing extends SQL with AI-powered operators, enabling queries over multimodal data, with cost-quality trade-offs via approximation techniques~\cite{patel2024semantic,jo2024thalamusdb,liu2025palimpzest,dorbani2025flockmtl,googlecloud_bigquery_generativeAI_overview_2025,databricks_ai_functions_2025,snowflake_cortex_aisql_2025}.

\begin{figure*}[t!]
    \centering
    \includegraphics[width=\linewidth]
    {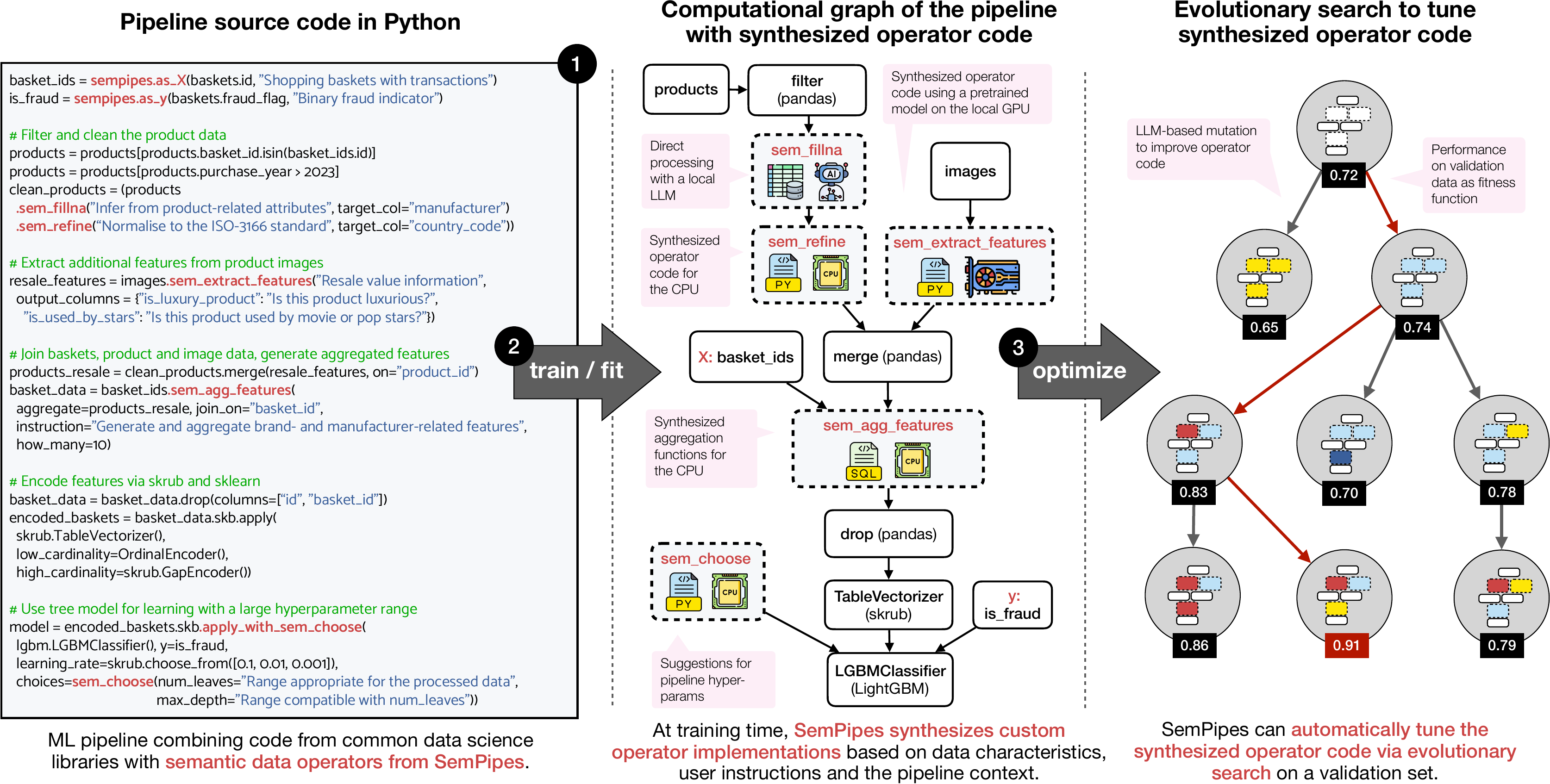}
    \vspace{-0.6cm}
    \caption{\textbf{High-level overview of \textsc{SemPipes}}:
\circled{1}~ML pipelines are defined in Python using standard data science libraries, including pandas for tabular data manipulation, numpy for numerical computation, scikit-learn for feature encoding, and models compatible with the scikit-learn ecosystem. \textsc{SemPipes} extends these pipelines with {\em semantic data operators}, which {\em delegate selected data-centric operations to LLMs}.
\circled{2}~During training, when a pipeline is \say{fitted} to the training data, \textsc{SemPipes} {\em synthesizes custom implementations for the semantic operators} based on data characteristics, natural language instructions, and pipeline context. The synthesized code may execute locally on the CPU or leverage a pretrained model on a local GPU.
\circled{3}~Once a pipeline is defined, {\em the implementations of its semantic operators can be tuned on a validation set}. \textsc{SemPipes} performs evolutionary search under a given search policy (e.g., Monte Carlo tree search), using downstream validation performance as the fitness function. During search, operator implementations are iteratively mutated via reflective prompting using prior performance scores and a tree-structured memory.}
    \label{fig:overview}
\vspace{-0.2cm}
\end{figure*}

\header{\textsc{SemPipes}: Extending Tabular ML Pipelines with Semantic Data Operators} Declarative programming offers a principled approach for integrating LLMs into tabular ML pipelines by separating \emph{what} should be computed from \emph{how} the computation is executed. Building on this paradigm, we introduce \textsc{SemPipes}, a novel programming model that extends tabular ML pipelines with declarative, LLM-powered \emph{semantic data operators}. \textsc{SemPipes} pipelines are specified in Python and selectively delegate data-centric operations to context-aware, optimizable semantic operators (see \Cref{fig:overview} for a high-level overview).
During training, \textsc{SemPipes} uses LLMs to synthesize custom implementations for semantic operators, conditioned on data characteristics, natural language instructions, and pipeline context (\Cref{sec:approach-semops}). As a consequence, LLM calls are incurred only during training, and scale with the number of semantic operators—independent of data size—resulting in predictable LLM usage and no reliance on LLMs during inference. The design of \textsc{SemPipes} enables semi-automated pipeline construction while preserving the ability to iteratively develop the pipeline in an interactive notebook environment, and avoids the brittleness of chat-based interfaces and manual copy-pasting of code. Once a pipeline is defined, its operator implementations can be automatically tuned via LLM-based code synthesis guided by evolutionary search to improve end-to-end predictive performance (\Cref{sec:approach-opt}). Within this scope, \textsc{SemPipes} is not an AutoML system and does not generate end-to-end pipelines; instead, it introduces  semantic data operators to improve data operations and can be integrated into user-written or agent-generated pipelines.

\header{Contributions} Our contributions are as follows:
\vspace{-0.35cm}

\begin{itemize}[leftmargin=*,itemsep=1pt]
  \item We introduce a \emph{novel programming model} that extends tabular ML pipelines with optimizable semantic data operators, leveraging LLMs to synthesize custom operator code tailored to the data and pipeline~(\Cref{sec:approach-semops}).
  \item We develop an \emph{optimizer for semantic data operators} that improves a pipeline's predictive performance by evolving synthesized operator code via LLM-based mutation guided by tree-structured evolutionary search~(\Cref{sec:approach-opt}).
  \item We empirically evaluate \textsc{SemPipes} on a new custom-designed benchmark of nineteen expert-written and agent-generated pipelines across five tabular ML tasks. We find that \emph{integrating optimized semantic data operators consistently improves the pipelines' predictive performance}, and often simplifies the code. Furthermore, we showcase that our \emph{synthesized operator code is competitive with custom methods} for feature engineering, missing value imputation, and zero-shot feature extraction~(\Cref{sec:experiments}).
  \item We implement \textsc{SemPipes} in Python as an extension of the skrub library and release it under an open license at \repofull{}.
\end{itemize}

%% file: sections/02-related.tex
\section{Related Work}
\label{sec:related_work}

Tabular ML pipelines typically prepare, integrate, and clean data using pandas dataframes~\cite{pandas2020}, and then compose so-called \say{estimators}~\cite{scikit-learn_compose_doc} from scikit-learn~\cite{pedregosa2011scikit} to encode features, train models, and perform hyperparameter search. Such pipelines are widely adopted in practice~\cite{psallidas2022data}, as of October 2025, the scikit-learn and pandas packages together exceed 600 million monthly downloads according to \texttt{pypistats.org}. Furthermore, scikit-learn’s estimator abstraction has been adopted by other popular ML libraries, including SparkML~\cite{meng2016mllib} and TensorFlow~Transform~\cite{baylor2017tfx}, and is the foundation for ongoing research on ML pipelines~\cite{karlavs2022data,petersohn2020towards,grafberger2023automating}. Despite its popularity, the current ecosystem for tabular ML pipelines has several major limitations. Dataframes and data operations are not first-class citizens in scikit-learn’s estimator abstraction, which is designed to apply a linear sequence of transformations to a single table with a fixed number of rows~\cite{skrub_data_ops_vs_alternatives}. As a result, common operations such as filters, joins, aggregations, and working with multiple input tables are not supported. Integrating external libraries, e.g., for automated feature engineering, remains difficult~\cite{schafer2025usable}. 

In the past, data operations have typically been excluded from end-to-end optimization approaches~\cite{yu2021windtunnel}, as they are non-differentiable. A promising approach to overcome this limitation is optimization by prompting~\cite{yang2023large}, where a code-generating LLM is iteratively refining a program using textual feedback and performance scores~\cite{hollmann2023large,opsahl2024optimizing,agrawal2025gepareflectivepromptevolution,lee2025compound,aygun2025ai}. ML pipelines naturally fit this setting, as validation performance provides a clear optimization objective.

We defer a broader discussion of related work on declarative programming with semantic operators and agentic pipeline generation to \cref{app:related-work}. We argue that \textsc{SemPipes} provides a complementary abstraction for agents: reusable primitives that potentially allow them to separate reasoning about high-level pipeline structure from the synthesis and optimization of individual data operations.

\begin{table*}[t!]
\centering
\caption{\textbf{Semantic data operators in \textsc{SemPipes}.} Here, $\mathbf{s}$ is a natural language instruction, and $D$, $D^\prime$ denote input/output dataframes. The right columns show applicability on dataframes (DF), multi-modal dataframes (MM), and estimators (ET).
}
\setlength{\tabcolsep}{5.4pt}   
\vspace{-2.5mm}
{\small
\begin{tabular}{llccc}
\toprule
\multirow{2}{*}{\textbf{Semantic Data Operator}} & \multirow{2}{*}{\textbf{Definition}} & \multicolumn{3}{c}{\textbf{Applicable on}}\\
& & DF & MM & ET\\
\midrule
$\texttt{sem\_gen\_features}(D, k, s) \rightarrow D^\prime$ & Generate $k$ feature columns in $D$  & \cmark &&\\  
$\texttt{sem\_agg\_features}(D, D_\text{agg}, c_\text{join}, k, s) \rightarrow D^\prime$ & Generate $k$ aggregated feature columns in $D_\text{agg}$, join $D$ on $c_\text{join}$ & \cmark &&\\ 
$\texttt{sem\_extract\_features}(D, C_\text{out}, s) \rightarrow D^\prime$ & Extract feature columns from feature descriptions $C_\text{out}$ and $D$& & \cmark &\\
$\texttt{sem\_augment}(D, k, s) \rightarrow D^\prime$ & Augment $D$ with $k$ additional rows& \cmark &&\\ 
$\texttt{sem\_fillna}(D, c, s) \rightarrow D^\prime$ & Impute missing values in column $c$ of $D$ & \cmark & \cmark &\\ 
$\texttt{sem\_clean}(D, c, s) \rightarrow D^\prime$ & Clean column $c$ of $D$ & \cmark &&\\
$\texttt{sem\_refine}(D, c, s) \rightarrow D^\prime$ & Restructure column $c$ of $D$  & \cmark &&\\ 
$\texttt{sem\_select}(D, s) \rightarrow D^\prime$ & Reduce $D$ to a selected subset of columns & \cmark & \cmark & \cmark \\ 
$\texttt{sem\_choose}(E, s) \rightarrow H$ & Propose hyperparameters or config options for estimator $E$ & & & \cmark\\ 
\bottomrule
\end{tabular}}
\label{tab:semops}
\vspace{-0.4cm}
\end{table*}

%% file: sections/03-operators.tex
\section{The \textsc{SemPipes} Programming Model}
\label{sec:approach-semops}

\textsc{SemPipes} offers a novel programming model that integrates LLM-powered semantic data operators into tabular ML pipelines to tailor them to the data and task at hand. This abstraction enables a form of semi-automated pipeline construction, which delegates selected data operations to LLMs. \textsc{SemPipes} targets complex Python pipelines over multiple tabular datasets---with multimodal cells such as text, images, or audio---and supports joins, aggregations, cleaning, preprocessing, and encoding using dataframe operations, scikit-learn–style estimators, user-defined functions, and models compatible with the scikit-learn ecosystem. 

\header{Semantic Data Operators} We introduce nine semantic data operators (SemOps) for tabular ML pipelines (see \Cref{tab:semops}), falling into three categories: $(i)$~four operators for feature generation, extraction, and data augmentation; $(ii)$~three operators for data cleaning and integration; and $(iii)$~two operators for pipeline parameterization, including semantic column selection and hyperparameter range proposals for estimators~\cite{belkhiter2025prehocpredictionsautomlleveraging}. Each operator has well-defined input and output types, is configured with a natural language instruction, and applies either to dataframes or estimator parameters, integrating seamlessly with skrub, pandas, and scikit-learn. \Cref{fig:overview} illustrates a sample pipeline for a fictitious fraud detection use case.
Our operator selection draws on heuristic methods from open-source libraries such as skrub and recent studies highlighting the challenges of integrating research prototypes into production code~\cite{schafer2025usable}. Our design allows fine-grained control over which data operations are handled by LLM-driven operators, and supports development in interactive environments like Jupyter notebooks. \textsc{SemPipes} deliberately focuses on data-centric operations rather than model training, which is already well supported by easy-to-use models~\cite{grinsztajn2025tabpfn, qu2025tabicl, Grinsztajn2022whyboostingtrees} and established hyperparameter tuning methods~\cite{Feurer2019hyperparameter}.

\begin{figure*}[t!]
    \centering
    \includegraphics[width=\linewidth]
    {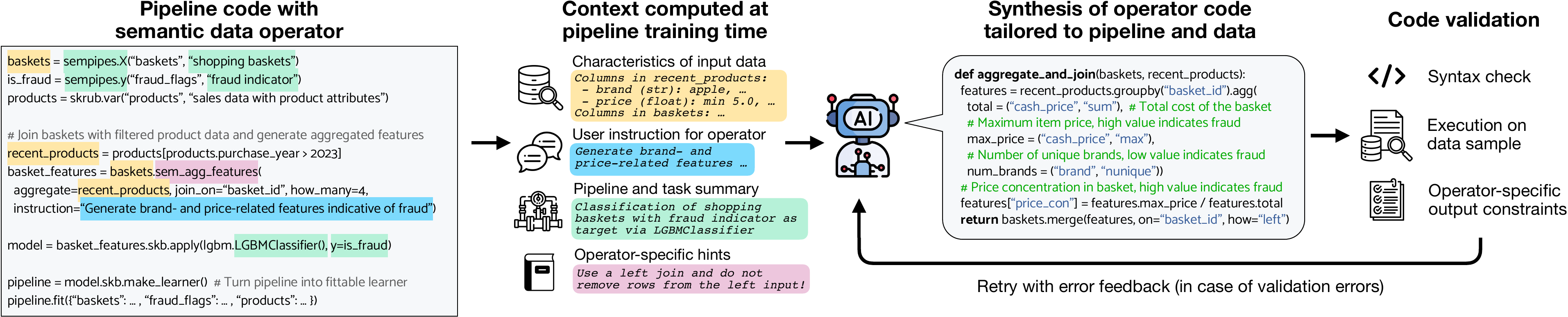}
    \vspace{-6mm}
    \caption{\textbf{Toy example of code synthesis for a semantic data operator}. At training time, \textsc{SemPipes} synthesizes custom operator code conditioned on input data characteristics (e.g., size, column names, and statistics), the natural language instruction, and pipeline context such as the learning task and model type. The synthesized code is validated by executing it on a sample of the input data and by checking operator-specific output constraints.}
    \label{fig:synth}
\vspace{-0.2cm}
\end{figure*}

\header{Execution Model} \textsc{SemPipes} leverages skrub DataOps pipelines~\cite{skrub_data_ops}, a lightweight abstraction for multi-table ML pipelines over dataframes (see \Cref{app:skrub-dataops} for a detailed introduction). DataOps represents pipelines as a computational graph whose nodes are data operations and edges denote data dependencies. This design enables lazy execution as well as reasoning about and rewriting the pipeline itself. Standard scikit-learn pipelines are restricted to a linear sequence of transformations applied to a single table with a fixed number of rows. In contrast, the DataOps abstraction elevates the entire pipeline---including multi-table dataframe operations---into a unified scikit-learn–style predictor, which seamlessly handles complex data operations across multiple tables. For that, a DataOps pipeline exposes a \texttt{fit} method for end-to-end training and a \texttt{predict} method for inference on unseen data. \textsc{SemPipes} leverages the computational graph to integrate semantic data operators as scikit-learn–style estimators.

\header{Operator-Specific Code Synthesis at Training Time} \textsc{SemPipes} synthesizes custom operator implementations at training time, conditioned on input data characteristics (e.g., size, column names, statistics), natural language instructions, and pipeline context—including the learning task and model type inferred from the source code and computational graph (see a toy example in~\Cref{fig:synth}). Depending on the operator type and user instruction, \textsc{SemPipes} synthesizes different code variants: for feature generation, it produces CPU-based dataframe and aggregation code; for multimodal feature extraction, it produces code to run lightweight pretrained models on a local GPU (e.g., HuggingFace models in zero-shot mode). Only a few LLM calls—linear in the number of semantic operators—are required at training time, producing operator code that executes efficiently without LLM calls at inference time. The operator-centric design also enables fine-grained validation of synthesized code: beyond syntax checks via parsing, \textsc{SemPipes} executes each operator on a sample of input data and verifies operator-specific output constraints, such as the expected shape of the resulting dataframe. See~\Cref{app:semop-examples} for details on the code synthesis per operator and alternative implementations that invoke LLMs directly during data processing.

%% file: sections/04-optimization.tex
\section{Optimizing Semantic Data Operations}
\label{sec:approach-opt}

To improve the data operations of a tabular ML pipeline, data scientists typically select a target operation (e.g., feature generation or missing value imputation), rewrite its code, integrate external libraries, and rerun the pipeline on a validation set to assess performance improvements. They iterate until the pipeline meets their goals or time runs out. \textsc{SemPipes} automates this workflow by tuning the synthesized code of its semantic data operators to improve predictive performance via LLM-based mutation (code modifications), guided by tree-structured evolutionary search.

\header{Semantic Pipelines} We consider a semantic ML pipeline a composition of {\em semantic operators} $o(\cdot; \phi)$ with states $\phi$ (their synthesized code) and {\em predictive operators} $f(\cdot; \theta)$ with learnable parameters~$\theta$  (e.g., a predictive model, dimensionality reduction, or an ensemble). Conceptually, each input $(x, y) \in \mathcal{D}$ is first processed by the semantic operators, whose output is subsequently used by the predictive operators to produce predictions $\hat{y} = f(o(x; \phi); \theta)$. The performance of the overall pipeline is measured by a scalar utility function $\mu(y, \hat{y})$, e.g., accuracy on hold out data. Since the semantic operators $o$ are non-differentiable, the pipeline parameters $\phi$ and $\theta$ cannot be optimized jointly via gradient-based methods. We adopt a bi-level optimization strategy~\cite{sinha2017review} that alternates between $(i)$~optimizing the learnable parameters~$\theta$ of predictive operators for a fixed set of semantic operator states, and $(ii)$~updating the semantic operator states~$\phi$ through an LLM-guided mutation procedure.
We aim to solve:
\begin{equation}
    \argmax_{\phi} \,\,
    \mathbb{E}_{(x, y) \sim \mathcal{D}}
    \Big[ \mu\big(y, f(o(x; \phi); \theta^*(\phi)) \big) \Big],
    \label{eq:outer-optimization}
\end{equation}
where the inner optimization defines the best learnable parameters $\theta^*(\phi)$ of the predictive operators for a given set of semantic operator states $\phi$ as $\smash{\theta^*(\phi) = \arg\max_{\theta} \; \mathbb{E}_{(x, y) \sim \mathcal{D}} \big[ \mu\big(y, f(o(x; \phi); \theta)\big) \big]}$. For the optimization of \Cref{eq:outer-optimization}, we employ an LLM-based mutation mechanism, which proposes improved semantic operator states guided by performance feedback and semantic reasoning about the input data, pipeline context, and past semantic operator states. Given some training data $\mathcal{D}_\text{train}$, the optimization iteratively alternates between inner-loop learning and outer-loop mutation as follows. \emph{Inner-loop learning} at step $i$ fixes the current semantic operator states to $\phi^{(i)}$, and fits the pipeline to learn the parameters of the predictive operators that maximize utility on $\mathcal{D}_\text{train}$ via $\smash{\theta^{(i)} = \arg\max_{\theta} \; \sum_{(x,y) \in \mathcal{D}_\text{train}} \mu\big(y, f(o(x; \phi^{(i)}); \theta)\big)}$. Next, \emph{outer-loop mutation} evolves the operators by first evaluating the validation utility $\sum_{(x,y) \in \mathcal{D}_\text{train}} \mu\big(y, f(o(x; \phi^{(i)}); \theta^{(i)})\big)$ on $\mathcal{D}_\text{train}$ (typically via cross-validation), and then generating new semantic operator states $\phi^{(i+1)}$ using LLM-based mutation to improve expected utility.

\setlength{\textfloatsep}{5pt}
\begin{algorithm}[t!]
  \caption{Optimization of the semantic operator states $\phi$ of a \textsc{SemPipes} pipeline with evolutionary tree search and LLM-guided mutation.}
  \label{alg:colopro}
  {\small
  \begin{algorithmic}[0]
  \Require Pipeline with semantic data operators $o$, training~data~$\mathcal{D}_\text{train}$, language model $L$, search policy {\large${\pi}$}, utility function $\mu$, budget $B$
   \State $\phi^{(0)}, m^{(0)} \gets$ empty states and memories
   \State $T \gets$ Initialize search tree with root $s^{(0)}$ based on $\phi^{(0)}$
   \For{step $i \in 0 \dots B$}      
     \If{i > 0}
        \State $s^{(i)} \gets$ next search node from policy {\large $\pi$}$(T)$
        \State $\phi^{(i)}, m^{(i)} \gets \textsc{evolve\_semops}(\mathcal{D}_\text{train}, s^{(i)}, ${\large ${\pi}$}$, L)$
     \EndIf   
     \State $\tilde{\theta} \gets$ Fit pipeline on $\mathcal{D}_\text{train}$ with fixed $\phi^{(i)}$
     \State $v^{(i)} \gets $ utility $\mu$ of pipeline with with $\phi^{(i)}, \tilde{\theta}$ on $\mathcal{D}_\text{train}$       
     \State Record outcomes $(v^{(i)}, \phi^{(i)}, m^{(i)})$ for $s^{(i)}$ in $T$    
   \EndFor
   \State \textbf{return} $\phi$ for highest utility recorded in $T$  
   \item[]
   \Procedure{evolve\_semops}{$\mathcal{D}_\text{train}, s^{(i)}, ${\large ${\pi}$}$, L$}
       \State $\mathbf{C}  \gets$ Summarize pipeline context
       \State $\mathbf{V}^{(i-1)} \gets$ utility scores $[v^{(0)}, {\scriptstyle \ldots} \; , v^{(i-1)}]$ in path to $s^{(i)}$    
       \For{$\text{semantic operator} \, o_j$ as ordered in computational DAG}
         \State $\mathbf{I}_{j} \gets$ inspirational states from $T$ selected by {\large${\pi}$}
         \State $\mathbf{M}_{j}^{(i-1)} \gets$ memories $[m^{(0)}_j, {\scriptstyle \ldots} \; , m^{(i-1)}_j]$ in path to $s^{(i)}$
         \State $\mathbf{X}_{j} \gets$ Partially fit pipeline on $\mathcal{D}_\text{train}$ up to operator $o_j$           
         \State $\phi^{(i)}_j\!, m^{(i)}_j\!\gets\!\textsc{evolve}(o_j, \mathbf{C}, \mathbf{M}_{j}^{(i-1)}\!,\mathbf{I}_{j}, \mathbf{X}_{j},\!\mathbf{V}^{(i-1)}; L)$
       \EndFor         

       \State $\textbf{return} \,\, \text{semantic operator states} \, \phi^{(i)}$ with memories $m^{(i)}$ 
   \EndProcedure   
  \end{algorithmic}}
\end{algorithm}

\header{Evolving the Code of Individual Semantic Data Operators} Next, we detail how to iteratively improve the semantic operator state $\phi_j^{(i)}$ of a semantic data operator $o_j$ during our optimization procedure. We design a function $\textsc{evolve}(o_j, \mathbf{C}, \mathbf{V}^{(i)}, \mathbf{M}_{j}^{(i)}, \mathbf{X}_{j}, \mathbf{V}^{(i)}; L) \rightarrow \phi^{(i)}_j, m^{(i)}_j$ specific to each type of semantic operator, which evolves the internal state $\smash{\phi^{(i+1)}_j \rightarrow \phi^{(i+1)}_j}$ (e.g., its synthesized code) of a semantic operator $\smash{o_j}$ via a language model~$L$. The function produces both the evolved state $\smash{\phi^{(i+1)}_j}$ as well as the corresponding memory trace $\smash{m^{(i+1)}_j}$ as output, which contains the history of attempts and reasoning for the current state of the operator. These outputs depend on the pipeline context $\mathbf{C}$ (summarizing the structure of the computational graph, the task type, and type of model used) as well as the sequence of previous utility scores $\smash{\mathbf{V}^{(i)}}$ and memory traces $\smash{\mathbf{M}_{j}^{(i)}}$, and the concrete input data $\mathbf{X}_{j}$ of the operator $o_j$ in the pipeline,. Furthermore, the function can be provided with external ``inspirations'' $\mathbf{I}_{j}$, e.g., examples of operator's code that worked well in other optimization trajectories. See \Cref{app:semop-optimization} for further details.

\header{Tree-Structured Evolutionary Search for Pipeline Optimization} We combine our iterative bi-level optimization procedure with a tree-structured evolutionary search~\cite{aygun2025ai} over semantic operator states, as shown in \Cref{alg:colopro}. Each node $s^{(i)}$ in the search tree $T$ encapsulates both the utility value $v^{(i)}$ achieved at that step and a memory trace $m^{(i)}$ from producing the corresponding semantic operator states. The traversal and expansion of the search tree are governed by a search policy~{\large $\pi$}, which selects the next node as starting point for each optimization step. The policy can be instantiated by Monte Carlo tree search~\cite{browne2012survey} or evolutionary algorithms~\cite{back1997handbook}.

%% file: sections/05-evaluation.tex
\section{Experimental Evaluation}
\label{sec:experiments}

\input{tables/main-results}

We evaluate semantic data operators as first-class, reusable primitives for both expert-written and agent-generated ML pipelines. By design, \textsc{SemPipes} is neither an AutoML system nor an end-to-end agentic pipeline generator. Instead, \textsc{SemPipes} introduces a programming abstraction to synthesize and optimize complex data operations in ML pipelines. Benchmarks evaluating fully automated pipeline generation for prediction, such as MLEBench~\cite{chan2024mle}, are an inherent mismatch for our setting. Furthermore, tabular ML pipelines encompass a much broader range of tasks than prediction alone, including data integration, cleaning, feature extraction, debugging, and augmentation, which are not well covered by existing benchmarks. We design dedicated experiments that capture these data-centric challenges and isolate the impact of semantic operators within realistic, complex pipelines. All experiments utilize Google’s \texttt{gemini-2.5-flash} model, with a temperature of 0.0 for non-optimized execution and 2.0 for optimization.

\subsection{Semantic Data Operators Improve and Simplify Tabular ML Pipelines}
\label{sec:experiments-main}

We show that optimized semantic data operators improve expert-written and agent-generated ML pipelines.

\header{Experimental Setup} We evaluate \textsc{SemPipes} on a set of challenging data-centric tasks spanning diverse ML application areas. The tasks involve complex data integration, cleaning, feature extraction and generation, debugging, and data augmentation, and are drawn from recent work at NeurIPS~\cite{chen2023hibug,qian2023synthcity}, EMNLP~\cite{lee2021micromodels}, a SIGMOD programming contest~\cite{sigmod2022}, and various Kaggle competitions. The tasks are chosen to be highly heterogeneous, e.g., their input data size varies from thousands of records to millions of records, and several of them involve multiple input tables or combine tabular data with text or images. 
We collect \emph{hand-written expert pipelines} from code repositories accompanying the respective research papers, top-ranked SIGMOD programming contest submissions, and Kaggle medalist notebooks, reflecting substantial domain expertise and engineering effort. Additionally, we synthesize \emph{agent-generated pipelines} using AIDE~\cite{jiang2025aide} and mini-SWE-agent~\cite{yang2024swe}. Despite extensive prompting and retries, at least one of these agents fails to produce a runnable or correct pipeline for six out of seven tasks, highlighting the complexity of our chosen tasks. 
Our experimentation protocol is as follows: (1)~We first evaluate the original expert and agent pipelines on each task; (2)~Then, we rewrite them as skrub DataOps pipelines and introduce semantic operators from \textsc{SemPipes}, which we provide with a natural language instruction summarizing the expert knowledge in the original pipeline. Note that we keep model choices and hyperparameters fixed, to isolate the impact of the actual data operations. We evaluate the rewritten pipelines and (3)~optimize their semantic operators for 24 iterations. We repeat each pipeline run five times, see~\Cref{app:main-setup} for details on each task's origin, evaluation metric, and settings. 

\begin{figure*}[t]
\centering
  \begin{minipage}[t]{0.32\textwidth}
  \vspace{0pt}
  \centering
  \input{tables/ROC_AUC-results}
  \end{minipage}
\hfill
  \begin{minipage}[t]{0.32\textwidth}
  \vspace{0pt}
  \centering
  \input{tables/costs-results}
  \end{minipage}
\hfill
  \begin{minipage}[t]{0.32\textwidth}
  \vspace{0pt}
  \centering
    \includegraphics[width=\columnwidth]{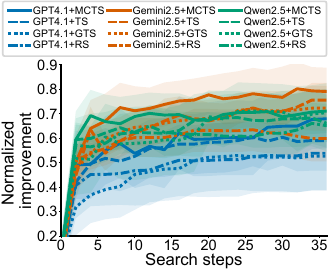}
    \vspace{-7mm}
    \caption{
    \textbf{Optimization effectiveness of LLMs for code synthesis in semantic data operators}. 
    Search policies include Monte Carlo tree search (MCTS), truncation selection (TS), greedy tree search (GTS) or random search (RS).
   }
    \label{fig:minibench}
  \end{minipage}  
\vspace{-0.4cm}
\end{figure*}

\header{Results and Discussion} \Cref{tab:main-results} summarizes the results for the 19 evaluated pipelines, including task-specific metrics and the origin of each pipeline (expert-written or agent-generated). The results confirm that \emph{semantic data operators improve the predictive performance of both expert-written and agent-generated pipelines}. In 17 of 19 cases, augmenting the original pipeline with semantic operators and instructions summarizing expert knowledge (\say{w/ SemOps \& Instructions}) improves predictive performance even without optimization. In one case, performance remains unchanged (\texttt{kaggle-movie-a}), and in another case (\texttt{sustech}), we observe a small decrease in recall by 0.001, which we attribute to the original code using a fine-tuned model, which we did not re-tune.
Optimizing the semantic data operators (\say{w/ Optimized SemOps}) yields further gains. In 18 out of 19 cases, the optimized pipelines outperform the original code (\say{Original Pipeline}), and in 17 of 19 cases they also improve over pipelines with non-optimized semantic operators. We discuss three exemplary cases (further details in \Cref{app:main-setup-sample-results}). 
First, we observe large improvements for both expert-written and agent-generated pipelines in feature engineering for predicting the rating of scrabble players, where the pipelines with optimized semantic operators achieve an 18\% to 26\% reduction in RMSE. We attribute this to the fact that the complex input data (three tables with data about 70 thousand players, who played 80 thousand games with 2 million turns) provides a rich opportunity to engineer helpful features. The semantic operator is instructed to generate player-statistics-related features that help predict the player's rating. Based on that, \textsc{SemPipes} for example improves the \texttt{kaggle-scrabble} pipeline (which applies a heavily regularized XGBoost model with a large number of shallow trees) by computing over hundred extra features, e.g., the win rate of a player when going first, the pacing of their games, or their turn efficiency for scoring.
Second, we apply \textsc{SemPipes} to the SIGMOD~2022 programming contest task, which requires feature extraction and entity blocking at scale over datasets with millions of records. Entity blocking identifies and links different records that refer to the same real-world entity. We rewrite the fifth-ranked solution (\texttt{rutgers}) on the contest's more challenging multilingual dataset, where the original approach achieves a low recall. Using \textsc{SemPipes}, we replace hundreds of lines of hand-written feature extraction and blocking code with calls to \texttt{sem\_extract\_features} and \texttt{sem\_gen\_features}, which  enrich the feature set. Our optimizations would improve \texttt{rutgers} from fifth to third place on the \href{https://dbgroup.ing.unimore.it/sigmod22contest/leaders.shtml}{public leaderboard}. 
Third, in the data annotation for model debugging task, optimizing the expert-written \texttt{hibug} pipeline with \textsc{SemPipes} results in an uplift of more than 16\% for the annotation accuracy. We attribute this improvement to the optimizer's choice of a well-working pretrained model.
We additionally find that semantic data operators simplify expert-written pipelines, see \Cref{app:main-setup-sample-results-simplification}.

In summary, \textsc{SemPipes} enables the optimization of data operations in ML pipelines, which are typically fixed in prior work that only focuses on models and hyperparameters. Our results on challenging tasks with strong expert baselines show that optimizing data operations constitutes an additional route for improving performance in tabular ML. All gains stem solely from data operations, as model choices and hyperparameters remain unchanged in our experiments. LLM-based experiments can be affected by training-time contamination, e.g., exposure to task solutions during pretraining. We believe this is unlikely to impact our results: the agent-generated pipelines perform poorly/fail on our tasks; several tasks do not derive from existing competitions; and the semantic operators already encode domain knowledge explicitly via their natural language instructions.

\subsection{Semantic Data Operators Perform On-Par or Better than Specialized Approaches}


We validate that \textsc{SemPipes} operators match or exceed the performance of specialized approaches through comparison to strong baselines. See \Cref{app:effectiveness-comparison} for further details.

\header{Experimental Setup} For feature generation with \texttt{sem\_gen\_features}, we compare against the automated feature engineering method CAAFE~\cite{hollmann2023large}.
We evaluate on three datasets using the experimental setup and precomputed feature engineering code provided by the CAAFE authors. We use a random forest and a non-tuned TabPFN~2.0~\cite{hollmann2022tabpfn} as downstream models, and report mean AUROC on the test splits. For \textsc{SemPipes}, we optimize the semantic pipelines on hold out data. For missing-value imputation with \texttt{sem\_fillna}, we compare against the prompting scheme of~\cite{narayan2022canllmswrangleyourdata} on two benchmark tasks from the paper, and measure the accuracy of the imputed values.
For zero-shot feature extraction with \texttt{sem\_extract\_features}, we compare \textsc{SemPipes}’ synthesized code against direct per-item LLM processing in \textsc{SemPipes} and semantic query processing systems Palimpzest~\cite{liu2025palimpzest} and LOTUS~\cite{patel2024semantic}. We evaluate on text (financial clause classification~\cite{kaggle-legal-clauses}), images (pneumonia detection in X-Rays~\cite{Kermany2018}) and audios (environmental sound extraction~\cite{piczak2015dataset}), processing 1,000 and 10,000 samples per modality.

\newpage
\header{Results and Discussion} \Cref{tab:caafe} shows the results for feature engineering, where CAAFE and \textsc{SemPipes} significantly improve performance for both downstream models on two out of three datasets. CAAFE yields a modest gain for random forests (RF) (4.2\%) and a negligible improvement for TabPFN (0.03\%). \textsc{SemPipes} achieves substantially larger gains, improving RFs by an average of 10.6\% and TabPFN by 4.4\% over no feature engineering. We attribute this advantage to guided tree-structured optimization, rather than CAAFE’s greedy linear search, and to \textsc{SemPipes}’ ability to customize prompts to the downstream pipeline, e.g., determining the type of used model.
For missing value imputation (\Cref{tab:mv}), \textsc{SemPipes} outperforms hand-designed prompting baselines on both datasets, with gains of 6.3\% on \texttt{Restaurant} and 1.6\% on \texttt{Buy}. 
This improvement is enabled by the design of our semantic operators, which combine automatically computed task-specific statistics, such as unique values and column types, with a customized operator instruction, containing more effective few-shot examples, presented in a human-readable format. 
For zero-shot feature extraction (\Cref{fig:feature_extraction}), \textsc{SemPipes} code generation achieves the strongest performance across image and audio modalities. \textsc{SemPipes} consistently surpasses Palimpzest and LOTUS, yielding accuracy gains of $\sim$20\% for audio and $\sim$10\% for images, while remaining substantially more cost-efficient.
For text, \textsc{SemPipes} offers a clear accuracy–cost trade-off: direct LLM processing yields the best accuracy (slightly ahead of Palimpzest and LOTUS, due to more elaborate prompts), whereas code generation is 10\% lower in accuracy but 10x cheaper, enabling scalable inference with constant cost.
Rather than relying on per-sample LLM inference, \textsc{SemPipes} synthesizes executable code that leverages pretrained, domain-specific models, such as DeBERTa~\cite{he2021deberta} for text, BiomedCLIP~\cite{zhang2024biomedclip} for images, and CLAP~\cite{wu2024largescalecontrastivelanguageaudiopretraining} for audio, enabling scalable, cost-efficient inference. When scaling from 1,000 to 10,000 data items, code synthesis in \textsc{SemPipes} incurs constant code generation, while LLM processing costs scale linearly and are therefore ten times higher.

\subsection{Optimization Effectiveness of Different LLMs for Code Synthesis in Semantic Data Operators}
\label{sec:experiments-colopro}

We evaluate LLMs and search policies for optimizing semantic operators using a benchmark of classification pipelines built on datasets provided by the skrub project (\Cref{app:main-optimizer}). We test GPT-4.1, Gemini-2.5-flash, and Qwen-coder-2.5 (32B) as synthesizers, paired with Monte Carlo tree search (MCTS)~\cite{browne2012survey}, truncation selection~\cite{back1991survey}, greedy tree search~\cite{jiang2025aide}, or random search. As shown in \Cref{fig:minibench}, all models improve predictive performance, with most gains achieved within five to eight search steps, suggesting significant optimization is possible with a modest budget. Among search policies, MCTS outperforms the other policies, which we attribute to its ability to modulate exploration and exploitation, similarly to related studies~\cite{toledo2025ai}.

%% file: tables/main-results.tex
\begin{table*}[t!]
\centering
\caption{\textbf{Results from using semantic data operators with natural language instructions in expert-written and agent-generated pipelines across a broad range of data-centric tasks.} In 18 out of 19 cases, semantic operators improve the performance over the original code. All improvements stem solely from data operations, as model choices and hyperparameters from the original code remain unchanged in our experiments.}

\vspace{-2.5mm}
{\small
\setlength{\tabcolsep}{6.5pt}
\setlength{\aboverulesep}{1.15pt}
\setlength{\belowrulesep}{0.85pt}
\begin{tabular}{lc|lcc|ccc}
\toprule
\multirow{2}{*}{\makebox[40pt][l]{\textbf{Data-Centric Task}}} & \multirow{2}{*}{\makebox[40pt][c]{\textbf{Metric}}} & \multirow{2}{*}{\makebox[50pt][c]{\textbf{Pipeline}}} & \multicolumn{2}{c|}{\textbf{Origin}} & \multicolumn{1}{c}{Original}&\multicolumn{1}{c}{w/ SemOps \&} & \multicolumn{1}{c}{w/ Optimized}\\[-0.2mm]
 &  &  & \multicolumn{1}{c}{\small Expert} & \multicolumn{1}{c|}{\small Agent} & \multicolumn{1}{c}{Pipeline} & \multicolumn{1}{c}{Instructions} & \multicolumn{1}{c}{SemOps}\\[-0.2mm]

\midrule

Extraction of clinically &  \multirow{2}{*}{F1 Score $\uparrow$} & \texttt{micromodels} & \checkmark &  & 0.665\pmS{0.01} & 0.692\pmS{0.02} & \textbf{0.729}\pmS{0.00}\\[-0.2mm]
informed features && \texttt{swe-micromodels} & & \checkmark & 0.690\pmS{0.00} &  0.692\pmS{0.02} & \textbf{0.729}\pmS{0.00}\\[-0.2mm]

\midrule   

Blocking for & \multirow{4}{*}{Recall $\uparrow$} & \texttt{rutgers} & \checkmark && 0.238\pmS{0.00}& 0.256\pmS{0.01} & \textbf{0.264}\pmS{0.01}\\[-0.2mm]
entity resolution  & & \texttt{sustech} & \checkmark && \textbf{0.322}\pmS{0.00}& 0.320\pmS{0.00} & 0.321\pmS{0.00}\\[-0.2mm]
&& \texttt{aide-sigmod} & & \checkmark & 0.043\pmS{0.00} & 0.163\pmS{0.03} & \textbf{0.165}\pmS{0.03}\\[-0.2mm]
&& \texttt{baseline-sigmod} & \checkmark & & 0.099\pmS{0.00} & 0.141\pmS{0.02} & \textbf{0.173}\pmS{0.05}\\[-0.2mm]

\midrule  

Feature engineering& \multirow{3}{*}{RMSLE $\downarrow$} & \texttt{kaggle-movie-a} &\checkmark&& 2.222\pmS{0.14} & 2.222\pmS{0.14} & \textbf{2.203}\pmS{0.12}\\[-0.2mm]
(movie revenue) &    & \texttt{kaggle-movie-b}  &\checkmark&& 2.170\pmS{0.07} & 2.129\pmS{0.15} & \textbf{2.122}\pmS{0.14}\\[-0.2mm]
&& \texttt{aide-movie} &&\checkmark& 2.355\pmS{0.14} & 2.319\pmS{0.15} & \textbf{2.269}\pmS{0.20}\\[-0.2mm]

\midrule  

  Feature engineering &   \multirow{4}{*}{RMSLE $\downarrow$} & \texttt{kaggle-house-a} &\checkmark&& 0.158\pmS{0.03} & \textbf{0.149}\pmS{0.00} & 0.157\pmS{0.03}\\[-0.2mm]
(house prices)  &    & \texttt{kaggle-house-b}  &\checkmark&& 0.159\pmS{0.03} & 0.153\pmS{0.03} & \textbf{0.150}\pmS{0.03}\\[-0.2mm]
  &    & \texttt{aide-house} & & \checkmark & 0.165\pmS{0.04} & 0.163\pmS{0.05} & \textbf{0.163}\pmS{0.04}\\[-0.2mm]
  &    & \texttt{swe-house} & & \checkmark & 0.177\pmS{0.04} & 0.168\pmS{0.04} &  \textbf{0.164}\pmS{0.03}\\[-0.2mm]  
  
\midrule  

Feature engineering  &   \multirow{3}{*}{RMSE $\downarrow$} & \texttt{kaggle-scrabble} &\checkmark&  & 176.1\pmS{24.7} & 149.2\pmS{24.8} & \textbf{144.4}\pmS{18.3}\\[-0.2mm]
(scrabble rating)  & & \texttt{aide-scrabble} & & \checkmark & 255.0\pmS{24.9} & 195.2\pmS{21.9} & \textbf{186.7}\pmS{21.1}\\[-0.2mm]
  & & \texttt{swe-scrabble} & & \checkmark & 228.5\pmS{12.9} & 210.1\pmS{43.0} & \textbf{187.8}\pmS{21.4}\\[-0.2mm]

\midrule  

Data annotation & Accuracy $\uparrow$ & \texttt{hibug} & \checkmark && 0.598\pmS{0.05} & 0.758\pmS{0.15} & \textbf{0.766}\pmS{0.16}\\[-0.2mm]  

\midrule  

Data augmentation & \multirow{2}{*}{AUROC $\uparrow$} & \texttt{sivep} & \checkmark & & 0.653\pmS{0.03} & 0.682\pmS{0.00} & \textbf{0.710}\pmS{0.04}\\[-0.2mm]
for fairness & & \texttt{swe-fair} &&\checkmark& 0.650\pmS{0.04} & 0.682\pmS{0.02} & \textbf{0.710}\pmS{0.04}\\[-0.2mm]
\bottomrule
\end{tabular}
}
\label{tab:main-results}
\vspace{-0.4cm}
\end{table*}





%


%% file: tables/ROC_AUC-results.tex

  \captionof{table}{AUROC scores without feature engineering (No FE), and for automated feature engineering on three datasets with Random Forest (RF) and TabPFN. 
  }
  \label{tab:caafe}
  \vspace{-2.5mm}
  {\small
  \setlength{\tabcolsep}{2.3pt}
  \begin{tabular}{ll|c|c|c}
  \toprule
  &\textbf{Dataset} & \multicolumn{1}{c|}{No FE} & \multicolumn{1}{c|}{CAAFE} & \multicolumn{1}{c}{SemPipes}\\
  \midrule
  \multirow{5.4}{*}{\rotatebox[origin=c]{90}{RF}}
  &\texttt{balance-} & 0.685  & 0.737  & \textbf{0.884} \\[-1mm]
  &\texttt{\ \ scale} & \pmS{0.13} & \pmS{0.19} & \pmS{0.23} \\
  &\texttt{tic-tac-} & 0.807 & 0.878 & \textbf{0.929} \\[-1mm]
  &\texttt{\ \ toe} & \pmS{0.07} & \pmS{0.11} & \pmS{0.08}\\
  &\multirow{1.8}{*}{\texttt{airlines}} & 0.629  & \textbf{0.631} & 0.625 \\[-1mm]
  &  & \pmS{0.03} & \pmS{0.03} & \pmS{0.03}\\
  \midrule
  \multirow{5.4}{*}{\rotatebox[origin=c]{90}{TabPFN}}
  &\texttt{balance-} & 0.848  & 0.863 & \textbf{0.938} \\[-1mm]
  & \texttt{\ \ scale} & \pmS{0.25} & \pmS{0.24} & \pmS{0.11}\\
  &\texttt{tic-tac-} & 0.959  & 0.945  & \textbf{0.997} \\[-1mm] 
  &\texttt{\ \ toe} &  \pmS{0.02} & \pmS{0.07} &  \pmS{0.00}\\
  &\multirow{1.8}{*}{\texttt{airlines}} & 0.647  & 0.647  & \textbf{0.650} \\[-1mm]
  & & \pmS{0.03} &  \pmS{0.03} &  \pmS{0.03}\\
  \bottomrule
  \end{tabular}
  
  }

  \vspace{-0.4cm}

%% file: tables/costs-results.tex

\newcommand{\mTmp}[1]{\multirow{1.8}{*}{\money{#1}}}

\captionof{table}{
Missing value imputation with \textsc{SemPipes} and \citeauthor{narayan2022canllmswrangleyourdata}, both methods use direct LLM processing with ten few-shot examples.
}
  \label{tab:mv}
  \vspace{-2.5mm}
  {\small
  \setlength{\tabcolsep}{2pt}
  \begin{tabular}{c|c@{\hspace{-1mm}}c|c@{\hspace{-1mm}}c}
  \toprule
  \multicolumn{5}{c}{\textbf{Missing Value Imputation}}\\
  \textbf{Dataset} & \multicolumn{2}{c|}{\citeauthor{narayan2022canllmswrangleyourdata}} & \multicolumn{2}{c}{SemPipes}\\
  \midrule
  \texttt{Restaurant} & \multicolumn{2}{c|}{0.761\,\pmS{0.00}} & \multicolumn{2}{c}{\textbf{0.827}\,\pmS{0.01}}\\
  \texttt{Buy} & \multicolumn{2}{c|}{0.923\,\pmS{0.00}} & \multicolumn{2}{c}{\textbf{0.988}\,\pmS{0.01}}\\[-0.7mm]
  \midrule
  \end{tabular}
  \setlength{\tabcolsep}{1.6pt}

  \vspace{-0.4cm}
  \captionof{figure}{Zero-shot feature extraction.}
  \label{fig:feature_extraction}
  \includegraphics[scale = 0.98]{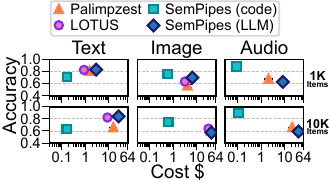}
  
  \vspace{-0.1cm}
  }
  

%% file: sections/06-conclusion.tex
\section{Conclusion}

We introduced \textsc{SemPipes}, a programming model for specifying data operations in tabular ML pipelines as semantic operators and optimizing them via LLM-based code synthesis guided by evolutionary search. 
A current limitation of our approach is the lack of support for time series or videos. Furthermore, executing LLM-synthesized code comes with potential security risks, which need to be mitigated in practice by running in a security-hardened sandbox~\cite{firejail}.
In future work, we plan to add more semantic data operators, e.g., semantic join, and to explore their value as agentic primitives in systems such as AIDE~\cite{jiang2025aide}. Furthermore, we will extend our optimizer to jointly optimize predictive performance and execution time or cost.

%% file: sections/impact-statement.tex


%% file: sections/07-appendix.tex
\makeatletter
\let\addcontentsline\icml@addcontentsline
\makeatother

\section{Appendix}

\begingroup
\setcounter{tocdepth}{2}
\etocsettocstyle{}{}
\etocsetnexttocdepth{subsection}
\localtableofcontents
\endgroup


\input{sections/a00-agents}
\input{sections/a01-related-work}
\input{sections/a02-dataops}
\input{sections/a03-semops}
\input{sections/a03b-opt}
\input{sections/a04-main-exp}
\input{sections/a05-comparison-exp}

\input{sections/a06-colopro-exp}


%% file: sections/a00-agents.tex

\subsection{Shortcomings of Agentic Pipeline Generators}
\label{app:agents-shortcomings}

We examine the shortcomings of agentic pipeline generators, and analyze issues in the benchmarks used to evaluate them for ML engineering. Some of our insights draw on the experience from working with AIDE~\cite{jiang2025aide} and mini-SWE-agent~\cite{yang2024swe} for our main experiment, as detailed in \Cref{app:main-setup-agents}.

\header{Intransparency due to Nonavailability of Agentic Trajectories} It is difficult to judge the actual benchmark results from recent papers, since many published benchmarks neither make their generated pipeline code nor the produced agentic trajectories available to the research community, despite concrete requests. For example, MLEBench~\cite{chan2024mle} refuses to provide trajectories (\url{https://github.com/openai/mle-bench/issues/28}), and AIRA-dojo~\cite{toledo2025ai} has not answered a request to provide the trajectories as of January 22nd (\url{https://github.com/facebookresearch/aira-dojo/issues/12}).

\header{High Fraction of Non-runnable Pipelines} According to our experience from working with AIDE, a significant portion of the generated pipeline code is initially non-executable, often in up to a quarter of the cases. Furthermore, we see several cases where AIDE tries to use a non-installed library, which leads to the failure of all produced code. This requires restarting the agent with either manually updated dependencies, or specialized prompts to encourage it to use different libraries. We have for example observed this when trying to include specialized single-cell data integration and cell type prediction tasks into our benchmark, where both AIDE and mini SWE-agent repeatedly failed to even ingest the input data. In the context of the SIGMOD programming contest task from \Cref{sec:experiments-main}, the agents required extensive prompting---including explicit requirements, constraints, and prohibitions---alongside a baseline solution to generate an executable pipeline. Despite this guidance, the resulting pipelines failed to meet runtime requirements and yielded performance inferior to the provided baseline. mini SWE-agent achieves a recall of $0.001$ on SIGMOD programming contest.
Finally, if AIDE or mini SWE-agent is absent from~\Cref{tab:main-results} for a task, the agent-generated pipeline was either non-runnable or produced poor results.

\header{Data Leakage Issues in Generated Pipelines} In our experience, several agent-generated pipelines exhibit data leakage issues. We often encounter a pattern where the train and test data is concatenated and jointly preprocessed (e.g., via scaling or imputation), which is common in Kaggle code, where the goal is not to train a generalizable model, but to perform well on a predefined test set. We hypothesize that LLMs pick up and reproduce this faulty pattern. 
It is hard to verify whether this data leakage issue impacts current benchmark results, since most benchmarks do not publish the final pipeline code, as discussed above. The only available generated code that we could find is from the CatDB paper~\cite{fathollahzadeh2025catdb}, where the supplemental repository contains pipelines generated with their system and with AIDE~\cite{jiang2025aide}, which recently won the MLEBench~\cite{chan2024mle} benchmark. A superficial (non-exhaustive) investigation shows that both systems tend to reproduce the data leakage issues from Kaggle, we provide a three code pointers per system as evidence:

\headerl{Examples for data leakage in AIDE pipelines}
{\small
\begin{itemize}[noitemsep,leftmargin=*]
  \item \url{https://github.com/CoDS-GCS/CatDB/blob/main/Examples/AIDE-baseline/Airline/iteration-1-RUN.py#L18}
  \item \url{https://github.com/CoDS-GCS/CatDB/blob/main/Examples/AIDE-baseline/Bike-Sharing/iteration-1-RUN.py#L14}
  \item \url{https://github.com/CoDS-GCS/CatDB/blob/main/Examples/AIDE-baseline/House-Sales/iteration-3-RUN.py#L33}
\end{itemize}}

\headerl{Examples for data leakage in CatDB pipelines} 
{\small
\begin{itemize}[noitemsep,leftmargin=*]
  \item \url{https://github.com/CoDS-GCS/CatDB/blob/main/Examples/CatDB/Airline/iteration-2-RUN.py#L31}
  \item \url{https://github.com/CoDS-GCS/CatDB/blob/main/Examples/CatDB/House-Sales/teration-2-RUN.py#L37}
  \item \url{https://github.com/CoDS-GCS/CatDB/blob/main/Examples/CatDB/Financial/iteration-2-RUN.py#L56}
\end{itemize}}

\header{Potential Impact of Data Contamination for Kaggle tasks} Another concern regarding recent agentic ML engineering benchmarks is their reliance on Kaggle tasks, which are likely represented in the pretraining corpora of the commercial models being evaluated. While this contamination is difficult to verify directly, the performance degradation observed in more recent competitions strongly suggests its presence. For instance, in MLE-bench~\cite{chan2024mle}, \texttt{gpt-4o} ceases to achieve medals for competitions launched after 2022 (see Figure 9 from~\citeauthor{chan2024mle}\citeyear{chan2024mle}). 
This shift aligns with the late 2023 cutoff for its pretraining data~(date taken from \url{https://openai.com/index/gpt-4o-system-card/}).

\header{Narrow Focus on Tabular Prediction Tasks} Another open question is how well current MLE agents work outside of Kaggle-style prediction tasks. Their prompts explicitly condition them to this narrow kind of task, see \url{https://github.com/WecoAI/aideml/blob/main/aide/agent.py#L178} for AIDE, and \url{https://github.com/facebookresearch/aira-dojo/blob/main/src/dojo/config_dataclasses/task/mlebench.py#L14} for AIRA-dojo on MLEBench.

%% file: sections/a01-related-work.tex

\subsection{Related Work}
\label{app:related-work}

The main paper contains only a shortened version of the related work, we provide the full version here. 
We discuss tabular and agentic machine learning pipelines in detail, and their limitations.

\header{Tabular Machine Learning Pipelines} Tabular ML pipelines typically prepare, integrate, and clean data using pandas dataframes~\cite{pandas2020}, and then compose so-called \say{estimators}~\cite{scikit-learn_compose_doc} from scikit-learn~\cite{pedregosa2011scikit} to encode features, train models, and perform hyperparameter search. Such pipelines are widely adopted in practice~\cite{psallidas2022data}, as of October 2025, the scikit-learn and pandas packages together exceed 600 million monthly downloads according to \texttt{pypistats.org}. Furthermore, scikit-learn’s estimator abstraction has been adopted by other popular ML libraries, including SparkML~\cite{meng2016mllib} and TensorFlow~Transform~\cite{baylor2017tfx}, and forms the foundation for ongoing research on ML pipelines~\cite{karlavs2022data,petersohn2020towards,grafberger2023automating}.
Despite its popularity, the current ecosystem for tabular ML pipelines has several major limitations. Most notably, dataframes and data operations are not first-class citizens in scikit-learn’s estimator abstraction, which is designed to apply a linear sequence of transformations to a single table with a fixed number of rows~\cite{skrub_data_ops_vs_alternatives}. As a result, common operations such as filters, joins, aggregations, and working with multiple input tables are not supported. This limitation is only beginning to be addressed by the \say{DataOps} abstraction (\Cref{app:skrub-dataops}) introduced in the skrub library~\cite{skrub_data_ops}. Integrating external expert libraries, e.g., for automated feature engineering, remains difficult~\cite{schafer2025usable}. Moreover, data operations are typically excluded from end-to-end optimization approaches~\cite{yu2021windtunnel}, as they are non-differentiable.

\header{Agentic ML Pipeline Generation} Recent work has explored fully automated, end-to-end generation of tabular prediction pipelines using agentic approaches driven by natural-language task descriptions~\cite{toledo2025ai,aygun2025ai,fathollahzadeh2025catdb,nam2025dsstardatascienceagent,fang2025mlzero,yang2025r,nam2025mlestar}. \Cref{app:agents-shortcomings} discusses the limitations of such agentic methods. We argue that semantic data operators provide a complementary abstraction for agents: reusable primitives that potentially allow agents to separate reasoning about the high-level pipeline structure from the synthesis and optimization of individual data operations. 

\header{Declarative Programming with Semantic Data Operators}
Declarative programming with LLM-powered semantic data operators integrated in SQL for querying multimodal data is an active research area in the database community~\cite{patel2024semantic,jo2024thalamusdb,liu2025palimpzest,lao2025sembench,russo2025deep}.
Major cloud vendors have recently integrated semantic operators into production database systems~\cite{googlecloud_bigquery_generativeAI_overview_2025,databricks_ai_functions_2025,snowflake_cortex_aisql_2025}.
For language model programming, DSPy~\cite{khattab2023dspy} provides a widely used declarative framework that enables automatic optimization of program prompts with respect to the underlying data and model~\cite{opsahl2024optimizing}.

\header{Program Optimization by Prompting} A recent approach to improve machine-gradeable programs is \say{optimization by prompting}~\cite{yang2023large}, in which a code-synthesizing LLM is iteratively invoked with feedback and downstream performance scores to refine its generated code. This paradigm is explored across a range of tasks, including automated feature engineering for tabular prediction~\cite{hollmann2023large}, prompt optimization for language model programming and compound AI systems~\cite{opsahl2024optimizing,agrawal2025gepareflectivepromptevolution,lee2025compound}, and algorithmic discovery in scientific domains~\cite{aygun2025ai}. ML pipelines fit well into this framework, as their performance on hold out data provides a natural optimization objective.

%% file: sections/a02-dataops.tex
\subsection{Skrub DataOps Pipelines}
\label{app:skrub-dataops}

The skrub library recently introduced DataOps pipelines~\cite{skrub_data_ops}, a lightweight abstraction for multi-table ML pipelines over dataframes, which represents pipelines as a computational graph whose nodes are data operations and whose edges encode data flow. In contrast to to standard scikit-learn pipelines, which are restricted to a linear sequence of transformations applied to a single table with a fixed number of rows, the DataOps abstraction covers complex data operations across multiple tables. Importantly, it elevates the entire pipeline—including multi-table dataframe operations—into a unified scikit-learn–style predictor, which exposes a \texttt{fit} method for end-to-end training and a \texttt{predict} method for inference on unseen data.

For further details and code examples, we refer to the online documentation and recent talks from the skrub project:
\begin{itemize}[leftmargin=*,noitemsep]
\item Complex multi-table pipelines with Data Ops\\
\url{https://skrub-data.org/stable/data_ops.html}
\item Skrub: machine learning for dataframes, PyData Paris 2025\\
\url{https://youtu.be/k9MNMDpgdAk?si=mw-HjEw-x5M0rvwE&t=694}
\item Clean code in Data Science - Skrub DataOps, Dot AI 2025\\
\url{https://youtu.be/bQS46Pgbnvc?si=wqduC_fASILSO0Ql&t=769}
\end{itemize}

%% file: sections/a03-semops.tex

\subsection{Detailed Examples for Semantic Data Operators}
\label{app:semop-examples}

We exemplarily discuss three semantic data operators in detail. We choose operators that are complex and at the center of our experiments. The full implementation of all semantic operators is available at \repofull{/sempipes/operators}.

\header{Aggregated Feature Generation with \texttt{sem\_agg\_features}} The \texttt{sem\_agg\_features}($D$, $D_\text{agg}$, $c_\text{join}$, $s$, $k$) operation creates $k$ new feature columns in the dataframe $D$ by aggregating a second dataframe $D_\text{agg}$ according to the natural language instruction $s$, and subsequently adds the aggregation results to $D$ via a join on $c_\text{join}$. Formalized in relational algebra, \textsc{SemPipes} has to synthesize a function $o_\text{synth}(D, D_\text{agg}) \coloneqq D \,\leftjoin_{c_\text{join}}\, \gamma_{c_\text{join} ; A_1, \dots, A_k}(D_\text{agg})$ that applies $k$ aggregations $A_1, \dots, A_k$ to $D$, grouped by the column $c_\text{join}$, and subsequently conducts a left outer join between $D$ and the aggregation result, with $c_\text{join}$ as join condition. To aid the LLM with the code generation, \textsc{SemPipes} engineers a rich context at training time: it computes characteristics of the input dataframes, such as their column names, column types, descriptive statistics and data samples. Furthermore, it specifies a code skeleton \texttt{sem\_agg\_join(left\_join\_column, left\_df, right\_join\_column, right\_df) -> DataFrame} with a strongly typed signature, and inspects the whole computational graph of the pipeline code to determine the type of ML task (e.g., classification or regression), the exact type of model trained, as well as natural language descriptions of the input data and labels if these are annotated in the code. We apply chain-of-thought prompting by ask the LLM to precede each aggregation function with a comment to the code that describes the generated feature, explains the rationale for choosing it, and argues how this feature adds useful real world knowledge for the downstream model. \textsc{SemPipes} has several means to validate and steer the code generation. Given a candidate function from the LLM, \textsc{SemPipes} constructs an input sample for validation by sampling 100 rows from the input dataframe $D$ and filtering $D_\text{agg}$ to match 90\% of the join keys in the sample of $D$. The candidate code is then run on these input samples, and \textsc{SemPipes} then checks that the result still contains the original number of rows and that the original columns from $D$ are retained. In case of any runtime or validation errors, a retry mechanism is applied to generate an improved version of the candidate code, by feeding back the error messages to the LLM. 
\textsc{SemPipes} also allows

\headerlu{Example} We provide a concrete example of code synthesized by \textsc{SemPipes} for a fraud detection pipeline aggregating product data from shopping baskets.

\headerl{Pipeline containing the semantic operator}
\begin{lstlisting}[
  basicstyle=\tiny\ttfamily,
  backgroundcolor=\color{lightgray},
  frame=tb,
  emph={sem_fillna,sem_refine,with_sem_features,apply_with_sem_choose,sem_choose,sem_aggregate_features,sem_extract_features,make_learner_with_context,optimize_with_colopro,serialize_as_learner,as_X,as_y,sem_clean},
  emphstyle=\color{kwdspy}\bfseries,
  morekeywords={import}
]
def pipeline() -> skrub.DataOp:
    products = skrub.var("products")
    baskets = skrub.var("data")
    labels = skrub.var("labels")

    baskets = baskets.skb.mark_as_X().skb.set_description("Potentially fraudulent shopping baskets of products")
    labels = labels.skb.mark_as_y().skb.set_description(
        "Flag indicating whether the basket is fraudulent (1) or not (0)"
    )

    aggregated = baskets.sem_agg_features(
        products,
        left_on="ID",
        right_on="basket_ID",
        nl_prompt="""
            Generate product features that are indicative of potentially fraudulent baskets, make it easy to distinguish
            anomalous baskets from regular ones! It might be helpful to combine different product statistics.
            """,
        name="basket_agg_features",
        how_many=1,
    )

    encoded = aggregated.skb.apply(TableVectorizer())
    return encoded.skb.apply(HistGradientBoostingClassifier(random_state=0), y=labels)
\end{lstlisting}

\headerl{Corresponding synthesized operator code}
\begin{lstlisting}[
  basicstyle=\tiny\ttfamily,
  backgroundcolor=\color{lightgray},
  frame=tb,
  emph={sem_fillna,sem_refine,with_sem_features,apply_with_sem_choose,sem_choose,sem_aggregate_features,sem_extract_features,make_learner_with_context,optimize_with_colopro,serialize_as_learner,as_X,as_y,sem_clean},
  emphstyle=\color{kwdspy}\bfseries,
  morekeywords={import}
]
import pandas as pd
import numpy as np

def _sem_agg_join(left_join_column, left_df, right_join_column, right_df):
    """
    Aggregates right_df features and left joins them to left_df.
    
    Args:
        left_join_column (str): The column in left_df to join on.
        left_df (pd.DataFrame): The left dataframe.
        right_join_column (str): The column in right_df to join on.
        right_df (pd.DataFrame): The right dataframe to aggregate and join.
    
    Returns:
        pd.DataFrame: The left_df with new aggregated features joined.
    """
    
    # Aggregate required features at the basket_ID level
    agg_df = right_df.groupby(right_join_column).agg(
        total_basket_cash_price=('cash_price', 'sum'),
        total_basket_product_units=('Nbr_of_prod_purchas', 'sum'),
        # Feature: num_unique_items_in_basket
        # Description: This feature counts the number of distinct 'item' categories purchased within each basket.
        # Rationale: Anomalous or fraudulent baskets might often display unusual item diversity. A very high number
        # of unique items could indicate indiscriminate "smash-and-grab" attempts to purchase various low-value
        # items, or fraudsters testing stolen card limits across multiple product types. Conversely, an extremely
        # low number of unique items (e.g., a basket with only one very expensive unique item that might be
        # resold quickly) could also be a red flag. This feature provides a quantitative measure of basket complexity,
        # which is distinct from the total monetary value or quantity of units.
        # Real World Knowledge: This feature helps to capture behavioral patterns of fraudsters who might
        # either broadly acquire a range of different products (diverse items) or narrowly target specific
        # item types (low diversity), which often deviates from typical customer purchase behavior, thereby
        # aiding in anomaly detection.
        num_unique_items_in_basket=('item', 'nunique'),
        # Feature: max_item_cash_price_in_basket
        # Description: This feature captures the maximum cash price of any single product entry within a given basket,
        # considering the 'cash_price' column which represents the price per unit before being multiplied by 'Nbr_of_prod_purchas'.
        # Rationale: Fraudulent activities might often target high-value items for quick resale (e.g., electronics, expensive furniture).
        # By isolating the highest-priced single item within a basket, we can flag baskets containing disproportionately
        # expensive products which could be targets for credit card fraud or return fraud. Conversely, baskets consisting
        # only of extremely low-priced items might also be unusual and indicate attempts to exploit minimal purchase limits.
        # Real World Knowledge: In real-world fraud scenarios, attackers often prioritize items with high resale value
        # to maximize illicit gains. This feature directly captures the 'highest risk' item within a transaction
        # from a value perspective, offering insights distinct from total basket value or average item value.
        max_item_cash_price_in_basket=('cash_price', 'max'),
        # Feature: num_service_fulfilment_items_in_basket
        # Description: This feature counts how many items within a basket are categorized as service or fulfilment.
        # Rationale: Service charges and fulfillment fees typically have a cash price of 0 or a fixed low value. Their
        # presence and quantity in a basket can reveal specific purchasing patterns. A high number of such items might be
        # anomalous, possibly indicating attempts to bypass minimum purchase requirements for certain promotions or to pad
        # transactions for other fraudulent purposes (e.g., creating fake transaction volume for card testing or obscuring
        # fewer high-value physical items with numerous low/no-cost "filler" items). This feature distinguishes nominal items
        # from physical products.
        # Real World Knowledge: Some fraud schemes leverage nominal or free items as part of a larger strategy.
        # For instance, an excessive number of services could be used to make a transaction appear legitimate while
        # high-value physical goods are being purchased with stolen credentials, or to test credit card validity with minimal
        # financial risk. This directly targets such behavioral anomalies.
        num_service_fulfilment_items_in_basket=('is_service_or_fulfilment', 'sum'),
        # Feature: basket_cash_price_std
        # Description: This feature calculates the standard deviation of `cash_price` for all items within a specific basket.
        # Rationale: A very high standard deviation indicates a basket with a wide range of item prices (e.g., combining
        # very cheap filler items with an expensive target item), which can be an anomalous purchasing pattern often seen
        # in fraudulent activities. Conversely, a very low or zero standard deviation (all items similarly priced) might
        # also be specific (e.g., bulk purchasing of identical high-value items for resale, or attempts to purchase many
        # cheap items). This feature captures the dispersion or heterogeneity of item values within a transaction.
        # Real World Knowledge: Fraudsters often have different strategies: some mix item values to obfuscate an expensive
        # purchase, while others target uniform expensive items. The standard deviation helps to identify these structural
        # anomalies in the basket's value composition that deviate from typical customer buying habits.
        basket_cash_price_std=('cash_price', 'std'),
        # Feature: num_unique_makes_in_basket
        # Description: This feature counts the number of distinct 'make' (manufacturer/brand) types present within each basket.
        # Rationale: Similar to item diversity, the diversity of brands can be indicative of fraud. An unusually low number of unique
        # makes (e.g., buying many items all from a single, specific manufacturer like 'APPLE' or 'RETAILER') could signify
        # targeted acquisition of highly resellable brand-specific goods, often associated with fraud. Conversely, a basket
        # with an exceptionally high number of unique manufacturers might indicate a scatter-gun approach to maximize purchases
        # from a stolen card or to appear more random.
        # Real World Knowledge: Fraudsters may specialize in certain brands due to higher demand or easier liquidation.
        # This feature helps differentiate legitimate brand loyalty from suspicious focused purchasing, and captures another
        # dimension of basket composition distinct from item category or value distribution.
        num_unique_makes_in_basket=('make', 'nunique'),
        # Feature: min_item_cash_price_in_basket
        # Description: This feature calculates the minimum cash price of any product line within a given basket.
        # Rationale: This feature complements 'max_item_cash_price_in_basket' to give a fuller picture of the price range within a basket.
        # Fraudulent baskets might intentionally include very low-priced or zero-priced items (e.g., services or cheap accessories
        # as 'filler' items) alongside a primary high-value target to make the transaction appear more complex or less suspicious.
        # Conversely, a very high minimum cash price (all items being expensive) could also signal targeted, high-value fraud.
        # This feature highlights the lower boundary of value within a transaction.
        # Real World Knowledge: Fraudsters often leverage a mix of high and low-value items. This feature helps in identifying
        # baskets where low-cost items are disproportionately present, indicating strategies like credit card testing (minimal
        # financial risk) or creating "busy" transactions to hide illicit activities involving higher-value goods.
        min_item_cash_price_in_basket=('cash_price', 'min'),
        # Feature: mean_item_cash_price_in_basket
        # Description: This feature calculates the average cash price across all *line items* within a basket, providing a central tendency of item values.
        # Rationale: The mean price can expose unusual pricing patterns. A very low average could suggest credit card testing with minimal value
        # purchases, or an attempt to generate fake transaction volume. A very high average could indicate a targeted acquisition
        # of expensive items using fraudulent means. It differs from `basket_item_value_per_product_unit` by averaging per distinct line item
        # (which has its own `cash_price` and `Nbr_of_prod_purchas`), rather than per individual unit of product (`Nbr_of_prod_purchas`).
        # Real World Knowledge: Customers usually buy a mix of items. An abnormal average price, especially when contrasted with the overall basket value,
        # number of items, or other value distributions (like std dev), can be a strong indicator of fraudulent behavior where specific
        # item value profiles are sought.
        mean_item_cash_price_in_basket=('cash_price', 'mean')
    ).reset_index()

    # Feature: basket_item_value_per_product_unit
    # Description: This feature calculates the average cash value contributed by each individual product unit within a basket.
    # Rationale: Fraudulent baskets might exhibit anomalous patterns in terms of value per product unit. An unusually low
    # 'basket_item_value_per_product_unit' could signal fraudulent acquisition of many items for very low or no cost
    # (e.g., coupon abuse, free item exploits beyond legitimate promotions). Conversely, a very high 'basket_item_value_per_product_unit'
    # (few very expensive items) might indicate high-value target transactions for credit card fraud or return fraud.
    # This feature combines total basket value and total item quantity.
    # Real World Knowledge: This normalized measure highlights anomalies in pricing behavior by showing the efficiency of value acquisition per unit,
    # helping distinguish legitimate bulk purchases or single high-value items from suspicious transactions.
    agg_df['basket_item_value_per_product_unit'] = agg_df['total_basket_cash_price'] / agg_df['total_basket_product_units']
    
    # Handle potential division by zero for 'basket_item_value_per_product_unit'.
    # As per dataset description, `Nbr_of_prod_purchas` minimum is 1, implying `total_basket_product_units` will typically be >= 1 for valid baskets.
    # If `total_basket_product_units` is 0, it means either an empty basket or only items with Nbr_of_prod_purchas=0, in which case the value should be 0.
    agg_df.loc[agg_df['total_basket_product_units'] == 0, 'basket_item_value_per_product_unit'] = 0 
    
    # Feature: basket_price_concentration_max
    # Description: This feature calculates the ratio of the maximum single item's cash price to the total cash price of the entire basket.
    # It measures how much of the basket's total value is concentrated in its most expensive item.
    # Rationale: Fraudulent transactions often involve purchasing one or two very high-value items, potentially mixed with cheaper "filler" items
    # to create a seemingly legitimate but large basket. A high value for this ratio indicates that a single item accounts for a
    # significant portion of the basket's value, which can be a strong anomaly indicator for fraud types focusing on high-value asset acquisition.
    # Real World Knowledge: Real-world fraud commonly targets high-value, easily resellable goods. This ratio directly quantifies the extent to which
    # a basket is dominated by such a "target" item, distinguishing it from general purchases of several moderately priced items or bulk low-value goods.
    agg_df['basket_price_concentration_max'] = agg_df['max_item_cash_price_in_basket'] / agg_df['total_basket_cash_price']

    # Handle potential division by zero for 'basket_price_concentration_max' (when total_basket_cash_price is 0).
    # If the total price is zero, it means the basket only contains items with zero cash_price. In this case,
    # the max item price is also zero, and the concentration can be considered 0.
    agg_df.loc[agg_df['total_basket_cash_price'] == 0, 'basket_price_concentration_max'] = 0

    # Fill NaNs for basket_cash_price_std (occurs when a basket has only one item) with 0.
    # A single item basket has no variation in price, so its standard deviation is 0.
    agg_df['basket_cash_price_std'] = agg_df['basket_cash_price_std'].fillna(0)

    # Drop the intermediate sum columns used for calculations, as they are not the final desired features
    agg_df = agg_df.drop(columns=['total_basket_cash_price', 'total_basket_product_units'])

    # Rename the right_join_column in agg_df to match left_join_column in left_df for merging
    agg_df = agg_df.rename(columns={right_join_column: left_join_column})
    
    # Left join the aggregated features back to the main dataframe
    left_df = left_df.merge(agg_df, on=left_join_column, how='left')
    
    return left_df
\end{lstlisting}

\header{Feature Extraction with \texttt{sem\_extract\_features}} The \texttt{sem\_extract\_features}(D, $C_\text{out}$, $s$) operation extracts the feature columns $C_\text{out}$ from dataframe $D$ following the corresponding natural language instructions $s$. To synthesize code for this operator, \textsc{SemPipes} first determines the modalities (text, image, audio) of all input columns of $D$ and provides a code skeleton \texttt{sem\_extract\_features(df: DataFrame, features\_to\_extract: list[dict[str, object]]) -> DataFrame} with a typed signature. Next, the LLM is prompted to generate custom extraction code per output column $c$ (with user-provided instruction $s$) for all feature columns $(c, s) \in (C_\text{out}, S_\text{out})$. The prompt asks the LLM to synthesize extraction code using appropriate pretrained models from HuggingFace's transformers library in zero-shot mode, to choose the available GPU with least load, to use batching to make efficient use of the GPU and display a progress bar for users. \textsc{SemPipes} validates candidate code on a sample of 50 input tuples and checks if the requested feature columns are added. In case of any runtime or validation errors, a retry mechanism is applied to generate an improved version of the candidate code, by feeding back the error messages to the LLM. 

\headerlu{Example} We provide a concrete example of code synthesized by \textsc{SemPipes} for feature extraction in a tweet classification pipeline, where \textsc{SemPipes} selects different models for sentiment detection, toxicity detection and named entity recognition. 

\headerl{Pipeline containing the semantic operator}
\begin{lstlisting}[
  basicstyle=\tiny\ttfamily,
  backgroundcolor=\color{lightgray},
  frame=tb,
  emph={sem_fillna,sem_refine,with_sem_features,apply_with_sem_choose,sem_choose,sem_aggregate_features,sem_extract_features,make_learner_with_context,optimize_with_colopro,serialize_as_learner,as_X,as_y,sem_clean},
  emphstyle=\color{kwdspy}\bfseries,
  morekeywords={import}
]
def pipeline() -> skrub.DataOp:
    tweets = skrub.var("data")
    labels = skrub.var("labels")

    tweets = tweets.skb.mark_as_X()
    labels = labels.skb.mark_as_y()

    output_columns = {
        "maybe_toxic": "Respond with '1' if a tweet is toxic and '0' if its text is not toxic.",
        "sentiment_label": "Classify the overall sentiment of the tweet as 'positive', 'negative', or 'neutral.",
    }

    labeled_tweets = tweets.sem_extract_features(
        nl_prompt="Extract features from textual tweets that could help predict toxicity.",
        input_columns=["text"],
        name="tweet_features",
        output_columns=output_columns,
        generate_via_code=True,
        print_code_to_console=True,
    )

    encoded_tweets = labeled_tweets.skb.apply(TableVectorizer())
    return encoded_tweets.skb.apply(RandomForestClassifier(random_state=0), y=labels)
\end{lstlisting}

\headerl{Corresponding synthesized operator code}
\begin{lstlisting}[
  basicstyle=\tiny\ttfamily,
  backgroundcolor=\color{lightgray},
  frame=tb,
  emph={sem_fillna,sem_refine,with_sem_features,apply_with_sem_choose,sem_choose,sem_aggregate_features,sem_extract_features,make_learner_with_context,optimize_with_colopro,serialize_as_learner,as_X,as_y,sem_clean},
  emphstyle=\color{kwdspy}\bfseries,
  morekeywords={import}
]
import torch
import pandas as pd
from tqdm.auto import tqdm
from transformers import (
    AutoTokenizer,
    AutoModelForSequenceClassification,
    pipeline,
)

def pick_gpu_by_free_mem_torch():
    assert torch.cuda.is_available(), "No CUDA device"
    free = []
    for i in range(torch.cuda.device_count()):
        # returns (free, total) in bytes for that device
        f, t = torch.cuda.mem_get_info(i)
        free.append((f, i))
    free.sort(reverse=True)
    _, idx = free[0]
    print(f"Chosen GPU: {idx}")
    return idx

gpu_idx = pick_gpu_by_free_mem_torch()
device = torch.device(f"cuda:{gpu_idx}") # Use the selected GPU for model inference

features_to_extract = [
    {'feature_name': 'sentiment_label', 'feature_prompt': "Classify the overall sentiment of the tweet as 'positive', 'negative', or 'neutral.", 'input_columns': ['text']},
    {'feature_name': 'maybe_toxic', 'feature_prompt': "Respond with '1' if a tweet is toxic and '0' if its text is not toxic.", 'input_columns': ['text']},
    {'feature_name': 'swear_words', 'feature_prompt': 'A comma separated list of all the swear words in the text', 'input_columns': ['text']}
]

def extract_features(df: pd.DataFrame, features_to_extract: list[dict[str, object]]) -> pd.DataFrame:
    # Sentiment analysis pipeline (zero-shot, robust)
    sentiment_model_name = "cardiffnlp/twitter-roberta-base-sentiment-latest"
    sentiment_pipe = pipeline(
        "sentiment-analysis",
        model=sentiment_model_name,
        tokenizer=sentiment_model_name,
        device=device.index,
        batch_size=32,
    )

    # Toxicity detection pipeline (zero-shot, robust)
    toxic_model_name = "unitary/toxic-bert"
    toxic_pipe = pipeline(
        "text-classification",
        model=toxic_model_name,
        tokenizer=toxic_model_name,
        device=device.index,
        batch_size=32,
    )

    # Swear word extraction: use a zero-shot NER model, then filter for profanities
    # We'll use a simple list of common swear words for demonstration
    swear_words_set = {
        # REMOVED FROM PAPER DUE TO OFFENSIVE NATURE
    }
    # For NER, use a robust model
    ner_model_name = "dslim/bert-base-NER"
    ner_pipe = pipeline(
        "ner",
        model=ner_model_name,
        tokenizer=ner_model_name,
        device=device.index,
        aggregation_strategy="simple",
        batch_size=32,
    )

    # Prepare output columns
    df = df.copy()
    text_col = "text"
    batch_size = 32

    # 1. Sentiment label
    sentiment_labels = []
    for i in tqdm(range(0, len(df), batch_size), desc="Extracting sentiment_label"):
        batch_texts = df.iloc[i:i+batch_size][text_col].tolist()
        results = sentiment_pipe(batch_texts)
        # Each result is a dict with 'label' and 'score'
        # Map to 'positive', 'negative', 'neutral'
        mapped = []
        for r in results:
            label = r['label'].lower()
            if label in ['positive', 'negative', 'neutral']:
                mapped.append(label)
            elif label == 'LABEL_0':
                mapped.append('negative')
            elif label == 'LABEL_1':
                mapped.append('neutral')
            elif label == 'LABEL_2':
                mapped.append('positive')
            else:
                mapped.append('neutral')
        sentiment_labels.extend(mapped)
    df['sentiment_label'] = sentiment_labels

    # 2. maybe_toxic
    maybe_toxic = []
    for i in tqdm(range(0, len(df), batch_size), desc="Extracting maybe_toxic"):
        batch_texts = df.iloc[i:i+batch_size][text_col].tolist()
        results = toxic_pipe(batch_texts)
        # Each result is a dict with 'label' and 'score'
        for r in results:
            label = r['label'].lower()
            if "toxic" in label:
                maybe_toxic.append(1)
            else:
                maybe_toxic.append(0)
    df['maybe_toxic'] = maybe_toxic

    # 3. swear_words
    # We'll do simple tokenization and match against the swear_words_set
    import re
    def extract_swear_words(text):
        tokens = re.findall(r"\b\w+\b", text.lower())
        found = [w for w in tokens if w in swear_words_set]
        return ", ".join(sorted(set(found)))

    swear_words_list = []
    for i in tqdm(range(0, len(df), batch_size), desc="Extracting swear_words"):
        batch_texts = df.iloc[i:i+batch_size][text_col].tolist()
        batch_swears = [extract_swear_words(t) for t in batch_texts]
        swear_words_list.extend(batch_swears)
    df['swear_words'] = swear_words_list

    return df
\end{lstlisting}

\header{Feature Generation with \texttt{sem\_gen\_features}} The $\texttt{sem\_gen\_features}(D, k, s)$ operator generates $k$ additional feature columns in the dataframe $D$ based on the natural language instruction $s$. Our implementation is based on the prompts from the CAAFE~\cite{hollmann2023large} approach, available under an open license at \url{https://github.com/noahho/CAAFE}. A major difference is that \textsc{SemPipes} adds more context about the actual pipeline at hand during code synthesis, for example by inspecting the computational DAG of the pipeline to determine the type of model used, the problem at hand (e.g., regression or classification), descriptions of the target column, and to capture the exact input data of the operator (and not only the pipeline's input dataset, which might already have been altered by other operations).

\headerlu{Example} We provide a concrete example of code synthesized by \textsc{SemPipes} for feature generation in a rewritten Kaggle pipeline for predicting the box office revenue of movies.

\headerl{Pipeline containing the semantic operator}
\begin{lstlisting}[
  basicstyle=\tiny\ttfamily,
  backgroundcolor=\color{lightgray},
  frame=tb,
  emph={sem_fillna,sem_refine,with_sem_features,apply_with_sem_choose,sem_choose,sem_aggregate_features,sem_extract_features,make_learner_with_context,optimize_with_colopro,serialize_as_learner,as_X,as_y,sem_clean},
  emphstyle=\color{kwdspy}\bfseries,
  morekeywords={import}
]
def sempipes_pipeline():
    data = skrub.var("data")
    revenue = data.revenue
    data = data.drop(columns=["revenue"])

    movie_stats = data.skb.mark_as_X()
    movie_stats = movie_stats.skb.set_description("""
    In a world where movies made an estimated $41.7 billion in 2018, the film industry is more popular than ever. But what movies make the most money at the box office? How much does a director matter? Or the budget? For some movies, it's 'You had me at Hello.' For others, the trailer falls short of expectations and you think 'What we have here is a failure to communicate.'' In this competition, you're presented with metadata on several thousand past films from The Movie Database to try and predict their overall worldwide box office revenue. Data points provided include cast, crew, plot keywords, budget, posters, release dates, languages, production companies, and countries. You can collect other publicly available data to use in your model predictions, but in the spirit of this competition, use only data that would have been available before a movie's release. It is your job to predict the international box office revenue for each movie. For each id in the test set, you must predict the value of the revenue variable. Submissions are evaluated on Root-Mean-Squared-Logarithmic-Error (RMSLE) between the predicted value and the actual revenue. Logs are taken to not overweight blockbuster revenue movies.
    """)

    revenue = revenue.skb.mark_as_y()
    revenue = revenue.skb.set_description("the international box office revenue for a movie")

    movie_stats = movie_stats.sem_gen_features(
        nl_prompt="""
            Create additional features that could help predict the box office revenue of a movie.
            Consider aspects like genre, production details, cast, crew, and any other relevant information
            that could influence a movie's financial success. Some of the attributes are in JSON format,
            so you might need to parse them to extract useful information.
        """,
        name="additional_movie_features",
        how_many=25,
    )

    json_columns = [
        "belongs_to_collection",
        "genres",
        "production_companies",
        "production_countries",
        "spoken_languages",
        "Keywords",
        "cast",
        "crew",
    ]

    def cleanup_column_names(df):
        df.columns = df.columns.str.replace(r"[^A-Za-z0-9_]+", "_", regex=True)
        return df

    movie_stats = movie_stats.drop(columns=json_columns)

    encoder = skrub.TableVectorizer()

    X = movie_stats.skb.apply(encoder)
    X = X.fillna(-999)

    X = X.skb.apply_func(cleanup_column_names)

    y_log = revenue.skb.apply_func(np.log1p)

    params = {
        "objective": "regression_l1",
        "metric": "rmse",
        "n_estimators": 3000,
        "learning_rate": 0.003,
        "feature_fraction": 0.7,
        "bagging_fraction": 0.7,
        "bagging_freq": 1,
        "lambda_l1": 0.1,
        "lambda_l2": 0.1,
        "num_leaves": 31,
        "verbose": -1,
        "n_jobs": -1,
        "seed": 42,
        "boosting_type": "gbdt",
    }
    model = lgb.LGBMRegressor(**params)
    predictions = X.skb.apply(model, y=y_log)

    def exp_if_transform(outputs, mode=skrub.eval_mode()):
        if mode in {"transform", "predict"}:
            return np.expm1(outputs)
        return outputs

    return predictions.skb.apply_func(exp_if_transform)
\end{lstlisting}

\headerl{Corresponding synthesized operator code}
\begin{lstlisting}[
  basicstyle=\tiny\ttfamily,
  backgroundcolor=\color{lightgray},
  frame=tb,
  emph={sem_fillna,sem_refine,with_sem_features,apply_with_sem_choose,sem_choose,sem_aggregate_features,sem_extract_features,make_learner_with_context,optimize_with_colopro,serialize_as_learner,as_X,as_y,sem_clean},
  emphstyle=\color{kwdspy}\bfseries,
  morekeywords={import}
]
import json
import ast
import numpy as np

# (Feature: has_collection - Binary indicator if the movie belongs to a collection)
# Input samples: 'belongs_to_collection': [nan, nan, nan]
df['has_collection'] = df['belongs_to_collection'].apply(lambda x: 1 if isinstance(x, str) and len(x) > 2 else 0)

# (Helper function for safe JSON parsing)
def safe_literal_eval(x):
    if isinstance(x, str) and x.strip():
        try:
            return ast.literal_eval(x)
        except (ValueError, SyntaxError):
            return []
    return []

# (Feature: num_genres - Number of genres associated with the movie)
# Input samples: 'genres': ["[{'id': 18, 'name': 'Drama'}]", "[{'id': 18, 'name': 'Drama'}, {'id': 53, 'name': 'Thriller'}]", "[{'id': 10751, 'name': 'Family'}, {'id': 16, 'name': 'Animation'}, {'id': 35, 'name': 'Comedy'}, {'id': 12, 'name': 'Adventure'}]"]
df['parsed_genres'] = df['genres'].apply(safe_literal_eval)
df['num_genres'] = df['parsed_genres'].apply(len)

# (Feature: genre_is_Drama - Binary indicator for 'Drama' genre)
# (Feature: genre_is_Comedy - Binary indicator for 'Comedy' genre)
# (Feature: genre_is_Thriller - Binary indicator for 'Thriller' genre)
# (Feature: genre_is_Action - Binary indicator for 'Action' genre)
# Input samples: 'genres': ["[{'id': 18, 'name': 'Drama'}]", "[{'id': 18, 'name': 'Drama'}, {'id': 53, 'name': 'Thriller'}]", "[{'id': 10751, 'name': 'Family'}, {'id': 16, 'name': 'Animation'}, {'id': 35, 'name': 'Comedy'}, {'id': 12, 'name': 'Adventure'}]"]
df['genre_is_Drama'] = df['parsed_genres'].apply(lambda x: 1 if {'id': 18, 'name': 'Drama'} in x else 0)
df['genre_is_Comedy'] = df['parsed_genres'].apply(lambda x: 1 if {'id': 35, 'name': 'Comedy'} in x else 0)
df['genre_is_Thriller'] = df['parsed_genres'].apply(lambda x: 1 if {'id': 53, 'name': 'Thriller'} in x else 0)
df['genre_is_Action'] = df['parsed_genres'].apply(lambda x: 1 if {'id': 28, 'name': 'Action'} in x else 0)
del df['parsed_genres']

# (Feature: num_production_companies - Number of production companies)
# Input samples: 'production_companies': ["[...", "{'name': 'DiNovi Pictures', 'id': 813}]", "[...]"]
df['parsed_production_companies'] = df['production_companies'].apply(safe_literal_eval)
df['num_production_companies'] = df['parsed_production_companies'].apply(len)

# (Feature: has_WarnerBros - Binary indicator if Warner Bros. is a production company)
# Input samples: 'production_companies': ["[{'name': 'Twentieth Century Fox Film Corporation', 'id': 306}, {'name': 'Regency Enterprises', 'id': 508}, {'name': 'Fox 2000 Pictures', 'id': 711}, {'name': 'Taurus Film', 'id': 20555}, {'name': 'Linson Films', 'id': 54050}, {'name': 'Atman Entertainment', 'id': 54051}, {'name': 'Knickerbocker Films', 'id': 54052}]", "[{'name': 'DiNovi Pictures', 'id': 813}]", "[{'name': 'Aardman Animations', 'id': 297}, {'name': 'Studio Canal', 'id': 5870}, {'name': 'Anton Capital Entertainment (ACE)', 'id': 20664}]"]
df['has_WarnerBros'] = df['parsed_production_companies'].apply(lambda x: 1 if any(comp['name'] == 'Warner Bros.' for comp in x) else 0)

# (Feature: num_production_countries - Number of production countries)
# Input samples: 'production_countries': ["[{'iso_3166_1': 'DE', 'name': 'Germany'}, {'iso_3166_1': 'US', 'name': 'United States of America'}]", "[{'iso_3166_1': 'US', 'name': 'United States of America'}]", "[{'iso_3166_1': 'FR', 'name': 'France'}, {'iso_3166_1': 'GB', 'name': 'United Kingdom'}]"]
df['parsed_production_countries'] = df['production_countries'].apply(safe_literal_eval)
df['num_production_countries'] = df['parsed_production_countries'].apply(len)

# (Feature: is_US_production - Binary indicator if the US is a production country)
# Input samples: 'production_countries': ["[{'iso_3166_1': 'DE', 'name': 'Germany'}, {'iso_3166_1': 'US', 'name': 'United States of America'}]", "[{'iso_3166_1': 'US', 'name': 'United States of America'}]", "[{'iso_3166_1': 'FR', 'name': 'France'}, {'iso_3166_1': 'GB', 'name': 'United Kingdom'}]"]
df['is_US_production'] = df['parsed_production_countries'].apply(lambda x: 1 if any(country['iso_3166_1'] == 'US' for country in x) else 0)
del df['parsed_production_countries']
del df['parsed_production_companies']

# (Feature: num_spoken_languages - Number of spoken languages in the movie)
# Input samples: 'spoken_languages': ["[{'iso_639_1': 'en', 'name': 'English'}]", "[{'iso_639_1': 'en', 'name': 'English'}]", "[{'iso_639_1': 'en', 'name': 'English'}]"]
df['parsed_spoken_languages'] = df['spoken_languages'].apply(safe_literal_eval)
df['num_spoken_languages'] = df['parsed_spoken_languages'].apply(len)

# (Feature: is_English_spoken - Binary indicator if English is a spoken language)
# Input samples: 'spoken_languages': ["[{'iso_639_1': 'en', 'name': 'English'}]", "[{'iso_639_1': 'en', 'name': 'English'}]", "[{'iso_639_1': 'en', 'name': 'English'}]"]
df['is_English_spoken'] = df['parsed_spoken_languages'].apply(lambda x: 1 if any(lang['iso_639_1'] == 'en' for lang in x) else 0)
del df['parsed_spoken_languages']

# (Feature: num_keywords - Number of keywords associated with the movie)
# Input samples: 'Keywords': ["[{'id': 825, 'name': 'support group'}, {'id': 851, 'name': 'dual identity'}, {'id': 1541, 'name': 'nihilism'}, {'id': 3927, 'name': 'rage and hate'}, {'id': 4142, 'name': 'insomnia'}, {'id': 4565, 'name': 'dystopia'}, {'id': 14819, 'name': 'violence'}]", "[{'id': 5982, 'name': 'ex husband'}, {'id': 187056, 'name': 'woman director'}]", "[{'id': 548, 'name': 'countryside'}, {'id': 3020, 'name': 'sheep'}, {'id': 10988, 'name': 'based on tv series'}, {'id': 11469, 'name': 'memory loss'}, {'id': 11612, 'name': 'hospital'}, {'id': 12392, 'name': 'best friend'}, {'id': 14828, 'name': 'city'}, {'id': 15162, 'name': 'dog'}, {'id': 18118, 'name': 'disguise'}, {'id': 154846, 'name': 'farmer'}, {'id': 156596, 'name': 'no dialogue'}, {'id': 196705, 'name': 'hospitalization'}, {'id': 197065, 'name': 'claymation'}, {'id': 197361, 'name': 'hairstylist'}, {'id': 219906, 'name': 'sheep farm'}]"]
df['parsed_Keywords'] = df['Keywords'].apply(safe_literal_eval)
df['num_keywords'] = df['parsed_Keywords'].apply(len)
del df['parsed_Keywords']

# (Feature: num_cast - Total number of cast members)
# (Feature: num_male_cast - Number of male cast members (gender 2))
# (Feature: num_female_cast - Number of female cast members (gender 1))
# Input samples: 'cast': ['[{\'cast_id\': 4, \'character\': \'The Narrator\', \'credit_id\': \'52fe4250c3a36847f80149f3\', \'gender\': 2, \'id\': 819, \'name\': \'Edward Norton\', \'order\': 0, \'profile_path\': \'/eIkFHNlfretLS1spAcIoihKUS62.jpg\'}]']
df['parsed_cast'] = df['cast'].apply(safe_literal_eval)
df['num_cast'] = df['parsed_cast'].apply(len)
df['num_male_cast'] = df['parsed_cast'].apply(lambda x: sum(1 for c in x if c.get('gender') == 2))
df['num_female_cast'] = df['parsed_cast'].apply(lambda x: sum(1 for c in x if c.get('gender') == 1))

# (Feature: cast_popularity - Aggregate cast popularity based on their order in cast list, inverse of order for weighted average)
# Input samples: 'cast': ['[{\'cast_id\': 4, \'character\': \'The Narrator\', \'credit_id\': \'52fe4250c3a36847f80149f3\', \'gender\': 2, \'id\': 819, \'name\': \'Edward Norton\', \'order\': 0, \'profile_path\': \'/eIkFHNlfretLS1spAcIoihKUS62.jpg\'}, {\'cast_id\': 5, \'character\': \'Tyler Durden\', \'credit_id\': \'52fe4250c3a36847f80149f7\', \'gender\': 2, \'id\': 287, \'name\': \'Brad Pitt\', \'order\': 1, \'profile_path\': \'/ejYIW1enUcGJ9GS3Bs34mtONwWS.jpg\'}]', ..., '[{\'cast_id\': 8, \'character\': \'himself\', \'credit_id\': \'52fe43e2c3a368484e003bab\', \'gender\': 2, \'id\': 78029, \'name\': \'Martin Lawrence\', \'order\': 0, \'profile_path\': \'/uBVQuIdS3wGCpYaZFEeKZBuiZ2l.jpg\'}]']
df['cast_avg_order'] = df['parsed_cast'].apply(lambda x: np.mean([c.get('order', 0) for c in x]) if x else 0)
df['cast_min_order'] = df['parsed_cast'].apply(lambda x: np.min([c.get('order', 0) for c in x]) if x else 0)
del df['parsed_cast']

# (Feature: num_directors - Number of unique directors)
# (Feature: num_producers - Number of unique producers)
# (Feature: num_writers - Number of unique writers)
# Input samples: 'crew': ["[{'credit_id': '55731b8192514111610027d7', 'department': 'Production', 'gender': 2, 'id': 376, 'job': 'Executive Producer', 'name': 'Arnon Milchan', 'profile_path': '/5crR5twLRcIdvRR06dB1O0EQ8x0.jpg'}]", "[...]", "[...]"]
df['parsed_crew'] = df['crew'].apply(safe_literal_eval)
df['num_crew'] = df['parsed_crew'].apply(len) # Total number of crew members listed
df['num_directors'] = df['parsed_crew'].apply(lambda x: len(set(c['id'] for c in x if c.get('job') == 'Director')))
df['num_producers'] = df['parsed_crew'].apply(lambda x: len(set(c['id'] for c in x if c.get('job') in ['Producer', 'Executive Producer', 'Associate Producer', 'Co-Producer'])))
df['num_writers'] = df['parsed_crew'].apply(lambda x: len(set(c['id'] for c in x if c.get('job') in ['Screenplay', 'Writer', 'Novel', 'Author'])))

# (Feature: crew_has_male_director - Binary flag for male director)
# (Feature: crew_has_female_director - Binary flag for female director)
# Input samples: 'crew': ["[{'credit_id': '52fe4250c3a36847f8014a47', 'department': 'Directing', 'gender': 2, 'id': 7467, 'job': 'Director', 'name': 'David Fincher', 'profile_path': '/dcBHejOsKvzVZVozWJAPzYthb8X.jpg'}]"]
df['crew_has_male_director'] = df['parsed_crew'].apply(lambda x: 1 if any(c.get('job') == 'Director' and c.get('gender') == 2 for c in x) else 0)
df['crew_has_female_director'] = df['parsed_crew'].apply(lambda x: 1 if any(c.get('job') == 'Director' and c.get('gender') == 1 for c in x) else 0)
del df['parsed_crew']

import numpy as np
import ast # For safe_literal_eval
import json # In case any json parsing is still needed, though ast is preferred for python literal strings.

# (Helper function for safe literal_eval - ensure it's available for this block)
def safe_literal_eval(x):
    if isinstance(x, str) and x.strip():
        try:
            # Handle potential outer quotes if data is double-encoded
            if x.startswith("'") and x.endswith("'"):
                x = x[1:-1]
            return ast.literal_eval(x)
        except (ValueError, SyntaxError):
            return None
    return None

# (Feature: collection_id - Extract id from 'belongs_to_collection')
# Input samples: 'belongs_to_collection': [nan, nan, nan, "[{'id': 107467, 'name': 'Best Movie - Collection', 'poster_path': None, 'backdrop_path': None}]", nan]
df['parsed_collection'] = df['belongs_to_collection'].apply(safe_literal_eval)
df['collection_id'] = df['parsed_collection'].apply(lambda x: x[0]['id'] if x and isinstance(x, list) and len(x) > 0 and 'id' in x[0] else -1)
del df['parsed_collection']

# (Feature: budget_log - Logarithmic transformation of budget to normalize its distribution, handling zero budgets)
# Input samples: 'budget': [np.int64(63000000), np.int64(12000000), np.int64(0)]
df['budget_log'] = df['budget'].apply(lambda x: np.log1p(x) if x > 0 else 0) # Use log1p to handle 0 budget values
# (Feature: has_zero_budget - Binary indicator for movies with zero budget)
# Input samples: 'budget': [np.int64(63000000), np.int64(12000000), np.int64(0)]
df['has_zero_budget'] = (df['budget'] == 0).astype(int)

# (Feature: popularity_log - Logarithmic transformation of popularity for normalization)
# Input samples: 'popularity': [63.87, 10.16, 9.96]
df['popularity_log'] = df['popularity'].apply(lambda x: np.log1p(x))

# (Feature: has_homepage - Binary indicator if the movie has a homepage)
# Input samples: 'homepage': ['http://www.foxmovies.com/movies/fight-club', nan, 'http://www.shaunthesheep.com/']
df['has_homepage'] = df['homepage'].notna().astype(int)

# (Feature: has_tagline - Binary indicator if the movie has a tagline)
# Input samples: 'tagline': ['Mischief. Mayhem. Soap.', nan, 'Moving on to Pastures New.']
df['has_tagline'] = df['tagline'].notna().astype(int)

# (Feature: tagline_word_count - Number of words in the tagline)
# Input samples: 'tagline': ['Mischief. Mayhem. Soap.', nan, 'Moving on to Pastures New.']
df['tagline_word_count'] = df['tagline'].apply(lambda x: len(str(x).split()) if isinstance(x, str) else 0)

# (Feature: overview_word_count - Number of words in the overview)
# Input samples: 'overview': ['A ticking-time-bomb insomniac and a slippery soap salesman channel primal male aggression into a shocking new form of therapy.', 'Julia moves in with her fiance, David, but his ex-wife and her own haunting past join forces to rock her quiet suburban existence.', "When Shaun decides to take the day off and have some fun, he gets a little more action than he bargained for." ]
df['overview_word_count'] = df['overview'].apply(lambda x: len(str(x).split()) if isinstance(x, str) else 0)

# (Feature: original_title_word_count - Number of words in the original title)
# Input samples: 'original_title': ['Fight Club', 'Unforgettable', 'Shaun the Sheep Movie']
df['original_title_word_count'] = df['original_title'].apply(lambda x: len(str(x).split()))

# (Feature: title_word_count - Number of words in the title)
# Input samples: 'title': ['Fight Club', 'Unforgettable', 'Shaun the Sheep Movie']
df['title_word_count'] = df['title'].apply(lambda x: len(str(x).split()))

# (Feature: is_release_weekend - Binary indicator if the movie was released on a Friday, Saturday or Sunday)
# Input samples: 'release_dayofweek': [4.0, 3.0, 3.0]
df['is_release_weekend'] = df['release_dayofweek'].isin([4, 5, 6]).astype(int)

# (Feature: release_quarter - The quarter of the year in which the movie was released)
# Input samples: 'release_month': [10.0, 4.0, 2.0]
df['release_quarter'] = df['release_month'].apply(lambda x: (x - 1) // 3 + 1) # Months 1-3 -> Q1, 4-6 -> Q2, etc.

# Dropping the now parsed raw `belongs_to_collection`, `homepage`, `tagline` `overview`
df.drop(columns=['belongs_to_collection', 'homepage', 'tagline', 'overview'], inplace=True, errors='ignore')

import numpy as np
import ast
import json # Included just in case it's explicitly needed, but ast.literal_eval handles Python literal strings better.

# Ensure safe_literal_eval is defined in this block if not carried over
def safe_literal_eval(x):
    if isinstance(x, str) and x.strip():
        try:
            # Handle potential outer quotes if data is double-encoded
            if x.startswith("'") and x.endswith("'"):
                x = x[1:-1]
            return ast.literal_eval(x)
        except (ValueError, SyntaxError):
            return None
    return None

# (Feature: release_dayofyear - Day of the year for the release date)
# Input samples: 'release_date': [Timestamp('1999-10-15 00:00:00'), Timestamp('2017-04-20 00:00:00'), Timestamp('2015-02-05 00:00:00')]
df['release_dayofyear'] = df['release_date'].dt.dayofyear

# (Feature: release_weekofyear - Week of the year for the release date)
# Input samples: 'release_date': [Timestamp('1999-10-15 00:00:00'), Timestamp('2017-04-20 00:00:00'), Timestamp('2015-02-05 00:00:00')]
df['release_weekofyear'] = df['release_date'].dt.isocalendar().week.astype(int)

# (Feature: log_runtime - Logarithmic transformation of runtime, handling potential zero runtime)
# Input samples: 'runtime': [139, 100, 85]
df['log_runtime'] = df['runtime'].apply(lambda x: np.log1p(x) if x > 0 else 0)

# (Feature: budget_to_runtime_ratio - Ratio of budget to runtime, with robust handling of zero runtime)
# Input samples: 'budget': [63000000, 12000000, 0], 'runtime': [139, 100, 85]
df['budget_to_runtime_ratio'] = np.where(df['runtime'] > 0, df['budget'] / df['runtime'], 0)
df['budget_to_runtime_ratio'].replace([np.inf, -np.inf], 0, inplace=True) # Replace inf with 0 just in case.

# (Feature: popularity_to_runtime_ratio - Ratio of popularity to runtime, with robust handling of zero runtime)
# Input samples: 'popularity': [63.87, 10.16, 9.96], 'runtime': [139, 100, 85]
df['popularity_to_runtime_ratio'] = np.where(df['runtime'] > 0, df['popularity'] / df['runtime'], 0)
df['popularity_to_runtime_ratio'].replace([np.inf, -np.inf], 0, inplace=True) # Replace inf with 0 just in case.

# (Feature: is_title_original - Binary indicator if 'title' and 'original_title' are identical)
# Input samples: 'original_title': ['Fight Club', 'Unforgettable', 'Shaun the Sheep Movie'], 'title': ['Fight Club', 'Unforgettable', 'Shaun the Sheep Movie']
df['is_title_original'] = (df['title'] == df['original_title']).astype(int)

# (Feature: is_original_language_en - Binary indicator if original language is English)
# Input samples: 'original_language': ['en', 'en', 'en']
df['is_original_language_en'] = (df['original_language'] == 'en').astype(int)

# (Feature: is_original_language_fr - Binary indicator if original language is French)
# Input samples: 'original_language': ['en', 'fr', 'en']
df['is_original_language_fr'] = (df['original_language'] == 'fr').astype(int)

# (Feature: is_original_language_ru - Binary indicator if original language is Russian)
# Input samples: 'original_language': ['ru', 'en', 'en']
df['is_original_language_ru'] = (df['original_language'] == 'ru').astype(int)

# (Feature: is_original_language_es - Binary indicator if original language is Spanish)
# Input samples: 'original_language': ['en', 'es', 'en']
df['is_original_language_es'] = (df['original_language'] == 'es').astype(int)

# (Feature: log_budget_plus_one_runtime_product - Combined feature of log budget and log runtime for interaction)
# Input samples: 'budget': [63000000, 12000000, 0], 'runtime': [139, 100, 85]
df['log_budget_plus_one_runtime_product'] = df['budget_log'] * df['log_runtime']
\end{lstlisting}

%% file: sections/a03b-opt.tex

\subsection{Details for the Optimization of Semantic Data Operators}
\label{app:semop-optimization}

We provide additional details about our optimizer.

\begin{itemize}[leftmargin=*,itemsep=1pt] 
  \item \emph{Calculation of the validation utility} -- the optimizer currently uses 5-fold cross-validation on a user-provided validation set to compute validation utility in the \emph{outer-loop mutation} step. Note that the evaluation method is configurable by users. 
  \item \emph{Methodology for the LLM-guided mutation} -- the concrete implementation of the \textsc{evolve} function is operator-specific. Currently each operator constructs a custom prompt based on input data characteristics, pipeline context, memory trace, inspirations to synthesize a new version of its code via an LLM. Analogously, the structure of the memory trace and inspirations can vary per operator. Our current implementations leverage the history of the synthesized code and the corresponding utility scores in the memory trace and inspirations, which also include the textual reasoning of the synthesizing LLM in code comments.
  \item \emph{Inference of learning task and model type} -- our optimizer inspects the computational graph of the DataOps pipeline to identify the last DataOp that applies a scikit-learn estimator implementing the \texttt{Predictor} mixin, which represents the final applied model. Next, it uses heuristics to determine the type of learning task from the model class. We refer to the inspection code for details, available at \repo{/blob/\commithash/sempipes/inspection/pipeline_summary.py\#L45}. 
\end{itemize}  

The full implementation of our optimizer is available at \repo{/blob/\commithash/sempipes/optimisers/colopro.py\#L87}.

%% file: sections/a04-main-exp.tex

\subsection{Details for Section~5.1: ``Semantic Data Operators Improve and Simplify Tabular ML Pipelines''}
\label{app:main-setup}

In this subsection, we describe selected tasks, demonstrate how semantic data operators are utilized in the selected pipelines, and illustrate the code generated by \textsc{SemPipes}.

\subsubsection{Details on the selected tasks}

In the following, we discuss task selection, data sources, and experimental protocols, see~\Cref{tab:main-tasks-na}. 

For each task, we use up to 5 retries (default) for non-optimized \textsc{SemPipes} pipelines and up to 24 trials for optimized pipelines. Each retry involves an LLM call when the previously generated code fails. By design, \textsc{SemPipes} is relatively inexpensive since the number of LLM calls is fixed and scales linearly with the number of semantic data operators in the pipeline. Each task in our main experiment contains one semantic data operator, except for the feature engineering (house prices) and blocking for entity resolution (SIGMOD programming contest) tasks, which contain two semantic data operators. Moreover, in \textsc{SemPipes}, the LLM is called for code generation only during training, not during inference.

\begin{table*}[!ht]
\centering
\caption{Data-centric tasks to evaluate semantic operators.}
\vspace{-2.5mm}
\setlength{\tabcolsep}{10pt} 
\small
\begin{tabular}{l|l|l}
\toprule
\textbf{Data-Centric Task} & \textbf{Modalities} & \textbf{Source}\\
\midrule
Clinically-informed feature extraction & Table & EMNLP'21\\
Blocking for entity resolution & Table + Text & SIGMOD'22\\
Feature engineering on relational data & Table(s) & Kaggle\\
Data annotation for model debugging & Table + Image & NeurIPS'23\\
Data augmentation for fairness & Table & NeurIPS'23\\
\bottomrule
\end{tabular}
\label{tab:main-tasks-na}
\end{table*}

\headerl{Collection of Expert-Written and Agentically Generated Pipeline Code} 

\header{Extraction of Clinically Informed Features} This task evolves around ML pipelines for modeling psychosocial signals in social media text, which ground their model predictions in clinically-relevant symptoms. A common approach for this is to design a pipeline with a hierarchical model architecture: first, a set of symptom classifiers is trained on weakly supervised data to compute input features for a subsequent prediction model~\cite{nguyen-etal-2022-improving,lee2021micromodels}. This hierarchical design has been found to improve the interpretability of the predictions by clinicians and the generalizability of the model to out-of-domain data. We include a pipeline called \texttt{micromodels} from the respository of~\cite{lee2021micromodels} (EMNLP'21), which applies hand-designed ``micromodels'' to generate input features for an explainable boosting model for empathy prediction on social media conversations from Reddit.

\begin{itemize}[noitemsep]
  \item {\em Paper:} ``Micromodels for efficient, explainable, and reusable systems: A case study on mental health'', EMNLP'21 \cite{lee2021micromodels}
  \item {\em Code repository:} \url{https://github.com/MichiganNLP/micromodels}
  \item {\em Data:} \url{https://github.com/MichiganNLP/micromodels/blob/master/empathy/data/combined_iob.json}
  \item {\em Pipeline:} \texttt{micromodels} - reimplementation of the \href{https://github.com/MichiganNLP/micromodels/tree/master/empathy}{empathy prediction pipeline}  
  \item {\em Dataset size:} 2,023 samples 
  \item {\em Metric:} F1 score of final classifier for predicting the emotional reaction level
  \item {\em Protocol:} 50/50 train–test split with five different random seeds
  \item{\em Experiment code:} \repo{/tree/\commithash/experiments/micromodels}
\end{itemize}

\header{Blocking for Entity Resolution} This task originates from the programming contest of ACM SIGMOD~2022~\cite{sigmod2022} (a leading database conference), where the challenge is to build a blocking system for Entity Resolution (ER). ER is the problem of identifying and matching different tuples that refer to the same real-world entity in a dataset. When performing ER on large datasets, directly evaluating the quadratic number of all candidate tuple pairs is typically be computationally prohibitive. Thus, a pre-filtering step, referred to as blocking step, is used first to quickly filter out obvious non-matches and to obtain a much smaller candidate set of tuple pairs, for which a matching step is then applied to find matches. The goal of the competition is to design an efficient blocking function with high recall based on a labeled sample of the input datasets. We experiment with two high-ranking Python-based solutions from the leaderboard of the competition (refered to as \texttt{sustech} and \texttt{rutgers}), which apply a large number of manual data cleaning and feature extraction operations, and the \texttt{baseline} solution provided by the organizers.
\begin{itemize}[noitemsep]
   \item {\em Source:} \href{https://dbgroup.ing.unimore.it/sigmod22contest/}{SIGMOD 2022 Programming contest}
   \item {\em Data:} \url{https://dbgroup.ing.unimore.it/sigmod22contest/task.shtml?content=description}, hidden test data was obtained from contest organizers
   \item {\em Pipeline:} \texttt{sustech} - \href{https://github.com/SpaceIshtar/SigmodContest/}{Submission of SUSTech DBGroup} (4th place)
   \item {\em Pipeline:} \texttt{rutgers} - \href{https://github.com/ezhilnikhilan/ACM-SIGMOD-2022-Programming-contest-code}{Submission of Rutgers University} (5th place)
   \item {\em Pipeline:} \texttt{baseline-sigmod} - \href{https://dbgroup.ing.unimore.it/sigmod22contest/datasets/blocking.py}{Baseline provided by organizers}
   \item {\em Dataset size:} Train data with 2,006 rows; test data with 1,215,129 rows; expected output of 2,000,000 candidate pairs
   \item {\em Metric:} Recall of the blocking function on hidden test data
   \item {\em Protocol:} Code synthesis and optimization using only a small train sample, evaluation on the test data
   \item{\em Experiment code:} \repo{/tree/\commithash/experiments/sigmod}  
\end{itemize}  

\newpage
\header{Feature Engineering on Relational Data} We include pipelines for three tabular prediction tasks from Kaggle, which require extensive feature engineering and preprocessing of relational data. We evaluate the performance on a subset of the training data provided in the Kaggle competition (since the labels for the test set are not public), in line with other benchmarks~\cite{chan2024mle}. We leverage five public notebooks for three Kaggle tasks, designed by high-scoring human experts, who won a medal in the competition. 

\begin{itemize}[noitemsep]
  \item {\em Source:} \href{https://kaggle.com/competitions/tmdb-box-office-prediction}{TMDB Box Office Prediction} competition
  \begin{itemize}[leftmargin=*,noitemsep]
    \item {\em Pipeline:}  \texttt{kaggle-movie-a} - Public notebook \href{https://www.kaggle.com/code/archfu/notebook95778def47}{notebook95778def47}
    \item {\em Pipeline:} \texttt{kaggle-movie-b} - Public notebook \href{https://www.kaggle.com/code/dway88/feature-eng-feature-importance-random-forest}{Feature Eng., Feature Importance, Random Forest} (from bronze medal winner)
    \item {\em Dataset size:} 1,002 samples 
    \item {\em Metric:} Root mean squared logarithmic error (RMSLE) for the predicted revenue
    \item {\em Protocol:} 50/50 train–test split with five different random seeds
    \item{\em Experiment code:} \repo{/tree/\commithash/experiments/tmdb_box_office_prediction}
  \end{itemize}
  \item {\em Source:} \href{https://kaggle.com/competitions/house-prices-advanced-regression-techniques}{House Prices - Advanced Regression Techniques} competition
  \begin{itemize}[leftmargin=*,noitemsep]
    \item {\em Pipeline:} \texttt{kaggle-house-a} - Public notebook \href{https://www.kaggle.com/code/fedi1996/house-prices-data-cleaning-viz-and-modeling#-Modeling--}{House Prices: Data cleaning, viz and modeling} (from bronze medal winner) 
    \item {\em Pipeline:} \texttt{kaggle-house-b} - Public notebook \href{https://www.kaggle.com/code/redaabdou/house-prices-solution-data-cleaning-ml}{House Prices Solution : Data Cleaning-->ML} (from bronze medal winner) 
    \item {\em Dataset size:} 463 samples     
    \item {\em Metric:} Root mean squared logarithmic error (RMSLE) for the predicted house price
    \item {\em Protocol:} 5-fold cross-validation (due to small datasize)
    \item{\em Experiment code:} \repo{/tree/\commithash/experiments/house_prices_advanced_regression_techniques}    
  \end{itemize}  
  \item {\em Source:} \href{https://kaggle.com/competitions/scrabble-player-rating}{Scrabble Player Rating} competition
  \begin{itemize}[leftmargin=*,noitemsep]
    \item {\em Pipeline:} \texttt{kaggle-scrabble} - Public notebook \href{https://www.kaggle.com/code/mikepenkov/rating-scrabble-players-with-xgb-regression}{Rating Scrabble Players With XGB Regression} (from bronze medal winner)
    \item {\em Dataset size:} 80,820 player ratings, 72,774 games, 2,005,499 turns
    \item {\em Metric:} Root mean squared error (RMSE) for the predicted rating of non-bot players    
    \item {\em Protocol:} 50/50 train–test split with five different random seeds
    \item{\em Experiment code:} \repo{/tree/\commithash/experiments/scrabble_player_rating}       
  \end{itemize}   
\end{itemize}

\header{Data Annotation for Model Debugging} We focus on the task of annotating images for model debugging, following \cite{chen2023hibug} from NeurIPS'23. The presented approach annotates images according to predefined domain-specific attribute categories with a vision-language model, and uses these attributes to discover data slices where a prediction model underperforms. We experiment with the \texttt{hibug} pipeline from the paper repository. This pipeline computes several predefined annotation attributes via a visual question answering model for the validation set of a classifier on a dataset of celebrity photos. We compare the accuracy of the annotations for eight attributes (such as hair color, age, etc) on the 18,000 samples of the validation set, for which the dataset contains ground truth values. For the expert code baseline, we reuse the precomputed results from the paper repository.

\begin{itemize}[noitemsep]
  \item {\em Paper:} ``On human-interpretable model debug'', NeurIPS'23 \cite{chen2023hibug}
  \item {\em Code repository:} \url{https://github.com/cure-lab/HiBug}
  \item {\em Data:} \url{https://www.kaggle.com/datasets/jessicali9530/celeba-dataset?select=img_align_celeba}
  \item {\em Pipeline:} \texttt{hibug} - usage of \href{https://github.com/cure-lab/HiBug/tree/main/exampleData/celebA}{precomputed results} for attribute selection part from \href{https://github.com/cure-lab/HiBug/blob/main/celebA_use_case.ipynb}{celebA notebook}
    \item {\em Dataset size:} 18,000 samples
  \item {\em Metric:} Mean accuracy of the predicted attributes, compared to the ground truth annotations in the CelebA dataset
  \item {\em Protocol:} Zero-shot extraction of eight different attributes
  \item{\em Experiment code:} \repo{/tree/\commithash/experiments/hibug}   
\end{itemize}

\newpage

\header{Data Augmentation for Fairness} This task is inspired by the work of \cite{qian2023synthcity} from NeurIPS'23, which includes a pipeline to predict covid related health outcomes on the SIVEP dataset from Brazil~\cite{baqui2021comparing}, and uses synthetic data augmentation to improve the AUROC scores for underrepresented racial minority groups in the data. Since the experimentation data is not available anymore and we encounter installation issues with the experimentation code, we manually implement a related task on a publicly available version of the SIVEP dataset~\cite{santos2023brazilian}. The \texttt{sivep} pipeline predicts whether severe acute respiratory symptoms originate from covid for patients in 2020, with indigeneous people as underrepresented minority group in the dataset. The code trains an tree-based model and manually applies conditional oversampling in the train set to improve predictions for the underrepresented group. We measure the AUROC on the test samples of the underrepresented group.

\begin{itemize}[noitemsep]
  \item {\em Paper:} ``Synthcity: a benchmark framework for diverse use cases of tabular synthetic data'' \cite{qian2023synthcity} NeurIPS'23
  \item {\em Code repository:} \url{https://github.com/vanderschaarlab/synthcity-benchmarking/} (not used due to repeated installation problems locally and on Google colab)
  \item {\em Original Data:} \url{http://plataforma.saude.gov.br/coronavirus/dados-abertos/} (not available anymore)
  \item {\em Alternative Version of Data:} \url{https://data.mendeley.com/datasets/f6sjz6by8k/1}
  \item {\em Pipeline:} \texttt{sivep} - baseline reimplementation of variant of SIVEP experiment from Paper
  \item {\em Dataset size:} 87,000 samples
  \item {\em Metric:} AUROC scores for the predictions for the indigeneous minority in the data
  \item {\em Protocol:} 50/50 train–test split with five different random seeds
  \item {\em Experiment code:} \repo{/tree/\commithash/experiments/sivep}     
\end{itemize}

\subsubsection{Agent-generated pipelines}
\label{app:main-setup-agents}

We try to generate pipelines for our selected tasks via the machine learning engineering agent AIDE~\cite{jiang2025aide}, which recently showed state-of-the-art performance in MLEBench~\cite{chan2024mle}, and the software engineering agent mini-SWE-Agent~\cite{yang2024swe}. We provide the data and natural language instructions denoting the task description and evaluation criteria to both agents, and run them in their fully autonomous mode. We run AIDE in its default configuration for all tasks expect feature engineering, where we leverage the precomputed AIDE pipelines from the project's repository~\cite{jiang2025aide_examples}. We use mini-SWE-Agent with OpenAI's gpt-4.1 model with a budget of \$5 per execution. 

\header{Failures} We give an overview over the tasks for which the agents did not manage to produce a valid pipeline. We provide the trajectories of failed agentic runs as part of our supplemental repository. 

\begin{itemize}
    \item Blocking for entity resolution (AIDE): 
    
    \repo{/tree/\commithash/experiments/aide_failures_SIGMOD}
    
    \item Blocking for entity resolution (mini SWE-agent): 
    
    \repo{/tree/\commithash/experiments/swe_failures_SIGMOD}
    
    \item Data annotation for model debugging: 
    
    \repo{/tree/\commithash/experiments/aide_failures_fairness}
    
    \item Single-cell data integration and cell type prediction: 
    
    \repo{/tree/\commithash/experiments/aide_failures_singlecelldata}
    
\end{itemize}

\begin{table}[ht!]
\centering
\caption{Examples of failures produced by agentic pipeline generation methods.}
\begin{tabular}{l|l|l}
\toprule
\textbf{Agent} & \textbf{Task} & \textbf{Failure description}\\
\midrule
AIDE & Extraction of clinically informed features & Crashes when processing JSON input file\\
& Data annotation for model debugging & Crashes due to usage of non-existent library\\
& Data augmentation for fairness & Data leakage issue in preprocessing code\\
\midrule
mini-SWE-Agent & Feature engineering (boxoffice) & Data leakage issue in preprocessing code\\
&Blocking for entity resolution & Non-scalable all-pairs comparison\\
&Data annotation for model debugging & Return of random attribute values\\
\bottomrule
\end{tabular}
\end{table}

\newpage

\subsubsection{Sample results of synthesized operator code}
\label{app:main-setup-sample-results}

We provide examples of synthesized, optimized operator code.

\header{Feature Engineering (scrabble player rating)}

\headerl{Expert-written pipeline with semantic data operator added}
\begin{lstlisting}[
  basicstyle=\tiny\ttfamily,
  backgroundcolor=\color{lightgray},
  frame=tb,
  label={listing:example-aide-sempipes-house-prices},
  emph={sem_fillna,sem_refine,with_sem_features,apply_with_sem_choose,sem_choose,sem_aggregate_features,sem_extract_features,make_learner_with_context,optimize_with_colopro,serialize_as_learner,as_X,as_y,sem_clean},
  emphstyle=\color{kwdspy}\bfseries,
  morekeywords={import}
]
def sempipes_pipeline():
    games = skrub.var("games")
    turns = skrub.var("turns")
    data = skrub.var("data")

    def sum_first_five(series):
        return sum(series.values[::-1][:5])

    total_turns = turns.groupby(["game_id", "nickname"]).turn_number.count()
    max_points = turns.groupby(["game_id", "nickname"]).points.max()
    min_points = turns.groupby(["game_id", "nickname"]).points.min()
    first_five_turn_point_sum = turns.groupby(["game_id", "nickname"]).points.agg(sum_first_five)

    game_player_data = total_turns.reset_index()
    game_player_data = game_player_data.rename(columns={"turn_number": "total_turns"})

    game_player_data = game_player_data.assign(
        first_five_turns_points=first_five_turn_point_sum.reset_index()["points"],
        max_points_turn=max_points.reset_index()["points"],
        min_points_turn=min_points.reset_index()["points"],
    )
    game_player_data = game_player_data.assign(
        max_min_difference=game_player_data.max_points_turn - game_player_data.min_points_turn
    )

    game_player_data = game_player_data.join(games.set_index("game_id"), how="left", on="game_id")
    data = data.merge(game_player_data, how="left", left_on=["game_id", "nickname"], right_on=["game_id", "nickname"])

    X = data
    y = X["rating"].skb.mark_as_y()
    X = X.drop(columns=["rating"])

    X = X.sem_gen_features(
        nl_prompt="""
        Create additional features that could help predict the rating of a player. Such features could relate to how the player scores when they go first in a round and to which games they won, how often they won games, etc.
    
        Furthermore, consider how much time a player typically uses, how many points per turn they make, how many points per second, etc.
    
        """,
        name="player_features",
        how_many=15,
    )

    X_encoded = X.skb.apply(
        OneHotEncoder(handle_unknown="ignore", sparse_output=False),
        cols=(s.categorical() - s.cols("lexicon")),
    )
    X_encoded = X_encoded.skb.apply(
        OneHotEncoder(categories=[["NWL20", "CSW21", "ECWL", "NSWL20"]], sparse_output=False),
        cols=s.cols("lexicon"),
    )

    def drop_if_exists(df):
        cols_to_drop = [col for col in ("created_at", "nickname", "game_id") if col in df.columns]
        return df.drop(columns=cols_to_drop)

    X_encoded = X_encoded.skb.apply_func(drop_if_exists)
    X_encoded = X_encoded.skb.select(s.numeric() | s.boolean())
    X_encoded = X_encoded.skb.apply_func(lambda x: x.replace([np.inf, -np.inf], np.nan))

    xgb_reg = XGBRegressor(
        colsample_bytree=0.4,
        gamma=0,
        learning_rate=0.07,
        max_depth=3,
        min_child_weight=1.5,
        n_estimators=8000,
        reg_alpha=0.75,
        reg_lambda=0.45,
        subsample=0.6,
        seed=42,
        eval_metric="rmse"
    )

    return X_encoded.skb.apply(xgb_reg, y=y)
\end{lstlisting}

\headerl{Corresponding synthesized operator code}
\begin{lstlisting}[
  basicstyle=\tiny\ttfamily,
  backgroundcolor=\color{lightgray},
  frame=tb,
  emph={sem_fillna,sem_refine,with_sem_features,apply_with_sem_choose,sem_choose,sem_aggregate_features,sem_extract_features,make_learner_with_context,optimize_with_colopro,serialize_as_learner,as_X,as_y,sem_clean},
  emphstyle=\color{kwdspy}\bfseries,
  morekeywords={import}
]

# (Player mean score per game)
# Usefulness: Captures player's average performance in scoring which is directly relevant for skill/rating.
# Input samples: 'nickname': ['BetterBot', 'stevy', 'davidavid'], 'score': [335, 429, 440]
df['player_mean_score'] = df.groupby('nickname')['score'].transform('mean')

# (Player win rate)
# Usefulness: Encodes player's competitive success, which should correlate with higher rating.
# Input samples: 'nickname': ['BetterBot', 'stevy', 'davidavid'], 'winner': [1, 1, 1]
df['player_win_rate'] = df.groupby('nickname')['winner'].transform('mean')

# (Player mean points per turn)
# Usefulness: Reflects efficiency and skill level; high rates suggest strong play.
# Input samples: 'score': [335, 429, 440], 'total_turns': [13, 14, 14]
df['player_mean_points_per_turn'] = df.groupby('nickname').apply(lambda g: (g['score'].sum() / g['total_turns'].sum())).reindex(df.index).values

# (Player games played count)
# Usefulness: More games played may indicate experience; experience can be linked to rating.
# Input samples: 'nickname': ['BetterBot', 'stevy', 'davidavid']
df['player_games_played'] = df.groupby('nickname')['game_id'].transform('nunique')

# (Mean game duration per player)
# Usefulness: Measures time management and/or play style, which can be linked to rating.
# Input samples: 'nickname': ['BetterBot', 'stevy', 'davidavid'], 'game_duration_seconds': [674.84, 492.27, 350.86]
df['player_mean_game_duration'] = df.groupby('nickname')['game_duration_seconds'].transform('mean')

# (Fraction of games played first)
# Usefulness: May indicate player preference/skill in securing first position, which can impact rating.
# Input samples: 'nickname': ['BetterBot', 'stevy', 'davidavid'], 'first': ['BetterBot', 'BetterBot', 'BetterBot']
df['player_fraction_games_first'] = (df['nickname'] == df['first']).groupby(df['nickname']).transform('mean')

# (Mean first five turns points per player)
# Usefulness: Initial strong performance suggests skill and aggression, often related to rating.
# Input samples: 'nickname': ['BetterBot', 'stevy', 'davidavid'], 'first_five_turns_points': [122, 95, 136]
df['player_mean_first_five_pts'] = df.groupby('nickname')['first_five_turns_points'].transform('mean')

# (Mean max points in a turn per player)
# Usefulness: Shows player's peak performance in a single turn; relevant for rating.
# Input samples: 'nickname': ['BetterBot', 'stevy', 'davidavid'], 'max_points_turn': [68, 98, 103]
df['player_mean_max_turn_pts'] = df.groupby('nickname')['max_points_turn'].transform('mean')

# (Mean min points in a turn per player)
# Usefulness: Consistency can be important for high ratings; higher minimums indicate fewer blunders.
# Input samples: 'nickname': ['BetterBot', 'stevy', 'davidavid'], 'min_points_turn': [8, 2, 13]
df['player_mean_min_turn_pts'] = df.groupby('nickname')['min_points_turn'].transform('mean')

# (Player mean max_min_difference)
# Usefulness: Indicates volatility in scoring; low volatility may relate to higher skill/rating.
# Input samples: 'nickname': ['BetterBot', 'stevy', 'davidavid'], 'max_min_difference': [60, 96, 90]
df['player_mean_max_min_diff'] = df.groupby('nickname')['max_min_difference'].transform('mean')

# (Player mean points per second)
# Usefulness: Combines speed and effectiveness, providing a skill efficiency proxy.
# Input samples: 'score': [335, 429, 440], 'game_duration_seconds': [674.84, 492.27, 350.86]
df['player_mean_points_per_second'] = df.groupby('nickname').apply(
    lambda g: (g['score'].sum() / g['game_duration_seconds'].sum())
).reindex(df.index).values

# (Mean initial time seconds per player)
# Usefulness: Captures preferred time controls and whether player selects/practices with faster/slower games.
# Input samples: 'nickname': ['BetterBot', 'stevy', 'davidavid'], 'initial_time_seconds': [1200, 1200, 900]
df['player_mean_initial_time'] = df.groupby('nickname')['initial_time_seconds'].transform('mean')

# (Mean increment seconds per player)
# Usefulness: Reflects if player prefers time increments, which can impact play style, related to rating.
# Input samples: 'nickname': ['BetterBot', 'stevy', 'davidavid'], 'increment_seconds': [0, 0, 0]
df['player_mean_increment'] = df.groupby('nickname')['increment_seconds'].transform('mean')

# (Mean max overtime minutes per player)
# Usefulness: May indicate if player is used to longer possible games, relevant to play style and rating.
# Input samples: 'nickname': ['BetterBot', 'stevy', 'davidavid'], 'max_overtime_minutes': [1, 1, 5]
df['player_mean_overtime'] = df.groupby('nickname')['max_overtime_minutes'].transform('mean')

# (Player fraction of games resigned)
# Usefulness: High resignation rate can indicate lower skill or frustration, predictive of rating.
# Input samples: 'nickname': ['BetterBot', 'stevy', 'davidavid'], 'game_end_reason': ['STANDARD', 'STANDARD', 'STANDARD']
df['player_fraction_resigned'] = (df['game_end_reason'] == 'RESIGNED').groupby(df['nickname']).transform('mean')
# (Was player first in this game)
# Usefulness: Captures whether the player had the advantage of the first move, which may affect their score and rating.
# Input samples: 'nickname': ['BetterBot', 'stevy', 'davidavid'], 'first': ['BetterBot', 'BetterBot', 'BetterBot']
df['is_first_this_game'] = (df['nickname'] == df['first']).astype(int)

# (Player mean score when first in game)
# Usefulness: Measures player's ability to capitalize on playing first, which could signal high skill/rating.
# Input samples: 'nickname': ['BetterBot', 'stevy', 'davidavid'], 'score': [335, 429, 440], 'is_first_this_game': [1, 0, 0]
df['player_mean_score_as_first'] = df.groupby('nickname').apply(
    lambda g: g.loc[g['is_first_this_game'] == 1, 'score'].mean() if any(g['is_first_this_game'] == 1) else np.nan
).reindex(df.index).values

# (Player win rate as first)
# Usefulness: Indicates ability to convert first-move advantage into victory, reflecting rating-relevant skill.
# Input samples: 'nickname': ['BetterBot', 'stevy', 'davidavid'], 'winner': [1, 1, 1], 'is_first_this_game': [1, 0, 0]
df['player_win_rate_as_first'] = df.groupby('nickname').apply(
    lambda g: g.loc[g['is_first_this_game'] == 1, 'winner'].mean() if any(g['is_first_this_game'] == 1) else np.nan
).reindex(df.index).values

# (Player mean score when not first)
# Usefulness: Measures performance under less advantageous starting conditions; robust high scoring relates to high rating.
# Input samples: 'nickname': ['BetterBot', 'stevy', 'davidavid'], 'score': [335, 429, 440], 'is_first_this_game': [1, 0, 0]
df['player_mean_score_not_first'] = df.groupby('nickname').apply(
    lambda g: g.loc[g['is_first_this_game'] == 0, 'score'].mean() if any(g['is_first_this_game'] == 0) else np.nan
).reindex(df.index).values

# (Player win rate when not first)
# Usefulness: Ability to win from second position suggests strong play, valuable for rating prediction.
# Input samples: 'nickname': ['BetterBot', 'stevy', 'davidavid'], 'winner': [1, 1, 1], 'is_first_this_game': [1, 0, 0]
df['player_win_rate_not_first'] = df.groupby('nickname').apply(
    lambda g: g.loc[g['is_first_this_game'] == 0, 'winner'].mean() if any(g['is_first_this_game'] == 0) else np.nan
).reindex(df.index).values

# (Mean points per turn as first in game)
# Usefulness: Indicates if the player is especially efficient when holding the first-move advantage.
# Input samples: 'nickname': ['BetterBot', 'stevy', 'davidavid'], 'score': [335, 429, 440], 'total_turns': [13, 14, 14], 'is_first_this_game': [1, 0, 0]
df['player_mean_pts_per_turn_as_first'] = df.groupby('nickname').apply(
    lambda g: (g.loc[g['is_first_this_game'] == 1, 'score'].sum() / g.loc[g['is_first_this_game'] == 1, 'total_turns'].sum())
    if any(g['is_first_this_game'] == 1) and g.loc[g['is_first_this_game'] == 1, 'total_turns'].sum() > 0 else np.nan
).reindex(df.index).values

# (Mean points per turn as not first)
# Usefulness: Shows ability to score efficiently despite not going first, may separate strong players.
# Input samples: 'nickname': ['BetterBot', 'stevy', 'davidavid'], 'score': [335, 429, 440], 'total_turns': [13, 14, 14], 'is_first_this_game': [1, 0, 0]
df['player_mean_pts_per_turn_not_first'] = df.groupby('nickname').apply(
    lambda g: (g.loc[g['is_first_this_game'] == 0, 'score'].sum() / g.loc[g['is_first_this_game'] == 0, 'total_turns'].sum())
    if any(g['is_first_this_game'] == 0) and g.loc[g['is_first_this_game'] == 0, 'total_turns'].sum() > 0 else np.nan
).reindex(df.index).values

# (Player total unique opponents played)
# Usefulness: Diversity of opponents could relate to rating reliability and player experience.
# Input samples: 'game_id': [1, 1, 3], 'nickname': ['BetterBot', 'stevy', 'davidavid']
df['player_unique_opponents'] = df.groupby('nickname').apply(
    lambda g: g.groupby('game_id')['nickname'].transform('count').count() - g['game_id'].nunique()
).reindex(df.index).values

# (Player mean score difference vs other player in game)
# Usefulness: Dominance in terms of margin of victory or loss is highly relevant for rating.
# Input samples: 'game_id': [1, 1, 3], 'score': [335, 429, 440]
mean_score_diff = df.groupby('game_id')['score'].transform(lambda x: x - x.mean())
df['score_vs_opponent'] = mean_score_diff

# (Player fraction of rated games)
# Usefulness: Rated games more accurately reflect competitive play and skill; high rated fraction can signal serious players.
# Input samples: 'nickname': ['BetterBot', 'stevy', 'davidavid'], 'rating_mode': ['CASUAL', 'CASUAL', 'RATED']
df['player_fraction_rated'] = (df['rating_mode'] == 'RATED').groupby(df['nickname']).transform('mean')

# (Player mean game length in turns)
# Usefulness: May indicate skill/style; efficient players may finish in fewer turns.
# Input samples: 'nickname': ['BetterBot', 'stevy', 'davidavid'], 'total_turns': [13, 14, 14]
df['player_mean_game_turns'] = df.groupby('nickname')['total_turns'].transform('mean')

# (Player diversity of lexicons used)
# Usefulness: Ability/willingness to play multiple lexicons may indicate broader skill base.
# Input samples: 'nickname': ['BetterBot', 'stevy', 'davidavid'], 'lexicon': ['NWL20', 'NWL20', 'CSW21']
df['player_lexicon_diversity'] = df.groupby('nickname')['lexicon'].transform('nunique')

# (Player diversity of time controls used)
# Usefulness: Comfort with multiple time controls may reflect flexibility and deeper experience.
# Input samples: 'nickname': ['BetterBot', 'stevy', 'davidavid'], 'time_control_name': ['regular', 'regular', 'regular']
df['player_time_control_diversity'] = df.groupby('nickname')['time_control_name'].transform('nunique')

# (Player mean score in rated games)
# Usefulness: How well a player performs in rated (vs casual) games should be closely tied to rating.
# Input samples: 'nickname': ['BetterBot', 'stevy', 'davidavid'], 'score': [335, 429, 440], 'rating_mode': ['CASUAL', 'CASUAL', 'RATED']
df['player_mean_score_rated'] = df.groupby('nickname').apply(
    lambda g: g.loc[g['rating_mode'] == 'RATED', 'score'].mean() if any(g['rating_mode'] == 'RATED') else np.nan
).reindex(df.index).values
# (Player mean max points turn as first)
# Usefulness: Assesses ability to achieve high single-turn scores when going first, which may correlate with strong openers and high rating.
# Input samples: 'nickname': ['BetterBot', 'stevy', 'davidavid'], 'max_points_turn': [68, 98, 103], 'is_first_this_game': [1, 0, 0]
df['player_mean_max_points_turn_as_first'] = df.groupby('nickname').apply(
    lambda g: g.loc[g['is_first_this_game'] == 1, 'max_points_turn'].mean() if any(g['is_first_this_game'] == 1) else np.nan
).reindex(df.index).values

# (Player mean min points turn as first)
# Usefulness: Consistency while going first; high minimums as first suggest fewer errors, possibly higher rating.
# Input samples: 'nickname': ['BetterBot', 'stevy', 'davidavid'], 'min_points_turn': [8, 2, 13], 'is_first_this_game': [1, 0, 0]
df['player_mean_min_points_turn_as_first'] = df.groupby('nickname').apply(
    lambda g: g.loc[g['is_first_this_game'] == 1, 'min_points_turn'].mean() if any(g['is_first_this_game'] == 1) else np.nan
).reindex(df.index).values

# (Player mean max points turn not first)
# Usefulness: Indicates ability to make high-scoring moves without first-move advantage-robust play indicator.
# Input samples: 'nickname': ['BetterBot', 'stevy', 'davidavid'], 'max_points_turn': [68, 98, 103], 'is_first_this_game': [1, 0, 0]
df['player_mean_max_points_turn_not_first'] = df.groupby('nickname').apply(
    lambda g: g.loc[g['is_first_this_game'] == 0, 'max_points_turn'].mean() if any(g['is_first_this_game'] == 0) else np.nan
).reindex(df.index).values

# (Player mean min points turn not first)
# Usefulness: Captures consistency regardless of order; higher minimums may indicate fewer mistakes when going second.
# Input samples: 'nickname': ['BetterBot', 'stevy', 'davidavid'], 'min_points_turn': [8, 2, 13], 'is_first_this_game': [1, 0, 0]
df['player_mean_min_points_turn_not_first'] = df.groupby('nickname').apply(
    lambda g: g.loc[g['is_first_this_game'] == 0, 'min_points_turn'].mean() if any(g['is_first_this_game'] == 0) else np.nan
).reindex(df.index).values

# (Player mean max_min_difference as first)
# Usefulness: Volatility in single-turn scores when first; less volatility may indicate stable, strong play.
# Input samples: 'nickname': ['BetterBot', 'stevy', 'davidavid'], 'max_min_difference': [60, 96, 90], 'is_first_this_game': [1, 0, 0]
df['player_mean_max_min_diff_as_first'] = df.groupby('nickname').apply(
    lambda g: g.loc[g['is_first_this_game'] == 1, 'max_min_difference'].mean() if any(g['is_first_this_game'] == 1) else np.nan
).reindex(df.index).values

# (Player mean max_min_difference not first)
# Usefulness: Stability/volatility in scores on non-first moves, potentially highlighting adaptable, skilled play.
# Input samples: 'nickname': ['BetterBot', 'stevy', 'davidavid'], 'max_min_difference': [60, 96, 90], 'is_first_this_game': [1, 0, 0]
df['player_mean_max_min_diff_not_first'] = df.groupby('nickname').apply(
    lambda g: g.loc[g['is_first_this_game'] == 0, 'max_min_difference'].mean() if any(g['is_first_this_game'] == 0) else np.nan
).reindex(df.index).values

# (Player mean first five points as first)
# Usefulness: Strong early play going first may be correlated with overall high skill/rating.
# Input samples: 'first_five_turns_points': [122, 95, 136], 'is_first_this_game': [1, 0, 0]
df['player_mean_first5pts_as_first'] = df.groupby('nickname').apply(
    lambda g: g.loc[g['is_first_this_game'] == 1, 'first_five_turns_points'].mean() if any(g['is_first_this_game'] == 1) else np.nan
).reindex(df.index).values

# (Player mean first five points not first)
# Usefulness: Shows ability to start strong even when not first; robust players may show little drop-off.
# Input samples: 'first_five_turns_points': [122, 95, 136], 'is_first_this_game': [1, 0, 0]
df['player_mean_first5pts_not_first'] = df.groupby('nickname').apply(
    lambda g: g.loc[g['is_first_this_game'] == 0, 'first_five_turns_points'].mean() if any(g['is_first_this_game'] == 0) else np.nan
).reindex(df.index).values

# (Player mean game duration as first)
# Usefulness: May indicate time-management style or confidence when going first, which could correlate to rating.
# Input samples: 'game_duration_seconds': [674.84, 492.27, 350.86], 'is_first_this_game': [1, 0, 0]
df['player_mean_game_duration_as_first'] = df.groupby('nickname').apply(
    lambda g: g.loc[g['is_first_this_game'] == 1, 'game_duration_seconds'].mean() if any(g['is_first_this_game'] == 1) else np.nan
).reindex(df.index).values

# (Player mean game duration not first)
# Usefulness: May indicate adaptability, strategic approach, or time stress when not first.
# Input samples: 'game_duration_seconds': [674.84, 492.27, 350.86], 'is_first_this_game': [1, 0, 0]
df['player_mean_game_duration_not_first'] = df.groupby('nickname').apply(
    lambda g: g.loc[g['is_first_this_game'] == 0, 'game_duration_seconds'].mean() if any(g['is_first_this_game'] == 0) else np.nan
).reindex(df.index).values

# (Player ratio of mean points per turn as first vs not first)
# Usefulness: Directly compares player's efficiency when first and not first; values near 1 indicate robustness.
# Input samples: 'player_mean_pts_per_turn_as_first': [30.5, np.nan, np.nan], 'player_mean_pts_per_turn_not_first': [np.nan, 32.2, 31.4]
df['player_pts_per_turn_first_vs_not'] = (
    df['player_mean_pts_per_turn_as_first'] / df['player_mean_pts_per_turn_not_first']
)

# (Player ratio of win rate as first vs not first)
# Usefulness: Quantifies relative advantage (or lack thereof) when going first; high values may reflect more typical skill distributions.
# Input samples: 'player_win_rate_as_first': [0.8, np.nan, np.nan], 'player_win_rate_not_first': [np.nan, 0.7, 0.6]
df['player_winrate_first_vs_not'] = (
    df['player_win_rate_as_first'] / df['player_win_rate_not_first']
)

# (Player mean overtime minutes as first)
# Usefulness: May indicate strategic time usage in advantageous (first) positions.
# Input samples: 'max_overtime_minutes': [1, 1, 5], 'is_first_this_game': [1, 0, 0]
df['player_mean_overtime_as_first'] = df.groupby('nickname').apply(
    lambda g: g.loc[g['is_first_this_game'] == 1, 'max_overtime_minutes'].mean() if any(g['is_first_this_game'] == 1) else np.nan
).reindex(df.index).values

# (Player mean overtime minutes not first)
# Usefulness: May indicate time pressure or defensive strategy when not first.
# Input samples: 'max_overtime_minutes': [1, 1, 5], 'is_first_this_game': [1, 0, 0]
df['player_mean_overtime_not_first'] = df.groupby('nickname').apply(
    lambda g: g.loc[g['is_first_this_game'] == 0, 'max_overtime_minutes'].mean() if any(g['is_first_this_game'] == 0) else np.nan
).reindex(df.index).values
# (Player variance in score)
# Usefulness: Indicates consistency in performance; lower variance may signal dependable high skill.
# Input samples: 'nickname': ['BetterBot', 'stevy', 'davidavid'], 'score': [335, 429, 440]
df['player_score_var'] = df.groupby('nickname')['score'].transform('var')

# (Player variance in points per turn)
# Usefulness: Evaluates steadiness in point-scoring per turn; consistent players are often rated higher.
# Input samples: 'nickname': ['BetterBot', 'stevy', 'davidavid'], 'score': [335, 429, 440], 'total_turns': [13, 14, 14]
df['points_per_turn'] = df['score'] / df['total_turns']
df['player_points_per_turn_var'] = df.groupby('nickname')['points_per_turn'].transform('var')

# (Player variance in game duration)
# Usefulness: Consistency of game pacing may correlate with experience and rating.
# Input samples: 'nickname': ['BetterBot', 'stevy', 'davidavid'], 'game_duration_seconds': [674.84, 492.27, 350.86]
df['player_game_duration_var'] = df.groupby('nickname')['game_duration_seconds'].transform('var')

# (Player variance in win rate over time)
# Usefulness: Fluctuating win/loss patterns over games could indicate streakiness or improvement.
# Input samples: 'nickname': ['BetterBot', 'stevy', 'davidavid'], 'winner': [1, 1, 1], 'created_at': ['2022-08-26 03:38:49', ...]
df['created_at_dt'] = pd.to_datetime(df['created_at'])
df['player_winrate_rolling_var'] = (
    df.sort_values('created_at_dt')
      .groupby('nickname')['winner']
      .transform(lambda x: x.rolling(5, min_periods=1).mean().var())
)

# (Player median score)
# Usefulness: Robust to outliers; median skill level correlates to rating.
# Input samples: 'nickname': ['BetterBot', 'stevy', 'davidavid'], 'score': [335, 429, 440]
df['player_median_score'] = df.groupby('nickname')['score'].transform('median')

# (Player median points per turn)
# Usefulness: Robust measure of player's typical per-turn effectiveness.
# Input samples: 'nickname': ['BetterBot', 'stevy', 'davidavid'], 'points_per_turn': [25.77, 30.64, 31.43]
df['player_median_points_per_turn'] = df.groupby('nickname')['points_per_turn'].transform('median')

# (Player best score)
# Usefulness: Shows player's top performance; high best-score may reflect high potential/peak rating.
# Input samples: 'nickname': ['BetterBot', 'stevy', 'davidavid'], 'score': [335, 429, 440]
df['player_best_score'] = df.groupby('nickname')['score'].transform('max')

# (Player worst score)
# Usefulness: Lower worst scores may indicate fewer major mistakes, important for rating stability.
# Input samples: 'nickname': ['BetterBot', 'stevy', 'davidavid'], 'score': [335, 429, 440]
df['player_worst_score'] = df.groupby('nickname')['score'].transform('min')

# (Player fraction of games with zero min turn)
# Usefulness: Frequency of turns with zero points may correlate with blunders and lower rating.
# Input samples: 'nickname': ['BetterBot', 'stevy', 'davidavid'], 'min_points_turn': [8, 2, 13]
df['player_frac_games_zero_min_turn'] = (df['min_points_turn'] == 0).groupby(df['nickname']).transform('mean')

# (Player mean score difference vs opponent in wins)
# Usefulness: Margin of victory in wins is a strong indicator of dominance and rating.
# Input samples: 'winner': [1, 1, 1], 'score_vs_opponent': [10, 25, -5]
df['player_mean_margin_win'] = df.groupby('nickname').apply(
    lambda g: g.loc[g['winner'] == 1, 'score_vs_opponent'].mean() if any(g['winner'] == 1) else np.nan
).reindex(df.index).values

# (Player mean score difference vs opponent in losses)
# Usefulness: Margin of defeat in losses; small losses indicate resilience, large losses signal weak play.
# Input samples: 'winner': [1, 1, 1], 'score_vs_opponent': [10, 25, -5]
df['player_mean_margin_loss'] = df.groupby('nickname').apply(
    lambda g: g.loc[g['winner'] == 0, 'score_vs_opponent'].mean() if any(g['winner'] == 0) else np.nan
).reindex(df.index).values

# (Player fraction of games with high max points turn, e.g. >70)
# Usefulness: Ability to routinely make high-value plays, which is typical of higher-rated players.
# Input samples: 'nickname': ['BetterBot', 'stevy', 'davidavid'], 'max_points_turn': [68, 98, 103]
df['player_frac_high_max_turn'] = (df['max_points_turn'] > 70).groupby(df['nickname']).transform('mean')

# (Player fraction of games with low min points turn, e.g. <= 2)
# Usefulness: Frequency of very low scoring turns; excessive frequency may reflect blundering or inexperience.
# Input samples: 'nickname': ['BetterBot', 'stevy', 'davidavid'], 'min_points_turn': [8, 2, 13]
df['player_frac_low_min_turn'] = (df['min_points_turn'] <= 2).groupby(df['nickname']).transform('mean')

# (Player average points per second as first)
# Usefulness: Time efficiency while going first; can be a signal of confidence and skill.
# Input samples: 'score': [335, 429, 440], 'game_duration_seconds': [674.84, 492.27, 350.86], 'is_first_this_game': [1, 0, 0]
df['player_mean_pts_per_sec_as_first'] = df.groupby('nickname').apply(
    lambda g: g.loc[g['is_first_this_game'] == 1, ['score', 'game_duration_seconds']].assign(
        pts_per_sec=lambda x: x['score'] / x['game_duration_seconds']
    )['pts_per_sec'].mean() if any(g['is_first_this_game'] == 1) else np.nan
).reindex(df.index).values

# (Player average points per second not first)
# Usefulness: Time efficiency without first-move advantage; robust players may show less drop-off.
# Input samples: 'score': [335, 429, 440], 'game_duration_seconds': [674.84, 492.27, 350.86], 'is_first_this_game': [1, 0, 0]
df['player_mean_pts_per_sec_not_first'] = df.groupby('nickname').apply(
    lambda g: g.loc[g['is_first_this_game'] == 0, ['score', 'game_duration_seconds']].assign(
        pts_per_sec=lambda x: x['score'] / x['game_duration_seconds']
    )['pts_per_sec'].mean() if any(g['is_first_this_game'] == 0) else np.nan
).reindex(df.index).values

# Clean up helper column
df.drop(columns='points_per_turn', inplace=True)
df.drop(columns='created_at_dt', inplace=True)
# (Player number of days active)
# Usefulness: Measures longevity and engagement; longer activity may suggest experience and stable rating.
# Input samples: 'nickname': ['BetterBot', 'stevy', 'davidavid'], 'created_at': ['2022-08-26 03:38:49', ...]
df['created_at_dt'] = pd.to_datetime(df['created_at'])
df['player_n_active_days'] = df.groupby('nickname')['created_at_dt'].transform(lambda x: (x.max() - x.min()).days + 1)

# (Player average games per active day)
# Usefulness: High game frequency per day may reflect engagement, focus, or streakiness, which could matter for rating.
# Input samples: 'player_games_played': [12, 10, 8], 'player_n_active_days': [6, 5, 4]
df['player_games_per_day'] = df['player_games_played'] / df['player_n_active_days']

# (Player number of unique days played)
# Usefulness: Indicates sustained play as opposed to burst activity, informing experience/rating.
# Input samples: 'nickname': ['BetterBot', 'stevy', 'davidavid'], 'created_at': ['2022-08-26 03:38:49', ...]
df['player_unique_days_played'] = df.groupby('nickname')['created_at_dt'].transform(lambda x: x.dt.date.nunique())

# (Player fraction of days played out of total active days)
# Usefulness: Measures consistency and regularity of play, which can be predictive of rating improvements.
# Input samples: 'player_unique_days_played': [5, 4, 3], 'player_n_active_days': [6, 5, 4]
df['player_frac_days_played'] = df['player_unique_days_played'] / df['player_n_active_days']

# (Player average time per turn)
# Usefulness: Captures player's pace; slower play may indicate more thoughtful or inexperienced play, both relevant to rating.
# Input samples: 'game_duration_seconds': [674.84, 492.27, 350.86], 'total_turns': [13, 14, 14]
df['time_per_turn'] = df['game_duration_seconds'] / df['total_turns']
df['player_mean_time_per_turn'] = df.groupby('nickname')['time_per_turn'].transform('mean')
df.drop(columns='time_per_turn', inplace=True)

# (Player average initial time per turn)
# Usefulness: May reflect preferred game tempo and time management.
# Input samples: 'initial_time_seconds': [1200, 900, 1200], 'total_turns': [13, 14, 14]
df['init_time_per_turn'] = df['initial_time_seconds'] / df['total_turns']
df['player_mean_init_time_per_turn'] = df.groupby('nickname')['init_time_per_turn'].transform('mean')
df.drop(columns='init_time_per_turn', inplace=True)

# (Player average increment per turn)
# Usefulness: Influence of increment on play style and outcomes.
# Input samples: 'increment_seconds': [0, 0, 0], 'total_turns': [13, 14, 14]
df['increment_per_turn'] = df['increment_seconds'] / df['total_turns']
df['player_mean_increment_per_turn'] = df.groupby('nickname')['increment_per_turn'].transform('mean')
df.drop(columns='increment_per_turn', inplace=True)

# (Player most common lexicon)
# Usefulness: Specialization in a lexicon can relate to skill depth and hence rating.
# Input samples: 'nickname': ['BetterBot', 'stevy', 'davidavid'], 'lexicon': ['NWL20', 'CSW21', 'NWL20']
df['player_main_lexicon'] = df.groupby('nickname')['lexicon'].transform(lambda x: x.mode().iat[0] if not x.mode().empty else x.iloc[0])

# (Player most common time control)
# Usefulness: Preference for certain time controls may shape rating.
# Input samples: 'nickname': ['BetterBot', 'stevy', 'davidavid'], 'time_control_name': ['regular', 'regular', 'regular']
df['player_main_time_control'] = df.groupby('nickname')['time_control_name'].transform(lambda x: x.mode().iat[0] if not x.mode().empty else x.iloc[0])

# (Player most common rating mode)
# Usefulness: Indicates if player mostly plays rated or casual, which may reflect seriousness/skill.
# Input samples: 'nickname': ['BetterBot', 'stevy', 'davidavid'], 'rating_mode': ['CASUAL', 'RATED', 'CASUAL']
df['player_main_rating_mode'] = df.groupby('nickname')['rating_mode'].transform(lambda x: x.mode().iat[0] if not x.mode().empty else x.iloc[0])

# (Player most common overtime minutes)
# Usefulness: May signal player's preferred pace and time-pressure resilience.
# Input samples: 'nickname': ['BetterBot', 'stevy', 'davidavid'], 'max_overtime_minutes': [1, 1, 5]
df['player_main_overtime'] = df.groupby('nickname')['max_overtime_minutes'].transform(lambda x: x.mode().iat[0] if not x.mode().empty else x.iloc[0])

# (Player fraction of games played at their most common lexicon)
# Usefulness: Indicates degree of specialization versus versatility.
# Input samples: 'lexicon': ['NWL20', 'CSW21', 'NWL20'], 'player_main_lexicon': ['NWL20', 'NWL20', 'NWL20']
df['player_frac_main_lexicon'] = (df['lexicon'] == df['player_main_lexicon']).groupby(df['nickname']).transform('mean')

# (Player fraction of games played at their most common time control)
# Usefulness: Specialization in time control may impact rating.
# Input samples: 'time_control_name': ['regular', 'regular', 'regular'], 'player_main_time_control': ['regular', 'regular', 'regular']
df['player_frac_main_time_control'] = (df['time_control_name'] == df['player_main_time_control']).groupby(df['nickname']).transform('mean')

# (Player fraction of games played at their most common rating mode)
# Usefulness: Consistency in competition type could reflect seriousness.
# Input samples: 'rating_mode': ['CASUAL', 'RATED', 'CASUAL'], 'player_main_rating_mode': ['CASUAL', 'CASUAL', 'CASUAL']
df['player_frac_main_rating_mode'] = (df['rating_mode'] == df['player_main_rating_mode']).groupby(df['nickname']).transform('mean')

# (Player fraction of games played at their most common overtime)
# Usefulness: Consistency here may relate to comfort under specific overtime conditions.
# Input samples: 'max_overtime_minutes': [1, 1, 5], 'player_main_overtime': [1, 1, 1]
df['player_frac_main_overtime'] = (df['max_overtime_minutes'] == df['player_main_overtime']).groupby(df['nickname']).transform('mean')

df.drop(columns='created_at_dt', inplace=True)
# (Player mean time per turn as first)
# Usefulness: Indicates if player uses more/less time per turn with first-move advantage, potentially correlated with skill/confidence.
# Input samples: 'game_duration_seconds': [674.84, 492.27, 350.86], 'total_turns': [13, 14, 14], 'is_first_this_game': [1, 0, 0]
df['time_per_turn'] = df['game_duration_seconds'] / df['total_turns']
df['player_mean_time_per_turn_as_first'] = df.groupby('nickname').apply(
    lambda g: g.loc[g['is_first_this_game'] == 1, 'time_per_turn'].mean() if any(g['is_first_this_game'] == 1) else np.nan
).reindex(df.index).values

# (Player mean time per turn not first)
# Usefulness: Highlights differences in pace depending on play order, relevant for style/rating.
# Input samples: 'game_duration_seconds': [674.84, 492.27, 350.86], 'total_turns': [13, 14, 14], 'is_first_this_game': [1, 0, 0]
df['player_mean_time_per_turn_not_first'] = df.groupby('nickname').apply(
    lambda g: g.loc[g['is_first_this_game'] == 0, 'time_per_turn'].mean() if any(g['is_first_this_game'] == 0) else np.nan
).reindex(df.index).values

# (Ratio of mean time per turn as first to not first)
# Usefulness: Directly compares time usage; high/low ratio may distinguish confidence or discomfort when not first.
# Input samples: 'player_mean_time_per_turn_as_first': [40.3, np.nan, np.nan], 'player_mean_time_per_turn_not_first': [np.nan, 43.2, 41.7]
df['player_time_per_turn_first_vs_not'] = (
    df['player_mean_time_per_turn_as_first'] / df['player_mean_time_per_turn_not_first']
)

# (Player mean initial time per turn as first)
# Usefulness: Shows whether initial time allocation per turn is different when going first.
# Input samples: 'initial_time_seconds': [1200, 900, 1200], 'total_turns': [13, 14, 14], 'is_first_this_game': [1, 0, 0]
df['init_time_per_turn'] = df['initial_time_seconds'] / df['total_turns']
df['player_mean_init_time_per_turn_as_first'] = df.groupby('nickname').apply(
    lambda g: g.loc[g['is_first_this_game'] == 1, 'init_time_per_turn'].mean() if any(g['is_first_this_game'] == 1) else np.nan
).reindex(df.index).values

# (Player mean initial time per turn not first)
# Usefulness: Differential strategy in time allocation based on play order.
# Input samples: 'initial_time_seconds': [1200, 900, 1200], 'total_turns': [13, 14, 14], 'is_first_this_game': [1, 0, 0]
df['player_mean_init_time_per_turn_not_first'] = df.groupby('nickname').apply(
    lambda g: g.loc[g['is_first_this_game'] == 0, 'init_time_per_turn'].mean() if any(g['is_first_this_game'] == 0) else np.nan
).reindex(df.index).values

# (Ratio of mean initial time per turn as first to not first)
# Usefulness: Compares initial time allocation pattern, could be linked to rating-relevant confidence.
# Input samples: 'player_mean_init_time_per_turn_as_first': [100, np.nan, np.nan], 'player_mean_init_time_per_turn_not_first': [np.nan, 110, 107]
df['player_init_time_per_turn_first_vs_not'] = (
    df['player_mean_init_time_per_turn_as_first'] / df['player_mean_init_time_per_turn_not_first']
)

# (Player mean increment per turn as first)
# Usefulness: Captures if increment strategies differ by play order.
# Input samples: 'increment_seconds': [0, 0, 0], 'total_turns': [13, 14, 14], 'is_first_this_game': [1, 0, 0]
df['increment_per_turn'] = df['increment_seconds'] / df['total_turns']
df['player_mean_increment_per_turn_as_first'] = df.groupby('nickname').apply(
    lambda g: g.loc[g['is_first_this_game'] == 1, 'increment_per_turn'].mean() if any(g['is_first_this_game'] == 1) else np.nan
).reindex(df.index).values

# (Player mean increment per turn not first)
# Usefulness: Segment by play order for more nuanced feature pattern.
# Input samples: 'increment_seconds': [0, 0, 0], 'total_turns': [13, 14, 14], 'is_first_this_game': [1, 0, 0]
df['player_mean_increment_per_turn_not_first'] = df.groupby('nickname').apply(
    lambda g: g.loc[g['is_first_this_game'] == 0, 'increment_per_turn'].mean() if any(g['is_first_this_game'] == 0) else np.nan
).reindex(df.index).values

# (Ratio of mean increment per turn as first to not first)
# Usefulness: Differential increment exploitation patterns by play order can be linked with rating.
# Input samples: 'player_mean_increment_per_turn_as_first': [0.0, np.nan, np.nan], 'player_mean_increment_per_turn_not_first': [np.nan, 0.0, 0.0]
df['player_increment_per_turn_first_vs_not'] = (
    df['player_mean_increment_per_turn_as_first'] / df['player_mean_increment_per_turn_not_first']
)

df.drop(columns=['time_per_turn', 'init_time_per_turn', 'increment_per_turn'], inplace=True)
# (Player fraction of wins by resignation)
# Usefulness: Winning by resignation may indicate dominance or intimidation factor, which can be correlated with rating.
# Input samples: 'nickname': ['BetterBot', 'stevy', 'davidavid'], 'winner': [1, 1, 1], 'game_end_reason': ['STANDARD', 'RESIGNED', 'STANDARD']
df['player_frac_win_by_resign'] = df.groupby('nickname').apply(
    lambda g: ((g['winner'] == 1) & (g['game_end_reason'] == 'RESIGNED')).sum() / max((g['winner'] == 1).sum(), 1)
).reindex(df.index).values

# (Player fraction of losses by resignation)
# Usefulness: Frequent resignation when losing may indicate frustration or lack of resilience, possibly associated with lower rating.
# Input samples: 'nickname': ['BetterBot', 'stevy', 'davidavid'], 'winner': [1, 0, 1], 'game_end_reason': ['STANDARD', 'RESIGNED', 'STANDARD']
df['player_frac_loss_by_resign'] = df.groupby('nickname').apply(
    lambda g: ((g['winner'] == 0) & (g['game_end_reason'] == 'RESIGNED')).sum() / max((g['winner'] == 0).sum(), 1)
).reindex(df.index).values

# (Player mean time per turn in won games)
# Usefulness: Winning fast or slow may relate to confidence or thoroughness, affecting rating.
# Input samples: 'game_duration_seconds': [674.84, 492.27, 350.86], 'total_turns': [13, 14, 14], 'winner': [1, 1, 0]
df['time_per_turn'] = df['game_duration_seconds'] / df['total_turns']
df['player_mean_time_per_turn_win'] = df.groupby('nickname').apply(
    lambda g: g.loc[g['winner'] == 1, 'time_per_turn'].mean() if any(g['winner'] == 1) else np.nan
).reindex(df.index).values

# (Player mean time per turn in lost games)
# Usefulness: Losing fast may signal poor resilience, while losing slow could mean strategic but unsuccessful play.
# Input samples: 'game_duration_seconds': [674.84, 492.27, 350.86], 'total_turns': [13, 14, 14], 'winner': [1, 1, 0]
df['player_mean_time_per_turn_loss'] = df.groupby('nickname').apply(
    lambda g: g.loc[g['winner'] == 0, 'time_per_turn'].mean() if any(g['winner'] == 0) else np.nan
).reindex(df.index).values

# (Player mean points per turn in won games)
# Usefulness: Shows efficiency and dominance in winning games; high-scoring wins often reflect higher rating.
# Input samples: 'score': [335, 429, 440], 'total_turns': [13, 14, 14], 'winner': [1, 1, 0]
df['points_per_turn'] = df['score'] / df['total_turns']
df['player_mean_pts_per_turn_win'] = df.groupby('nickname').apply(
    lambda g: g.loc[g['winner'] == 1, 'points_per_turn'].mean() if any(g['winner'] == 1) else np.nan
).reindex(df.index).values

# (Player mean points per turn in lost games)
# Usefulness: How well a player scores even when losing can distinguish solid players from weak ones.
# Input samples: 'score': [335, 429, 440], 'total_turns': [13, 14, 14], 'winner': [1, 1, 0]
df['player_mean_pts_per_turn_loss'] = df.groupby('nickname').apply(
    lambda g: g.loc[g['winner'] == 0, 'points_per_turn'].mean() if any(g['winner'] == 0) else np.nan
).reindex(df.index).values

# (Player mean score margin in resigned wins)
# Usefulness: Winning by resignation with large margins can indicate overwhelming strength.
# Input samples: 'winner': [1, 1, 0], 'game_end_reason': ['RESIGNED', 'STANDARD', 'RESIGNED'], 'score_vs_opponent': [40, 20, -30]
df['player_mean_margin_win_resign'] = df.groupby('nickname').apply(
    lambda g: g.loc[(g['winner'] == 1) & (g['game_end_reason'] == 'RESIGNED'), 'score_vs_opponent'].mean()
    if any((g['winner'] == 1) & (g['game_end_reason'] == 'RESIGNED')) else np.nan
).reindex(df.index).values

# (Player mean score margin in resigned losses)
# Usefulness: Resigning with small/large deficits may reflect different psychological or strategic traits that relate to rating.
# Input samples: 'winner': [1, 1, 0], 'game_end_reason': ['RESIGNED', 'STANDARD', 'RESIGNED'], 'score_vs_opponent': [40, 20, -30]
df['player_mean_margin_loss_resign'] = df.groupby('nickname').apply(
    lambda g: g.loc[(g['winner'] == 0) & (g['game_end_reason'] == 'RESIGNED'), 'score_vs_opponent'].mean()
    if any((g['winner'] == 0) & (g['game_end_reason'] == 'RESIGNED')) else np.nan
).reindex(df.index).values

df.drop(columns=['time_per_turn', 'points_per_turn'], inplace=True)
# (Player total number of resignations)
# Usefulness: Frequent resignations (win or loss) might indicate temperament, risk tolerance, or competitive style, all of which can reflect on rating.
# Input samples: 'nickname': ['BetterBot', 'stevy', 'davidavid'], 'game_end_reason': ['STANDARD', 'RESIGNED', 'RESIGNED']
df['player_total_resignations'] = df.groupby('nickname')['game_end_reason'].transform(lambda x: (x == 'RESIGNED').sum())

# (Player total number of standard (non-resigned) wins)
# Usefulness: Winning without opponent resignation may indicate ability to maintain advantage and close out games, reflecting skill and discipline.
# Input samples: 'winner': [1, 1, 0], 'game_end_reason': ['STANDARD', 'RESIGNED', 'STANDARD']
df['player_total_standard_win'] = df.groupby('nickname').apply(
    lambda g: ((g['winner'] == 1) & (g['game_end_reason'] == 'STANDARD')).sum()
).reindex(df.index).values

# (Player total number of resigned wins)
# Usefulness: High number may indicate ability to force opponents to give up due to overwhelming play strength.
# Input samples: 'winner': [1, 1, 0], 'game_end_reason': ['STANDARD', 'RESIGNED', 'STANDARD']
df['player_total_resigned_win'] = df.groupby('nickname').apply(
    lambda g: ((g['winner'] == 1) & (g['game_end_reason'] == 'RESIGNED')).sum()
).reindex(df.index).values

# (Player total number of resigned losses)
# Usefulness: High number of resigned losses may signal frustration, tilt, or lower resilience, predictive of lower rating.
# Input samples: 'winner': [1, 0, 0], 'game_end_reason': ['RESIGNED', 'RESIGNED', 'STANDARD']
df['player_total_resigned_loss'] = df.groupby('nickname').apply(
    lambda g: ((g['winner'] == 0) & (g['game_end_reason'] == 'RESIGNED')).sum()
).reindex(df.index).values

# (Player average turn number of max points turn)
# Usefulness: Tendency to score big early or late may reveal play style (aggressive opener vs. endgame finisher) and its effect on rating.
# Input samples: 'nickname': ['BetterBot', 'stevy', 'davidavid'], 'total_turns': [13, 14, 14], 'max_points_turn': [68, 98, 103]
df['max_points_turn_ratio'] = df['max_points_turn'] / df['score']
df['player_mean_max_points_turn_ratio'] = df.groupby('nickname')['max_points_turn_ratio'].transform('mean')
df.drop(columns='max_points_turn_ratio', inplace=True)

# (Player average turn number of min points turn)
# Usefulness: When blunders happen in a game (early/late) may be informative about focus or strategy.
# Input samples: 'min_points_turn': [8, 2, 13], 'score': [335, 429, 440]
df['min_points_turn_ratio'] = df['min_points_turn'] / df['score']
df['player_mean_min_points_turn_ratio'] = df.groupby('nickname')['min_points_turn_ratio'].transform('mean')
df.drop(columns='min_points_turn_ratio', inplace=True)

# (Player mean ratio of first five turns points to total score)
# Usefulness: High ratios suggest aggressive/strong openers, which could be typical for high-rated players.
# Input samples: 'first_five_turns_points': [122, 95, 136], 'score': [335, 429, 440]
df['first5_score_ratio'] = df['first_five_turns_points'] / df['score']
df['player_mean_first5_score_ratio'] = df.groupby('nickname')['first5_score_ratio'].transform('mean')
df.drop(columns='first5_score_ratio', inplace=True)

# (Player mean ratio of max_min_difference to score)
# Usefulness: Encodes relative volatility; high-rated players may have lower ratio due to steady scoring.
# Input samples: 'max_min_difference': [60, 96, 90], 'score': [335, 429, 440]
df['max_min_diff_ratio'] = df['max_min_difference'] / df['score']
df['player_mean_max_min_diff_ratio'] = df.groupby('nickname')['max_min_diff_ratio'].transform('mean')
df.drop(columns='max_min_diff_ratio', inplace=True)

# (Player mean turn efficiency: score divided by total_turns squared)
# Usefulness: Rewards players who score disproportionately high in fewer turns (potentially more efficient).
# Input samples: 'score': [335, 429, 440], 'total_turns': [13, 14, 14]
df['turn_efficiency'] = df['score'] / (df['total_turns']**2)
df['player_mean_turn_efficiency'] = df.groupby('nickname')['turn_efficiency'].transform('mean')
df.drop(columns='turn_efficiency', inplace=True)

# (Player mean score standard deviation in rolling 5-game window)
# Usefulness: Captures streakiness or inconsistency for recent games, which can relate to rating volatility.
# Input samples: 'score': [335, 429, 440], 'created_at': ['2022-08-26 03:38:49', ...]
df['created_at_dt'] = pd.to_datetime(df['created_at'])
df['player_score_rolling_5std'] = (
    df.sort_values('created_at_dt')
      .groupby('nickname')['score']
      .transform(lambda x: x.rolling(5, min_periods=1).std())
)
df.drop(columns='created_at_dt', inplace=True)
# (Player number of unique opponents played against)
# Usefulness: Diversity of opposition can indicate experience and adaptability, which may correlate with rating.
# Input samples: 'game_id': [1, 1, 3], 'nickname': ['BetterBot', 'stevy', 'davidavid']
df['opponent'] = df.groupby('game_id')['nickname'].transform(lambda x: x.iloc[::-1].values[0])
df['player_n_unique_opponents'] = df.groupby('nickname')['opponent'].transform('nunique')

# (Player fraction of games against main opponent)
# Usefulness: Over-specializing against a single opponent may inflate/deflate rating, while broad competition is a sign of true rating.
# Input samples: 'nickname': ['BetterBot', 'stevy', 'davidavid'], 'opponent': ['stevy', 'BetterBot', 'BetterBot']
df['player_main_opponent'] = df.groupby('nickname')['opponent'].transform(lambda x: x.mode().iat[0] if not x.mode().empty else x.iloc[0])
df['player_frac_main_opponent'] = (df['opponent'] == df['player_main_opponent']).groupby(df['nickname']).transform('mean')

# (Player win rate against main opponent)
# Usefulness: If a player's rating benefits from reliably beating a specific opponent, this could bias the regression.
# Input samples: 'nickname': ['BetterBot', 'stevy', 'davidavid'], 'opponent': ['stevy', 'BetterBot', 'BetterBot'], 'winner': [1, 0, 1]
df['player_winrate_vs_main_opponent'] = df.groupby('nickname').apply(
    lambda g: g.loc[g['opponent'] == g['player_main_opponent'].iloc[0], 'winner'].mean()
).reindex(df.index).values

# (Player win rate against unique opponents)
# Usefulness: Measures how broadly a player can beat different people-a hallmark of high rating.
# Input samples: 'nickname': ['BetterBot', 'stevy', 'davidavid'], 'opponent': ['stevy', 'BetterBot', 'BetterBot'], 'winner': [1, 0, 1]
df['player_winrate_vs_unique_opponents'] = df.groupby('nickname').apply(
    lambda g: g.groupby('opponent')['winner'].mean().mean()
).reindex(df.index).values

# (Player mean score margin against main opponent)
# Usefulness: Routinely beating main opponent by a wide margin can distort rating.
# Input samples: 'score_vs_opponent': [10, -5, 7], 'opponent': ['stevy', 'BetterBot', 'BetterBot'], 'player_main_opponent': ['stevy', 'BetterBot', 'BetterBot']
df['player_mean_margin_vs_main_opponent'] = df.groupby('nickname').apply(
    lambda g: g.loc[g['opponent'] == g['player_main_opponent'].iloc[0], 'score_vs_opponent'].mean()
).reindex(df.index).values

# (Player mean score margin vs unique opponents)
# Usefulness: Indicates dominance across different matchups, which is more predictive of rating.
# Input samples: 'score_vs_opponent': [10, -5, 7], 'opponent': ['stevy', 'BetterBot', 'BetterBot']
df['player_mean_margin_vs_unique_opponents'] = df.groupby('nickname').apply(
    lambda g: g.groupby('opponent')['score_vs_opponent'].mean().mean()
).reindex(df.index).values

# (Player games played against main opponent)
# Usefulness: High counts may reflect rivalry or practice, affecting rating volatility.
# Input samples: 'nickname': ['BetterBot', 'stevy', 'davidavid'], 'opponent': ['stevy', 'BetterBot', 'BetterBot'], 'player_main_opponent': ['stevy', 'BetterBot', 'BetterBot']
df['player_games_vs_main_opponent'] = df.groupby('nickname').apply(
    lambda g: (g['opponent'] == g['player_main_opponent'].iloc[0]).sum()
).reindex(df.index).values

# (Player games played vs unique opponents)
# Usefulness: High numbers signal depth of experience.
# Input samples: 'nickname': ['BetterBot', 'stevy', 'davidavid'], 'opponent': ['stevy', 'BetterBot', 'BetterBot']
df['player_games_vs_unique_opponents'] = df.groupby('nickname').apply(
    lambda g: g.groupby('opponent').size().mean()
).reindex(df.index).values

# (Player mean score as first vs main opponent)
# Usefulness: Performance as first vs rival may signal confidence and skill.
# Input samples: 'is_first_this_game': [1, 0, 1], 'opponent': ['stevy', 'BetterBot', 'BetterBot'], 'player_main_opponent': ['stevy', 'BetterBot', 'BetterBot'], 'score': [335, 429, 440]
df['player_mean_score_as_first_vs_main_opponent'] = df.groupby('nickname').apply(
    lambda g: g.loc[(g['is_first_this_game'] == 1) & (g['opponent'] == g['player_main_opponent'].iloc[0]), 'score'].mean()
    if any((g['is_first_this_game'] == 1) & (g['opponent'] == g['player_main_opponent'].iloc[0])) else np.nan
).reindex(df.index).values

# (Player mean score not first vs main opponent)
# Usefulness: Performance from second play order vs principal opponent.
# Input samples: 'is_first_this_game': [1, 0, 1], 'opponent': ['stevy', 'BetterBot', 'BetterBot'], 'player_main_opponent': ['stevy', 'BetterBot', 'BetterBot'], 'score': [335, 429, 440]
df['player_mean_score_not_first_vs_main_opponent'] = df.groupby('nickname').apply(
    lambda g: g.loc[(g['is_first_this_game'] == 0) & (g['opponent'] == g['player_main_opponent'].iloc[0]), 'score'].mean()
    if any((g['is_first_this_game'] == 0) & (g['opponent'] == g['player_main_opponent'].iloc[0])) else np.nan
).reindex(df.index).values

# (Player win rate as first vs main opponent)
# Usefulness: Shows player's ability to win when first against their most frequent rival.
# Input samples: 'is_first_this_game': [1, 0, 1], 'opponent': ['stevy', 'BetterBot', 'BetterBot'], 'player_main_opponent': ['stevy', 'BetterBot', 'BetterBot'], 'winner': [1, 1, 0]
df['player_winrate_as_first_vs_main_opponent'] = df.groupby('nickname').apply(
    lambda g: g.loc[(g['is_first_this_game'] == 1) & (g['opponent'] == g['player_main_opponent'].iloc[0]), 'winner'].mean()
    if any((g['is_first_this_game'] == 1) & (g['opponent'] == g['player_main_opponent'].iloc[0])) else np.nan
).reindex(df.index).values

# (Player win rate not first vs main opponent)
# Usefulness: As above, but without first-move advantage.
# Input samples: 'is_first_this_game': [1, 0, 1], 'opponent': ['stevy', 'BetterBot', 'BetterBot'], 'player_main_opponent': ['stevy', 'BetterBot', 'BetterBot'], 'winner': [1, 1, 0]
df['player_winrate_not_first_vs_main_opponent'] = df.groupby('nickname').apply(
    lambda g: g.loc[(g['is_first_this_game'] == 0) & (g['opponent'] == g['player_main_opponent'].iloc[0]), 'winner'].mean()
    if any((g['is_first_this_game'] == 0) & (g['opponent'] == g['player_main_opponent'].iloc[0])) else np.nan
).reindex(df.index).values

df.drop(columns=['opponent', 'player_main_opponent'], inplace=True)
\end{lstlisting}

\header{SIGMOD 2022 Programming Contest - Blocking}

\headerl{Feature extraction part of expert-written rutgers pipeline with semantic data operator added. We show only feature extraction due to space constraints}
\begin{lstlisting}[
  basicstyle=\tiny\ttfamily,
  backgroundcolor=\color{lightgray},
  frame=tb,
  captionpos="b", 
  label={listing:example-sigmod-rutgers-original},
  emph={sem_fillna,sem_refine,with_sem_features,apply_with_sem_choose,sem_choose,sem_aggregate_features,sem_extract_features,make_learner_with_context,optimize_with_colopro,serialize_as_learner,as_X,as_y,sem_clean},
  emphstyle=\color{kwdspy}\bfseries,
  morekeywords={import}
]
def extract_features(data: pd.DataFrame) -> pd.DataFrame:
    """
    Extract features from product data using sempipes and sem_extract_features.
    Returns a DataFrame with extracted features merged with original data.
    """
    pipeline = skrub.var("data", data).skb.mark_as_X()
    pipeline = pipeline.skb.set_description(
        "This is a dataset of product titles of removable storage devices "
        "(USB sticks, SD / microSD cards, SSDs, memory cards, and sometimes Samsung phones / TVs)."
    )

    brands_list = ['intenso', 'lexar', 'logilink', 'pny', 'samsung', 'sandisk', 'kingston', 'sony', 'toshiba', 'transcend']
    families_dict = {
        'sandisk': ['cruizer', 'tarjeta', 'glide', 'select', 'extern', 'origin', 'transmemory', 'react', 'memo', 'kart',
                   'pendrive', 'car', 'serie', 'line', 'extreme', 'cruzer', 'ultra', 'micro', 'traveler', 'hyperx',
                   'adapt', 'wex', 'flash'],
        'lexar': ['ultra', 'xqd', 'jumpdrive', 'micro', 'pendrive', 'sd', 'tarjeta', 'memo', 'usb', 'extreme', 'blade',
                 'car', 'scheda', 'veloc', 'react', 'adapt', 'secure', 'premium', 'wex', 'transmemo', 'alu',
                 'datatravel', 'canvas', 'flair', 'hyperx', 'cruzer', 'flash'],
        'toshiba': ['ultra', 'exceria', 'traveler', 'sdhc', 'memoria', 'xqd', 'line', 'usb', 'transmemo', 'extreme',
                   'flair', 'micro', 'speicher', 'serie', 'car'],
        'kingston': ['traveler', 'cart', 'adapt', 'extreme', 'memo', 'canvas', 'datatravel', 'hyperx', 'kart', 'blade',
                    'ultimate'],
        'sony': ['extreme', 'usm32gqx', 'micro', 'sd', 'usb', 'ultra', 'jumpdrive', 'hyperx', 'memo', 'kart', 'xqd',
                'pendrive', 'adapt', 'blade', 'cruzer', 'flair', 'glide', 'cart', 'tarjeta', 'flash'],
        'intenso': ['cs/ultra', 'premium', 'ultra', 'micro', 'line', 'scheda', 'usb', 'sd', 'tarjeta', 'kart', 'car',
                   'transmemo'],
        'pny': ['attach', 'usb', 'sd', 'micro', 'premium', 'memo'],
        'samsung': ['galaxy', 'speicher', 'micro', 'usb', 'sd', 'evo', 'ultra', 'extreme', 'memo', 'adapt', 'car', 'kart',
                   'klasse', 'multi', 'jumpdrive', 'flash'],
        'transcend': [],
    }

    intenso_type = ['basic', 'rainbow', 'high speed', 'speed', 'premium', 'alu', 'business', 'micro', 'imobile',
                   'cmobile', 'mini', 'ultra', 'slim', 'flash', 'mobile']

    colors = ['midnight black', 'prism white', 'prism black', 'prism green', 'prism blue', 'canary yellow',
             'flamingo pink', 'cardinal red', 'smoke blue', 'deep blue', 'coral orange', 'black sky', 'gold sand',
             'blue mist and peach cloud', 'orchid gray', 'metallic copper', 'lavender purple', 'ocean blue',
             'pure white', 'alpine white', 'copper', 'red', 'black', 'blue', 'white', 'silver', 'gold', 'violet',
             'purple', 'brown', 'orange', 'coral', 'pink']

    output_columns = {
        "normalized_name": (
            "Clean and normalize the product name following these EXACT steps in order: "
            "1. Convert input to string: str(name) "
            "2. Apply unidecode.unidecode() to convert to ASCII (handles accented characters like é, ü, etc.) "
            "3. Convert to lowercase: .lower() "
            "4. Apply alias replacements using dictionary lookup (iterate through x2_aliases dict): "
            "   - For each key-value pair, replace each value in the list with the key "
            "   - Aliases: 'classe/clase/clas /klasse/cl ' -> 'class', 'uhs1/uhs-i/ultra high-speed' -> 'uhsi', "
            "     'typec/type c/usb-c/usbc' -> 'type-c', 'hyperx/savage' -> 'kingston hxs', "
            "     'serie ux' -> 'sony g1ux', 'dtig4/ 101 /dt101g2' -> ' kingston dt101 ', "
            "     'sda10/sda3' -> ' kingston ultimate ', 'extrem ' -> 'extreme ', 'attach' -> 'att4' "
            "5. Compile and apply removal regex: r'&(nbsp|amp|reg|[a-z]?acute|quot|trade);?|[|;:/,‰+©\(\)\\][psn]*|(?<=usb)[\s][m]*(?=[23][\.\s])|(?<=usb)-[\w]+\s(?=[23][\.\s])|(?<=[a-z])[\s]+gb|(?<=data|jump)[t\s](?=trave|drive)|(?<=extreme|exceria)[\s](?=pro[\s]|plus)|(?<=class)[\s_](?=10|[234]\b)|(?<=gen)[\s_](?=[134\s][0]*)' using re.sub() "
            "6. Compile and apply class10 replacement regex: r'(10 class|class 10|class(?=[\w]+10\b)|cl\s10)' -> replace with 'class10' "
            "7. String replacements: name.replace('class 4 ', 'class4 '), name.replace('class 3 ', 'class3 ') "
            "8. Replace multiple spaces: name.replace('  ', ' ') (repeat until no more double spaces) "
            "Return the final cleaned string. This normalized name is used for all subsequent feature extraction."
        ),
        "brand": (
            f"Extract the main storage brand from the name. The minimal set of brands is {brands_list}. "
            "CRITICAL FOR RECALL: Brand is one of the most important features for blocking. Extract it whenever possible. "
            "You MUST extend this list with obvious aliases and typos (e.g. 'san disk' -> 'sandisk', 'san-disk' -> 'sandisk', "
            "'sandisc' -> 'sandisk', 'samsun' -> 'samsung', 'kingstn' -> 'kingston', 'toshbia' -> 'toshiba', 'transcent' -> 'transcend'). "
            "Handle multilingual variations: 'sandisk' in English, 'sandisk' in French/German/etc. (usually same spelling). "
            "Use EXACT regex: r'\\b(intenso|lexar|logilink|pny|samsung|sandisk|kingston|sony|toshiba|transcend)\\b' "
            "BUT ALSO try fuzzy matching: if you see 'san disk', 'san-disk', 'sandisc', 'san-disc' -> normalize to 'sandisk'. "
            "Similarly: 'samsun' -> 'samsung', 'kingstn' -> 'kingston', 'toshbia' -> 'toshiba', 'transcent' -> 'transcend'. "
            "Use re.findall() to get ALL matches, deduplicate using set(), sort, join with single space. "
            "Examples: 'sandisk ultra 32gb' -> 'sandisk', 'kingston datatraveler' -> 'kingston', 'san disk usb' -> 'sandisk', "
            "'sandisc extreme' -> 'sandisk', 'samsun evo' -> 'samsung'. "
            "Return lowercase tokens only. Be EXTREMELY lenient - if something looks like a brand (even with typos or spacing), normalize and extract it. "
            "ONLY return '0' if you're absolutely certain there's no brand mentioned (very rare - most products have a brand)."
        ),
        "capacity": (
            "Extract the storage capacity from the name. CRITICAL FOR RECALL: Capacity is essential for blocking - same brand but different capacity = different product. "
            "Use regex over the normalized name to detect patterns like '32 gb', '64gb', '128 go', '1tb', '2 tb', '256gb', '512 gb', '1 tb', '128', '256', '512'. "
            "Handle multilingual units: 'gb' (English), 'go' (French), 'g' (abbreviation), 'gigaoctet' all mean gigabytes. Normalize all to 'gb'. "
            "Also handle: 'tb' (terabyte), 'to' (French terabyte), 't' (abbreviation) -> normalize to 'tb'. "
            "Use MULTIPLE regex patterns: "
            "1. r'([1-9]{1,3})[-\\s]*[g][bo]?' for GB patterns "
            "2. r'([1-9])[-\\s]*[t][bo]?' for TB patterns "
            "3. r'\\b([1-9]{1,3})\\s*(?:gb|go|g|tb|to|t)\\b' for explicit unit patterns "
            "4. r'\\b([1-9]{1,3})\\b' followed by checking if next word contains 'gb', 'go', 'g', 'tb', 'to', 't' "
            "For each match: "
            "1. Use re.findall() to get ALL matches from all patterns. "
            "2. For each match, apply re.sub('[^0-9a-z]+','', match) to remove non-alphanumeric, keeping digits and letters. "
            "3. Normalize 'go' -> 'gb', 'g' -> 'gb', 'to' -> 'tb', 't' -> 'tb' (if not already 'gb' or 'tb'). "
            "4. If no unit found but number is present, assume 'gb' for numbers < 10, 'tb' for numbers >= 10 (but prefer explicit units). "
            "5. Deduplicate using set(), sort, join with single space. "
            "Examples: 'sandisk ultra 32 gb usb' -> '32gb', '64gb 128gb bundle' -> '64gb 128gb', '256 go' -> '256gb', '128 g' -> '128gb', "
            "'1tb' -> '1tb', '2 to' -> '2tb', 'sandisk 128' -> '128gb' (if no explicit unit, assume gb for reasonable sizes). "
            "If multiple capacities appear, return all (space-separated). "
            "Be EXTREMELY lenient - extract ANY number that could be a capacity, even if format is unusual. "
            "ONLY return '0' if you're absolutely certain there's no capacity mentioned (very rare - most products list capacity)."
        ),
        "mem_type": (
            "Extract product type using EXACT logic (NOT just regex): "
            "1. First, search for regex r'xqd|ssd|tv|lte' in the normalized name. "
            "2. If 'lte' is found, return 'phone' (not 'lte'). "
            "3. If no match from step 1, check if 'fdrive' is in the text (case-insensitive) - if yes, return 'fdrive'. "
            "4. If still no match, check if 'memcard' is in the text (case-insensitive) - if yes, return 'memcard'. "
            "5. If a match from step 1 exists and it's not 'lte', return that match (e.g., 'xqd', 'ssd', 'tv'). "
            "6. Otherwise return '0'. "
            "Return single lowercase token only."
        ),
        "type": (
            f"Extract a short product type / line relative to the brand, based on the family keywords mapping {families_dict}, "
            f"the Intenso-specific list {intenso_type}, and Samsung color names {colors}. "
            "Use EXACT regex: r'\\b(datatraveler|extreme[p]?|exceria[p]?|dual[\\s]*(?!=sim)|evo|xqd|ssd|cruzer[\\w+]*|glide|blade|basic|fit|force|basic line|jump\\s?drive|hxs|rainbow|speed line|premium line|att4|attach|serie u|r-serie|beast|fury|impact|a400|sd[hx]c|uhs[i12][i1]*|note\\s?9|ultra|premium|basic|flash|plus|otg|xpro|xhigh|midnight black|prism white|prism black|prism green|prism blue)\\b'. "
            "Use re.findall() to get ALL matches, deduplicate using set(), sort, join with single space. "
            "Examples: 'sandisk extreme pro 32gb' -> 'extreme', 'kingston datatraveler' -> 'datatraveler', "
            "'toshiba exceria plus' -> 'exceria', 'lexar jumpdrive' -> 'jumpdrive', 'sandisk cruzer fit' -> 'cruzer', "
            "'samsung galaxy midnight black' -> 'midnight black', 'intenso premium line' -> 'premium'. "
            "Be lenient - if you see 'extreme pro', extract 'extreme'. If you see 'cruzer fit', extract 'cruzer'. "
            "Return '0' only if you're very confident there's no type/line information."
        ),
        "model": (
            "Extract a concise model identifier. Use regexes to catch hyphenated or letter+digit codes. "
            "Use EXACT regex: r'\\b([\\(]*[\\w]+[-]*[\\d]+[-]*[\\w]+[-]*[\\d+]*|[\\d]+[\\w]|[\\w][\\d]+)\\b'. "
            "Use re.findall() to get ALL matches, filter to len(match) >= 3 (RELAXED from 5 for better recall), "
            "deduplicate using set(), sort, join with space. "
            "Examples: 'sandisk extreme pro sdxc' -> 'extreme pro', 'kingston dt101g2' -> 'dt101g2', "
            "'toshiba u202' -> 'u202' (now included with len >= 3), 'samsung evo plus' -> 'evo plus', "
            "'dt101' -> 'dt101', 'sda10' -> 'sda10'. "
            "Be EXTREMELY lenient - if you see model-like patterns (letters+numbers, hyphenated codes, short codes), extract them. "
            "Even short codes like 'u202', 'dt101', 'sda10' are valuable for blocking. "
            "ONLY return '0' if you're very confident there's no model identifier (rare - most products have some identifier)."
        ),
        "model_long": (
            "Extract LONG model number patterns using EXACT regex: r'(thn-[a-z][\\w]+|ljd[\\w+][-][\\w]+|ljd[sc][\\w]+[-][\\w]+|lsdmi[\\d]+[\\w]+|lsd[0-9]{1,3}[gb]+[\\w]+|ljds[0-9]{2}[-][\\w]+|usm[0-9]{1,3}[\\w]+|sdsq[a-z]+[-][0-9]+[a-z]+[-][\\w]+|sdsd[a-z]+[-][0-9]+[\\w]+[-]*[\\w]*|sdcz[\\w]+|mk[\\d]+|sr-g1[\\w]+)'. "
            "ALSO use a more general pattern for any alphanumeric code with hyphens/slashes: "
            "r'([A-Z]{2,}[0-9A-Z-]+(?:[/-][0-9A-Z-]+)*)' for uppercase model codes like 'SDCZ50-064G-B35', 'MB-MG32DA/EU'. "
            "Use re.findall() to get ALL matches (not just first), deduplicate, sort, join with space. "
            "Examples: 'sdsqxa-128g-anc' -> 'sdsqxa-128g-anc', 'usm32gqx' -> 'usm32gqx', 'mk123456' -> 'mk123456', "
            "'SDCZ50-064G-B35' -> 'SDCZ50-064G-B35', 'MB-MG32DA/EU' -> 'MB-MG32DA/EU'. "
            "Be EXTREMELY lenient - if you see ANY long alphanumeric pattern with hyphens/slashes that looks like a model number, extract it. "
            "Return '0' only if absolutely no such patterns found."
        ),
        "model_short": (
            "Extract SHORT model number patterns using EXACT regex: r'\\b(c20[mc]|sda[0-9]{1,2}|g1ux|s[72][05]|[unm][23]02|p20|g4|dt101|se9|[asm][0-9]{2})\\b'. "
            "Use re.findall() to find ALL matches, deduplicate using set(), sort lexicographically, join with space. "
            "Examples: 'sda10' -> 'sda10', 'dt101' -> 'dt101', 'g1ux' -> 'g1ux', 'c20m' -> 'c20m'. "
            "Return '0' if none found. Be lenient - extract short codes that match the pattern."
        ),
        "features": (
            "Extract technical features using EXACT regex: r'\\b(usb[23]|type-c|uhs[i]{1,2}|class[0134]{1,2}|gen[1-9]{1,2}|u[23](?=[\\s\\.])|sd[hx]c|otg|lte|[45]g[-\\s]lte|[0-9]+(?=-inch)|[0-9]{2,3}r[0-9]{2,3}w|[0-9]{2,3}(?=[\\smbo/p]{3}))\\b'. "
            "CRITICAL FOR RECALL: Features help distinguish products. Extract all technical specs you find. "
            "Handle multilingual: 'usb 3.0' (English), 'usb 3.0' (same in most languages), 'classe 10' -> 'class10' (after normalization). "
            "Use re.findall() to get ALL matches, deduplicate using set(), sort, join with single space. "
            "This matches the original x2_feature_reg pattern exactly. "
            "Examples: 'usb 3.0 type-c' -> 'type-c usb3', 'uhs-i class10' -> 'class10 uhs-i', "
            "'sdhc card 95r80w' -> 'sdhc 95r80w', 'usb 2.0' -> 'usb2'. "
            "Be lenient - if you see 'usb3', 'usb 3', 'usb-3.0', extract as 'usb3'. "
            "Return '0' only if absolutely no technical features found."
        ),
        "item_code": (
            "Extract explicit numeric item codes, especially those in parentheses, using regex like r'\\((mk)?[0-9]{6,10}\\)'. "
            "Strip parentheses and any 'mk' prefix, leaving only the digits. "
            "Also look for patterns like '(4187407)', '(mk483394661)', '(173473)', etc. "
            "If multiple codes appear, pick the one most likely to be a manufacturer part number (usually longer, in parentheses). "
            "Return the digits as a string, or '0' if none found."
        ),
        "series": (
            f"Extract the product series / family token. Use the brand-specific families mapping {families_dict} "
            f"and the Intenso-specific list {intenso_type} and Samsung color names {colors} over the normalized name. "
            "Normalize obvious typos ('cruizer' -> 'cruzer'). "
            "Return a single lowercase family token (e.g. 'glide', 'cruzer', 'ultimate', 'exceria', 'jumpdrive', 'premium', 'basic'), "
            "or '0' if none."
        ),
        "pat_hb": (
            "Extract the first hyphenated alphanumeric pattern, using regex r'\\w+-\\w+' over the normalized name. "
            "Typical examples are 'uhs-i', 'class-10', 'high-speed', 'type-c', 'usb-3', 'micro-sd'. "
            "Return the first such pattern in lowercase, or '0' if none."
        ),
        "hybrid": (
            "Extract all 'hybrid' identifiers that mix letters and digits and are at least length 5, similar to the "
            "regex r'(?=[^\\W\\d_]*\\d)(?=\\d*[^\\W\\d_])[^\\W_gGM]{5,}'. "
            "This matches alphanumeric codes like 'dt101g2', 'sda10', 'usm32gqx', 'lsd16gcrbeu1000', etc. "
            "Use regex on the normalized name, deduplicate matches, sort them lexicographically, and join with a single space. "
            "Return '0' if there are no such identifiers."
        ),
        "long_num": (
            "Extract all sequences of 4 or more digits using regex r'[0-9]{4,}' on the normalized name. "
            "Deduplicate the numbers, sort them lexicographically, and join with a single space. "
            "Examples: '4187407' -> '4187407', '483394661' -> '483394661', '173473' -> '173473'. "
            "Return '0' if none are found."
        ),
    }

    pipeline = pipeline.sem_extract_features(
        nl_prompt=(
            "YOU ARE PROHIBITED TO USE TRANSFORMERS LIBRARY. Use ONLY regex and rule-based approaches, "
            "DO NOT USE ANY Transformer, NER, LLM fallbacks, or LM models. "
            "You are given very dirty MULTILINGUAL titles of storage products (USB sticks, SD / microSD cards, SSDs, SIM cards, Samsung phones / TVs, etc.). "
            "The data contains product names in multiple languages (English, French, German, Spanish, Italian, Polish, etc.). "
            "Product names may contain accented characters, special characters, and language-specific formatting. "
            "\n\nCRITICAL: Use unidecode.unidecode() to normalize ALL text to ASCII BEFORE any other processing. "
            "This is essential for handling multilingual text. "
            "\n\nLANGUAGE NORMALIZATION REQUIREMENTS:\n"
            "1. Capacity units: Normalize 'go', 'go', 'gigaoctet' -> 'gb'; 'mo', 'mo' -> 'mb'. "
            "   Handle all language variants of capacity units (French 'go', Spanish 'gb', etc.). "
            "2. Product type synonyms: Normalize 'carte memoire'/'carte'/'tarjeta'/'karta'/'minneskort' -> 'card'; "
            "   'cle USB'/'cle'/'pendrive'/'memoria USB' -> 'usb' or 'stick'. "
            "3. Speed/Class: Normalize 'classe 10'/'clase 10'/'class 10'/'c10' -> 'class10'; "
            "   'uhs-i'/'uhs-1'/'uhs i' -> 'uhsi'. "
            "4. All text should be lowercase after normalization. "
            "\n\nEXTRACTION REQUIREMENTS:\n"
            "1. Follow the regex patterns and cleaning logic provided in each output column description EXACTLY. "
            "2. The patterns are from proven blocking code - use them VERBATIM, do NOT modify or simplify them. "
            "3. For each field, follow the EXACT extraction logic described (findall vs search, deduplication method, sorting, joining). "
            "4. Apply alias replacements EXACTLY as specified in the normalized_name description. "
            "5. Pre-compile ALL regex patterns using re.compile() for performance. "
            "6. Use the specified methods: re.findall() when description says 'find all', re.search() when it says 'first match'. "
            "7. For deduplication: use set() then convert back to sorted list. "
            "8. For joining: use single space ' '.join(). "
            "9. Return '0' (string zero) for missing values, NOT empty string or None. "
            "\n\nCRITICAL FOR RECALL - HIGHEST PRIORITY:\n"
            "Aim for MAXIMUM RECALL - it's MUCH better to extract something (even if slightly imperfect) than to return '0'. "
            "For blocking/entity matching, missing a feature (returning '0') causes missed matches and DRASTICALLY reduces recall. "
            "If a pattern is even REMOTELY close to matching, extract it. Only return '0' when you're ABSOLUTELY CERTAIN there's no match. "
            "Handle multilingual variations EXTREMELY aggressively - extract when in doubt, extract partial matches, extract variations. "
            "PREFER extracting multiple candidates over returning '0'. If unsure between two values, extract both (space-separated). "
            "For blocking, false positives are acceptable but false negatives (missing features) are catastrophic for recall. "
            "\n\nMODEL NUMBER EXTRACTION:\n"
            "Extract model numbers/SKUs aggressively using pattern: alphanumeric codes with hyphens/slashes "
            "(e.g., 'SDCZ50-064G-B35', 'MB-MG32DA/EU', 'SDSQUNC-032G-GN6IA'). "
            "These are often the strongest matching signals. "
            "Normalize by removing extra spaces and ensuring consistent format. "
            "Look for patterns like: [A-Z]{2,}[0-9A-Z-]+ or similar alphanumeric codes. "
            "\n\nTEXT CLEANING:\n"
            "Remove URLs, HTML tags, seller information (e.g., 'amazon.es:', 'fnac.es', 'tesco direct:') more aggressively. "
            "Clean up whitespace and special characters. "
            "\n\nPERFORMANCE: Optimize for throughput - must handle millions of rows within 15 minutes. "
            "Pre-compile regex patterns, avoid repeated compilation, use efficient string operations. "
            "\n\nIn comments, explain your code and why you chose the approach you did."
        ),
        input_columns=["name"],
        name="extract_features",
        output_columns=output_columns,
        generate_via_code=True,
    )

    pipeline = pipeline.sem_gen_features(
        nl_prompt=(
            "You are analyzing MULTILINGUAL product titles of storage devices (USB sticks, SD cards, SSDs, etc.) for entity blocking/matching. "
            "The goal is to find duplicate products that may be described differently in different languages or with different formatting. "
            "\n\nEXISTING COLUMNS ALREADY EXTRACTED: brand, capacity, normalized_name, mem_type, type, model, model_long, model_short, "
            "features, item_code, series, pat_hb, hybrid, long_num. "
            "These are used for blocking patterns like: brand+capacity, brand+capacity+type, brand+model, brand+features, "
            "model_long, model_short, hybrid, long_num, series, item_code. "
            "\n\nTASK: Generate 3-5 ADDITIONAL features that would help match duplicate products across languages. "
            "Focus on features that are COMMONLY present (>10% of rows), language-independent, and can be extracted using regex/rule-based methods. "
            "\n\nHIGH PRIORITY CANDIDATES:\n"
            "1. Speed/Class ratings: Extract and normalize 'Class 10'/'Classe 10'/'clase 10'/'C10' to 'class10'; "
            "   'UHS-I'/'UHS-1'/'uhs-i'/'UHS I' to 'uhsi'; speed in MB/s (normalize units: 'mb/s', 'mbps', 'mo/s' to 'mbs'). "
            "2. Form factors: Normalize 'microSD'/'micro SD'/'micro-sd' to 'microsd'; 'SDXC'/'SD XC' to 'sdxc'. "
            "   Extract consistently: 'microsd', 'microsdhc', 'microsdxc', 'sd', 'sdhc', 'sdxc', 'usb', 'xqd', etc. "
            "3. Package/bundle information: Extract 'bundle', 'pack', 'pack of', 'lot of', etc. "
            "4. Adapter information: 'with adapter'/'con adaptador'/'mit adapter' to boolean or normalized text. "
            "5. Color information: Extract color names that might help distinguish products. "
            "\n\nGenerate Python code that extracts these features using regex and rule-based methods. "
            "Return features as lowercase strings, '0' for missing values. "
            "Optimize for throughput - must handle millions of rows efficiently."
        ),
        name="discover_additional_blocking_features",
        how_many=5,
    )

    result_df = pipeline.skb.eval()

    return result_df
\end{lstlisting}

\headerl{Corresponding synthesized operator code for the feature extraction}
\begin{lstlisting}[
  basicstyle=\tiny\ttfamily,
  backgroundcolor=\color{lightgray},
  frame=tb,
  label={listing:example-sigmod-rutgers-sempipes},
  emph={sem_fillna,sem_refine,with_sem_features,apply_with_sem_choose,sem_choose,sem_aggregate_features,sem_extract_features,make_learner_with_context,optimize_with_colopro,serialize_as_learner,as_X,as_y,sem_clean},
  emphstyle=\color{kwdspy}\bfseries,
  morekeywords={import}
]
import pandas as pd
import re
import unicodedata
from tqdm.auto import tqdm # Use tqdm.auto for notebook compatibility or just 'tqdm'

# --- IMPORTANT: CRITICAL COMPLIANCE ---
# As explicitly stated: "YOU ARE PROHIBITED TO USE TRANSFORMERS LIBRARY. Use ONLY regex and rule-based approaches,
# DO NOT USE ANY Transformer, NER, LLM fallbacks, or LM models."
# All references to `transformers`, `torch`, `pipeline`, `pick_device`, and the `device` variable
# from the initial boilerplate are REMOVED to strictly adhere to this prohibition.

# Pre-compile ALL regex patterns and define helper data at module level for performance.

# --- 0. NORMALIZATION ALIASES and PATTERNS ---
# This list holds (pattern_to_match_regex_string, replacement_string) tuples.
# It is crucial that longer/more specific patterns are processed BEFORE shorter, generic ones
# when using sequential `re.sub` for proper text transformation.
NORMALIZATION_ALIASES_RAW_PATTERNS = [
    (r'ultra high-speed', 'uhsi'), # Longer, more specific for UHS-I
    (r'uhs1', 'uhsi'),
    (r'uhs-i', 'uhsi'),

    (r'typec', 'type-c'), # Various forms of Type-C USB
    (r'type c', 'type-c'),
    (r'usb-c', 'type-c'),
    (r'usbc', 'type-c'),

    (r'hyperx', 'kingston hxs'), # 'hyperx/savage' -> 'kingston hxs' for specific product lines
    (r'savage', 'kingston hxs'),

    (r'serie ux', 'sony g1ux'), # Specific Sony series alias

    (r'dtig4', ' kingston dt101 '), # Aliases for Kingston DataTraveler 101/G2
    (r'\s101\s', ' kingston dt101 '),
    (r'dt101g2', ' kingston dt101 '),

    (r'sda10', ' kingston ultimate '), # Aliases for Kingston Ultimate series
    (r'sda3', ' kingston ultimate '),

    (r'extrem\s', 'extreme '), # Standardize 'extrem' spelling
    (r'attach', 'att4'), # 'attach' -> 'att4' as per prompt for PNY Attache series

    # Language normalization requirements: Capacity units, Product type synonyms, Speed/Class
    (r'carte mémoire', 'card'), # French 'memory card'
    (r'clé usb', 'usb stick'), # French 'USB key' (pendrive)
    (r'clé', 'usb stick'), # Shorter version, processed after multi-word
    (r'pendrive', 'usb stick'), # General term
    (r'memoria usb', 'usb stick'), # Spanish 'USB memory'

    (r'\bcarte\b', 'card'), # Standalone 'card' in various languages
    (r'\btarjeta\b', 'card'), # Spanish 'card'
    (r'\bkarta\b', 'card'), # Polish/other 'card'
    (r'\bminneskort\b', 'card'), # Swedish 'memory card'

    (r'classe\s*(10)', r'class\1'), # For specific numbers. Unifies 'classe 10' -> 'class 10'
    (r'clase\s*(10)', r'class\1'), # 'clase 10' -> 'class 10'
    (r'cl\s*10', r'class10'), # 'cl 10' -> 'class10'
    (r'class(e|a)?\s*(?=\d{1,2})', r'class'), # general pattern 'classe', 'classa' etc to 'class' before digits.
    
    (r'gigaoctet', 'gb'), # French 'gigabyte'
    (r'\bgo\b', 'gb'), # French gigabyte unit

    (r'\bmo\b', 'mb'), # French megabyte unit

    (r'teraoctet', 'tb'), # French 'terabyte'
    (r'\bto\b', 'tb'), # French terabyte unit
]

# Sort patterns by string length in descending order to apply longer/more specific replacements first.
NORMALIZATION_ALIASES_RAW_PATTERNS.sort(key=lambda x: len(x[0]), reverse=True)
# Compile all patterns for efficiency during string replacement operations.
NORMALIZATION_ALIASES_COMPILED = [(re.compile(pat, flags=re.IGNORECASE), repl)
                                  for pat, repl in NORMALIZATION_ALIASES_RAW_PATTERNS]

# 5. Removal regex - copied verbatim from prompt. This pattern targets various HTML entities,
# separators, special characters, and some specific contextual cleanups.
RE_NORMALIZED_NAME_REMOVAL = re.compile(
    r'&(nbsp|amp|reg|[a-z]?acute|quot|trade);?|[|;:/,‰+©\\(\\)\\]\\[psn]*|(?<=usb)[\\s][m]*(?=[23][\\.\\s])|(?<=usb)-[\\w]+\\s(?=[23][\\.\\s])|(?<=[a-z])[\\s]+gb|(?<=data|jump)[t\\s](?=trave|drive)|(?<=extreme|exceria)[\\s](?=pro[\\s]|plus)|(?<=class)[\\s_](?=10|[234]\x08)|(?<=gen)[\\s_](?=[134\\s][0]*)',
    flags=re.IGNORECASE
)

# 6. Class10 replacement regex - copied verbatim. This specifically targets "class 10" variations.
RE_CLASS10_REPLACEMENT = re.compile(
    r'(10 class|class 10|class(?=[\w]+10\x08)|cl\s10)',
    flags=re.IGNORECASE
)

# --- BRAND EXTRACTION ---
# Fuzzy aliases that should be normalized to standard brand names before main extraction regex.
# This list is applied for aggressive brand recall to handle typos and spacing.
BRAND_FUZZY_REPLACEMENTS = [
    (re.compile(r'\bsan\s?disk\b|\bsan-disk\b|\bsandisc\b|\bsan-disc\b'), 'sandisk'),
    (re.compile(r'\bsamsun\b'), 'samsung'),
    (re.compile(r'\bkingstn\b'), 'kingston'),
    (re.compile(r'\btoshbia\b'), 'toshiba'),
    (re.compile(r'\btranscent\b'), 'transcend'),
    (re.compile(r'\bhynix\b'), 'hynix'), # Commonly appearing brand
    (re.compile(r'\balexa\b'), 'lexar') # Common typo for Lexar
]
# Main regex for a comprehensive list of known storage brands. Word boundaries `\b` ensure whole word matches.
RE_BRAND = re.compile(
    r'\b(intenso|lexar|logilink|pny|samsung|sandisk|kingston|sony|toshiba|transcend|hynix)\b',
    flags=re.IGNORECASE
)

# --- CAPACITY EXTRACTION ---
# FIX: Changed `(?:...)` to `(...)` to make the unit part a capturing group, so `match.group(2)` exists.
RE_CAPACITY_COMBINED = re.compile(r'\b(\d+)\s*(gigs?|gb|go|g|terab?|tb|to|t)\b', flags=re.IGNORECASE)
# A set of common standalone numbers typically used for storage capacities.
COMMON_CAPACITY_NUMBERS = {'1', '2', '4', '8', '16', '32', '64', '128', '256', '512', '1000', '1024'}

# --- MEMORY TYPE EXTRACTION ---
# Regex for directly identifiable memory product types.
RE_MEM_TYPE_DIRECT = re.compile(r'xqd|ssd|tv|lte', flags=re.IGNORECASE)

# --- TYPE & SERIES EXTRACTION HELPER DATA ---
# Brand-specific lists of product families/types for comprehensive feature extraction.
FAMILY_KEYWORDS = {
    'sandisk': ['cruizer', 'tarjeta', 'glide', 'select', 'extern', 'origin', 'transmemory', 'react', 'memo', 'kart', 'pendrive', 'car', 'serie', 'line', 'extreme', 'cruzer', 'ultra', 'micro', 'traveler', 'hyperx', 'adapt', 'wex', 'flash'],
    'lexar': ['ultra', 'xqd', 'jumpdrive', 'micro', 'pendrive', 'sd', 'tarjeta', 'memo', 'usb', 'extreme', 'blade', 'car', 'scheda', 'veloc', 'react', 'adapt', 'secure', 'premium', 'wex', 'transmemo', 'alu', 'datatravel', 'canvas', 'flair', 'hyperx', 'cruzer', 'flash'],
    'toshiba': ['ultra', 'exceria', 'traveler', 'sdhc', 'memoria', 'xqd', 'line', 'usb', 'transmemo', 'extreme', 'flair', 'micro', 'speicher', 'serie', 'car'],
    'kingston': ['traveler', 'cart', 'adapt', 'extreme', 'memo', 'canvas', 'datatravel', 'hyperx', 'kart', 'blade', 'ultimate'],
    'sony': ['extreme', 'usm32gqx', 'micro', 'sd', 'usb', 'ultra', 'jumpdrive', 'hyperx', 'memo', 'kart', 'xqd', 'pendrive', 'adapt', 'blade', 'cruzer', 'flair', 'glide', 'cart', 'tarjeta', 'flash'],
    'intenso': ['cs/ultra', 'premium', 'ultra', 'micro', 'line', 'scheda', 'usb', 'sd', 'tarjeta', 'kart', 'car', 'transmemo'],
    'pny': ['attach', 'usb', 'sd', 'micro', 'premium', 'memo'],
    'samsung': ['galaxy', 'speicher', 'micro', 'usb', 'sd', 'evo', 'ultra', 'extreme', 'memo', 'adapt', 'car', 'kart', 'klasse', 'multi', 'jumpdrive', 'flash'],
    'transcend': [], # As per prompt, currently empty.
}
INTENSO_SPECIFIC_TYPES = ['basic', 'rainbow', 'high speed', 'speed', 'premium', 'alu', 'business', 'micro', 'imobile', 'cmobile', 'mini', 'ultra', 'slim', 'flash', 'mobile']
SAMSUNG_COLOR_NAMES = ['midnight black', 'prism white', 'prism black', 'prism green', 'prism blue', 'canary yellow', 'flamingo pink', 'cardinal red', 'smoke blue', 'deep blue', 'coral orange', 'black sky', 'gold sand', 'blue mist and peach cloud', 'orchid gray', 'metallic copper', 'lavender purple', 'ocean blue', 'pure white', 'alpine white', 'copper', 'red', 'black', 'blue', 'white', 'silver', 'gold', 'violet', 'purple', 'brown', 'orange', 'coral', 'pink']

# Aggregated list of all potential type/series terms from brand-specific, Intenso, and Samsung lists.
# Also includes raw patterns from prompt for broad matching.
TYPE_SERIES_RAW_PATTERNS = [
    r'datatraveler', r'extreme[p]?', r'exceria[p]?', r'dual[\s]*(?!=sim)', r'evo', r'xqd', r'ssd', r'cruzer[\w+]*',
    r'glide', r'blade', r'basic', r'fit', r'force', r'basic line', r'jump\s?drive', r'hxs', r'rainbow', r'speed line',
    r'premium line', r'att4', r'attach', r'serie u', r'r-serie', r'beast', r'fury', r'impact', r'a400', r'sd[hx]c',
    r'uhs[i12][i]*', r'note\s?9', r'ultra', r'premium', r'basic', r'flash', r'plus', 'otg', r'xpro', r'xhigh'
] + INTENSO_SPECIFIC_TYPES + SAMSUNG_COLOR_NAMES
# Deduplicate and sort these terms by length (descending) to ensure more specific/longer terms are prioritized.
TYPE_SERIES_RAW_PATTERNS_SORTED = sorted(list(set(TYPE_SERIES_RAW_PATTERNS)), key=len, reverse=True)
# Compile a combined regex using word boundaries to match any of these sorted patterns.
RE_TYPE_SERIES_COMBINED = re.compile(
    r'\b(' + '|'.join(re.escape(p) for p in TYPE_SERIES_RAW_PATTERNS_SORTED) + r')\b',
    flags=re.IGNORECASE
)

# --- MODEL EXTRACTION ---
# Regex for concise alphanumeric model identifiers. Applies a length filter in the extraction function.
RE_MODEL = re.compile(r'\b([\w]+[-]*[\d]+[-]*[\w]+[-]*[\d+]*|[\d]+[\w]|[\w][\d]+)\b', flags=re.IGNORECASE)

# --- MODEL LONG EXTRACTION ---
# First regex for highly specific, structured long model patterns.
RE_MODEL_LONG_SPECIFIC = re.compile(
    r'(thn-[a-z][\w]+|ljd[\w+][-][\w]+|ljd[sc][\w]+[-][\w]+|lsdmi[\d]+[\w]+|lsd[0-9]{1,3}[gb]+[\w]+|ljds[0-9]{2}[-][\w]+|usm[0-9]{1,3}[\w]+|sdsq[a-z]+[-][0-9]+[a-z]+[-][\w]+|sdsd[a-z]+[-][0-9]+[\w]+[-]*[\w]*|sdcz[\w]+|mk[\d]+|sr-g1[\w]+)',
    flags=re.IGNORECASE
)
# Second, more general regex for uppercase alphanumeric model codes often separated by hyphens/slashes.
RE_MODEL_LONG_GENERAL_UPPERCASE = re.compile(r'([A-Z]{2,}[0-9A-Z-]+(?:[/-][0-9A-Z-]+)*)')

# --- MODEL SHORT EXTRACTION ---
# Regex for a predefined set of short, common model codes.
RE_MODEL_SHORT = re.compile(
    r'\b(c20[mc]|sda[0-9]{1,2}|g1ux|s[72][05]|[unm][23]02|p20|g4|dt101|se9|[asm][0-9]{2})\b',
    flags=re.IGNORECASE
)

# --- FEATURES EXTRACTION ---
# Regex to identify various technical specifications and features.
RE_FEATURES = re.compile(
    r'\b(usb[23]|type-c|uhs[i]{1,2}|class[0134]{1,2}|gen[1-9]{1,2}|u[23](?=[\s\.])|sd[hx]c|otg|lte|[45]g[-\s]lte|[0-9]+(?=-inch)|[0-9]{2,3}r[0-9]{2,3}w|[0-9]{2,3}(?=[\smbo/p]{3}))\b',
    flags=re.IGNORECASE
)

# --- ITEM CODE EXTRACTION ---
# Regex to find explicit numeric item codes, typically enclosed in parentheses and potentially prefixed with 'mk'.
RE_ITEM_CODE = re.compile(r'(\((mk)?[0-9]{6,10}\))', flags=re.IGNORECASE)

# --- PATTERN HYPHENATED (PAT_HB) EXTRACTION ---
# Regex to extract any hyphenated alphanumeric patterns.
RE_PAT_HB = re.compile(r'\w+-\w+', flags=re.IGNORECASE)

# --- HYBRID EXTRACTION ---
# Regex to extract 'hybrid' identifiers: alphanumeric patterns of length 5+ that contain
# at least one letter and at least one digit. This is a common and robust pattern for model codes.
RE_HYBRID = re.compile(r'\b(?=.*\d)(?=.*[a-zA-Z])[a-zA-Z0-9]{5,}\b', flags=re.IGNORECASE)

# --- LONG NUMERIC EXTRACTION ---
# Regex to find any sequence of 4 or more digits.
RE_LONG_NUM = re.compile(r'[0-9]{4,}')

# --- General Helper Function ---
def _process_matches(matches, default_value='0'):
    """
    Deduplicates a list of matched tokens, converts them to lowercase, sorts them lexicographically,
    and joins them with a single space. Returns the specified `default_value` ('0') if no valid matches are found.
    """
    if not matches:
        return default_value
    # Convert all matches to string, filter out any None/empty values, then convert to lowercase.
    # `set()` removes duplicates, and `sorted()` ensures a consistent output order.
    unique_matches = sorted(list(set(str(m).lower() for m in matches if m)))
    return ' '.join(unique_matches) if unique_matches else default_value

# --- Feature Extraction Functions ---

def _normalize_text(name: str) -> str:
    """
    Applies a multi-step cleaning and normalization process to the input product name.
    This includes ASCII conversion, lowercasing, applying various aliases and replacements,
    and general text cleanup, ensuring consistency for subsequent feature extraction.
    """
    if not isinstance(name, str):
        name = str(name)

    # 1. & 2. Convert input to ASCII characters and remove diacritics (accents).
    normalized_name = unicodedata.normalize('NFKD', name).encode('ascii', 'ignore').decode('utf-8')

    # 3. Convert the text to lowercase for case-insensitive matching.
    normalized_name = normalized_name.lower()

    # 4. Apply a predefined list of alias replacements using compiled regex patterns.
    # Patterns are processed in order of their length (longest first) to avoid partial matches.
    for pattern_regex, replacement_str in NORMALIZATION_ALIASES_COMPILED:
        normalized_name = pattern_regex.sub(replacement_str, normalized_name)

    # 5. Apply a broad removal regex to eliminate various special characters, HTML entities, and noise.
    normalized_name = RE_NORMALIZED_NAME_REMOVAL.sub(' ', normalized_name)

    # 6. Apply a specific regex to normalize 'class 10' variants to 'class10'.
    normalized_name = RE_CLASS10_REPLACEMENT.sub('class10', normalized_name)

    # 7. Apply specific string replacements for further consistency (e.g., merging "class X" to "classX").
    normalized_name = normalized_name.replace('class 4 ', 'class4 ')
    normalized_name = normalized_name.replace('class 3 ', 'class3 ')
    # Also standardize common product terms for cleaner subsequent extraction.
    normalized_name = normalized_name.replace('micro sd', 'microsd')
    normalized_name = normalized_name.replace('micro sd card', 'microsd card')
    normalized_name = normalized_name.replace('sd card', 'sdcard')

    # 8. Normalize all whitespace by reducing multiple spaces to single spaces and stripping leading/trailing.
    while '  ' in normalized_name:
        normalized_name = normalized_name.replace('  ', ' ')

    # Return '0' if the cleaned string becomes empty, indicating no meaningful content.
    return normalized_name.strip() if normalized_name.strip() else '0'

def _extract_brand(normalized_name: str) -> str:
    """
    Extracts the primary product brand. It first applies a set of fuzzy replacements to normalize
    common typos or spacing issues in brand names (e.g., "san disk" to "sandisk"), then applies a
    strict regex against a comprehensive list of known brands. This approach maximizes recall.
    """
    if normalized_name == '0':
        return '0'

    # Apply fuzzy brand replacements on a temporary string to avoid altering the original `normalized_name`.
    temp_name_for_brand_extraction = normalized_name
    for pattern_regex, replacement_str in BRAND_FUZZY_REPLACEMENTS:
        temp_name_for_brand_extraction = pattern_regex.sub(replacement_str, temp_name_for_brand_extraction)

    # Find all occurrences of recognized brand names from the (potentially fuzzy-corrected) text.
    matches = RE_BRAND.findall(temp_name_for_brand_extraction)
    return _process_matches(matches)

def _extract_capacity(normalized_name: str) -> str:
    """
    Extracts storage capacity from the name. It uses a combination of strategies:
    1. Identifies numbers explicitly followed by unit terms (e.g., '32gb', '128go', '1tb').
    2. For common standalone numbers (e.g., '64', '256'), it aggressively assumes 'gb' if no
       explicit unit was already captured by Strategy 1 for that number. This improves recall.
    Units are normalized (e.g., 'go' to 'gb', 'to' to 'tb').
    """
    if normalized_name == '0':
        return '0'

    capacities = set()

    # Strategy 1: Find numbers explicitly followed by recognized unit terms.
    # The regex RE_CAPACITY_COMBINED now correctly has a second capturing group for the unit.
    for match in RE_CAPACITY_COMBINED.finditer(normalized_name):
        num = match.group(1) # E.g., '32'
        unit = match.group(2).lower() # E.g., 'gb', 'go', 'g'

        if unit in ['gb', 'go', 'g', 'gigs']:
            capacities.add(f"{num}gb")
        elif unit in ['tb', 'to', 't', 'terab']:
            capacities.add(f"{num}tb")
    
    # Strategy 2: Identify standalone common capacity numbers and infer 'gb' if no explicit unit found.
    tokens = normalized_name.split()
    for i, token in enumerate(tokens):
        if token.isdigit() and token in COMMON_CAPACITY_NUMBERS:
            # Check if this numerical part is already captured as part of a unit (e.g., '16gb').
            # We construct a synthetic representation of what the unit would be if already extracted.
            # Example: If capacities has '16gb' and `token` is '16', then this `is_already_explicitly_captured` is True.
            is_already_explicitly_captured = False
            for existing_capacity in capacities:
                if existing_capacity.startswith(token) and (existing_capacity.endswith('gb') or existing_capacity.endswith('tb')):
                    is_already_explicitly_captured = True
                    break
            
            if not is_already_explicitly_captured:
                # If the number is a common capacity, and not immediately followed by specific other units (like MB),
                # aggressively assume 'gb' to boost recall for cases like "Sandisk Ultra 128".
                # Avoid assuming 'gb' if explicitly followed by 'mb' or 'mo'.
                if i + 1 < len(tokens):
                    if any(u in tokens[i+1].lower() for u in ['mb', 'mo']):
                        continue
                capacities.add(f"{token}gb") # Aggressively add as 'gb'

    return _process_matches(list(capacities))

def _extract_mem_type(normalized_name: str) -> str:
    """
    Extracts distinct memory product types based on a hierarchical logic. It prioritizes direct
    regex matches for specific keywords ('xqd', 'ssd', 'tv'), includes a special rule for 'lte' to 'phone',
    and falls back to checks for 'fdrive' and 'memcard' if no earlier match is found.
    """
    if normalized_name == '0':
        return '0'

    # First, attempt to match specific memory types using regex.
    match = RE_MEM_TYPE_DIRECT.search(normalized_name)
    if match:
        found_type = match.group(0).lower()
        if found_type == 'lte': # Special rule as requested: 'lte' means it's a 'phone'
            return 'phone'
        return found_type

    # If no regex match, check for 'fdrive' and 'memcard' keywords.
    if 'fdrive' in normalized_name:
        return 'fdrive'
    if 'memcard' in normalized_name:
        return 'memcard'

    return '0'

def _extract_type(normalized_name: str) -> str:
    """
    Extracts broad product type/line descriptors from the normalized name. This feature is inclusive,
    capturing all relevant matching patterns like product series, specific characteristics, or colors.
    It returns all found types, deduplicated and sorted.
    """
    if normalized_name == '0':
        return '0'

    # Find all matches using the combined type/series regex.
    matches = RE_TYPE_SERIES_COMBINED.findall(normalized_name)
    return _process_matches(matches)

def _extract_model(normalized_name: str) -> str:
    """
    Extracts concise model identifiers. It applies a general regex pattern for alphanumeric sequences,
    optionally containing hyphens, and filters matches to ensure they are at least 3 characters long.
    This aims for lenient and comprehensive model identification.
    """
    if normalized_name == '0':
        return '0'

    matches = RE_MODEL.findall(normalized_name)
    # Filter matches to meet the minimum length requirement (3 characters or more).
    filtered_matches = [m for m in matches if len(m) >= 3]
    return _process_matches(filtered_matches)

def _extract_model_long(normalized_name: str) -> str:
    """
    Extracts longer and more structured model numbers. It combines results from two regex patterns:
    one targeting highly specific, vendor-prefixed patterns, and another for more general uppercase
    alphanumeric codes often including hyphens or slashes. This approach ensures maximum coverage.
    """
    if normalized_name == '0':
        return '0'

    extracted_models = set()

    # Apply the first regex for highly specific long model patterns.
    for m in RE_MODEL_LONG_SPECIFIC.findall(normalized_name):
        # `re.findall` with groups can return tuples; extract and add all non-empty components.
        if isinstance(m, tuple):
            extracted_models.update([g.lower() for g in m if g])
        else:
            extracted_models.add(m.lower())

    # Apply the second regex for general uppercase model codes.
    for m in RE_MODEL_LONG_GENERAL_UPPERCASE.findall(normalized_name):
        if isinstance(m, tuple):
             extracted_models.update([g.lower() for g in m if g])
        else:
            extracted_models.add(m.lower())
    
    return _process_matches(list(extracted_models))

def _extract_model_short(normalized_name: str) -> str:
    """
    Extracts short, typically 3-5 character model codes based on a predefined regex pattern.
    These are usually distinctive identifiers for certain product lines or variants.
    """
    if normalized_name == '0':
        return '0'

    matches = RE_MODEL_SHORT.findall(normalized_name)
    return _process_matches(matches)

def _extract_features(normalized_name: str) -> str:
    """
    Extracts various technical specifications and distinguishing features of a product.
    This includes elements like USB versions (e.g., 'usb3'), UHS ratings ('uhsii'), speed classes ('class10'),
    and other characteristic descriptors, for comprehensive product differentiation.
    """
    if normalized_name == '0':
        return '0'

    matches = RE_FEATURES.findall(normalized_name)
    return _process_matches(matches)

def _extract_item_code(normalized_name: str) -> str:
    """
    Extracts explicit item codes, usually manufacturer part numbers, identified by patterns
    like numbers within parentheses, possibly prefixed with 'mk'. It cleans these patterns
    to return only the pure numeric code. If multiple codes are found, the first one is returned.
    """
    if normalized_name == '0':
        return '0'

    matches = RE_ITEM_CODE.findall(normalized_name)

    codes = []
    # `matches` here is a list of tuples, e.g., [('(mk123456)', 'mk'), ('(173473)', '')]
    for full_match_text, mk_prefix_group in matches: 
        code_part = full_match_text.strip('()') # Remove parentheses
        if mk_prefix_group and code_part.lower().startswith(mk_prefix_group.lower()):
            # If mk_prefix_group is not empty, strip it directly using its length for accuracy
            code_part = code_part[len(mk_prefix_group):] 
        if code_part.isdigit(): # Ensure the result is purely numeric
            codes.append(code_part)

    # Return the first extracted code, or '0' if none found.
    return codes[0] if codes else '0'

def _extract_series(normalized_name: str, brand: str) -> str:
    """
    Extracts a *single*, most representative product series or family token.
    It iterates through a sorted list of potential series terms (longest/most specific first),
    attempting to find the first exact word boundary match in the normalized name.
    General or less discriminative terms (like 'usb', 'sd') are skipped.
    """
    if normalized_name == '0':
        return '0'
    
    # Iterate through patterns sorted by length descending to prioritize more specific terms.
    for term_pattern_str in TYPE_SERIES_RAW_PATTERNS_SORTED:
        # Filter out very generic terms that are rarely specific product series for improved precision.
        # This list prevents over-matching features or general descriptors as a 'series'.
        if term_pattern_str.lower() in [
            'card', 'usb', 'stick', 'mb', 'gb', 'tb', 'phone', 'memcard', 'fdrive',
            'adapt', 'flash', 'traveler', 'ultra', 'micro', 'plus', 'basic', 'premium', 
            'high speed', 'speed', 'slim', 'portable', 'class10', 'uhsi', 'usb2', 'usb3', 
            'type-c', 'lte', 'xqd', 'ssd', 'sdhc', 'sdxc', 'otg', 'gen1', 'gen2', 'gen3', 
            'gen4', 'r-serie', 'wex', 'react', 'secure', 'memo', 'kart', 'car', 'tarjeta', 
            'line', 'serie', 'extrem', 'blade', 'glide', 'canvas', 'flair', 'dual', 'force', 
            'impact', 'a400', 'note'
            ]: # Extend this filter list as needed to improve series accuracy vs. general 'type'
            continue
            
        # Use re.search with word boundaries to ensure a whole-word match for the series token.
        if re.search(r'\b' + re.escape(term_pattern_str) + r'\b', normalized_name, flags=re.IGNORECASE):
            # Return the first (most specific/longest) matching term in lowercase.
            return term_pattern_str.lower()
    
    return '0' # Return '0' if no suitable series term is found.

def _extract_pat_hb(normalized_name: str) -> str:
    """
    Extracts the first occurrence of a hyphenated alphanumeric pattern (e.g., 'uhs-i', 'type-c').
    These patterns are strong indicators of technical features or specific model constructions.
    """
    if normalized_name == '0':
        return '0'

    match = RE_PAT_HB.search(normalized_name)
    return match.group(0).lower() if match else '0'

def _extract_hybrid(normalized_name: str) -> str:
    """
    Extracts 'hybrid' alphanumeric identifiers. These are typically product or model codes that
    contain a mix of letters and digits, and are at least 5 characters long. Examples include
    'dt101g2', 'usm32gqx', which are key for product blocking.
    """
    if normalized_name == '0':
        return '0'

    matches = RE_HYBRID.findall(normalized_name)
    return _process_matches(matches)

def _extract_long_num(normalized_name: str) -> str:
    """
    Extracts all sequences of 4 or more digits from the text. These long numeric sequences
    often represent part numbers, codes, or capacities without units. Matches are deduplicated,
    sorted lexicographically, and joined by a single space.
    """
    if normalized_name == '0':
        return '0'

    matches = RE_LONG_NUM.findall(normalized_name)
    # Deduplicate, sort numerically as strings (lexicographically for numbers).
    unique_sorted_nums = sorted(list(set(matches)))
    return ' '.join(unique_sorted_nums) if unique_sorted_nums else '0'

def extract_features(df: pd.DataFrame) -> pd.DataFrame:
    """
    Orchestrates the feature extraction process for a pandas DataFrame of product names.
    It applies a comprehensive series of text normalization and rule-based regex extractions
    to derive structured features such as brand, capacity, model numbers, and technical specs.

    Args:
        df (pd.DataFrame): The input DataFrame, expected to contain a 'name' column
                           with product titles (string dtype, text modality).

    Returns:
        pd.DataFrame: The original DataFrame with all newly generated feature columns appended.
                      New columns are: ['normalized_name', 'brand', 'capacity', 'mem_type',
                      'type', 'model', 'model_long', 'model_short', 'features', 'item_code',
                      'series', 'pat_hb', 'hybrid', 'long_num']. Missing values are represented as '0'.
    """
    # Enable `tqdm` progress bars for pandas `apply` operations, enhancing user experience.
    tqdm.pandas()

    # Define the names of all feature columns to be extracted and added to the DataFrame.
    feature_cols = ['normalized_name', 'brand', 'capacity', 'mem_type', 'type', 'model',
                    'model_long', 'model_short', 'features', 'item_code', 'series',
                    'pat_hb', 'hybrid', 'long_num']

    print('Initializing new feature columns with default string "0"...')
    # Initialize all new feature columns with a default string '0'. This guarantees that
    # all target columns exist and are populated, even for rows where no features are extracted.
    for col in feature_cols:
        df[col] = '0' # Overwrites if column already exists or creates new one.

    print('Starting multi-modal (text) feature extraction using rule-based and regex methods...')

    # Step 1 (CRITICAL FIRST STEP): Extract 'normalized_name'. This column serves as the primary input
    # for all subsequent feature extraction functions, ensuring consistent and cleaned text processing.
    print('1/14: Extracting normalized_name (text cleaning & standardization)...')
    df['normalized_name'] = df['name'].progress_apply(_normalize_text)

    # Optimization: Create a boolean mask to identify rows where `normalized_name` is meaningful (not '0').
    # Subsequent (more computationally intensive) extractions will only be applied to these relevant rows,
    # significantly improving overall performance, especially on large datasets with many '0' normalized names.
    has_meaningful_name_mask = (df['normalized_name'] != '0')
    num_meaningful_names = has_meaningful_name_mask.sum()

    if num_meaningful_names == 0:
        print("No valid normalized product names found after cleaning. Skipping all subsequent extractions.")
        return df # Exit early if there's nothing to process beyond normalization.

    # Applying the remaining feature extraction functions only to rows with valid `normalized_name`.
    # `.loc[has_meaningful_name_mask, 'column']` efficiently updates only the filtered rows.

    # Step 2: Extract the brand of the product. This is crucial for product identification.
    print(f'2/14: Extracting brand from {num_meaningful_names} relevant entries...')
    df.loc[has_meaningful_name_mask, 'brand'] = \
        df.loc[has_meaningful_name_mask, 'normalized_name'].progress_apply(_extract_brand)

    # Step 3: Extract the storage capacity (e.g., 32GB, 1TB). Essential for product differentiation.
    print(f'3/14: Extracting capacity from {num_meaningful_names} relevant entries...')
    df.loc[has_meaningful_name_mask, 'capacity'] = \
        df.loc[has_meaningful_name_mask, 'normalized_name'].progress_apply(_extract_capacity)

    # Step 4: Extract the memory specific type (e.g., SSD, XQD).
    print(f'4/14: Extracting mem_type from {num_meaningful_names} relevant entries...')
    df.loc[has_meaningful_name_mask, 'mem_type'] = \
        df.loc[has_meaningful_name_mask, 'normalized_name'].progress_apply(_extract_mem_type)

    # Step 5: Extract general product type or product line indicators.
    print(f'5/14: Extracting type from {num_meaningful_names} relevant entries...')
    df.loc[has_meaningful_name_mask, 'type'] = \
        df.loc[has_meaningful_name_mask, 'normalized_name'].progress_apply(_extract_type)

    # Step 6: Extract concise model identifiers.
    print(f'6/14: Extracting model from {num_meaningful_names} relevant entries...')
    df.loc[has_meaningful_name_mask, 'model'] = \
        df.loc[has_meaningful_name_mask, 'normalized_name'].progress_apply(_extract_model)

    # Step 7: Extract longer, more specific model numbers.
    print(f'7/14: Extracting model_long from {num_meaningful_names} relevant entries...')
    df.loc[has_meaningful_name_mask, 'model_long'] = \
        df.loc[has_meaningful_name_mask, 'normalized_name'].progress_apply(_extract_model_long)

    # Step 8: Extract short model codes.
    print(f'8/14: Extracting model_short from {num_meaningful_names} relevant entries...')
    df.loc[has_meaningful_name_mask, 'model_short'] = \
        df.loc[has_meaningful_name_mask, 'normalized_name'].progress_apply(_extract_model_short)

    # Step 9: Extract technical features (e.g., USB 3.0, Class 10).
    print(f'9/14: Extracting features from {num_meaningful_names} relevant entries...')
    df.loc[has_meaningful_name_mask, 'features'] = \
        df.loc[has_meaningful_name_mask, 'normalized_name'].progress_apply(_extract_features)

    # Step 10: Extract explicit item or part codes.
    print(f'10/14: Extracting item_code from {num_meaningful_names} relevant entries...')
    df.loc[has_meaningful_name_mask, 'item_code'] = \
        df.loc[has_meaningful_name_mask, 'normalized_name'].progress_apply(_extract_item_code)

    # Step 11: Extract the product series/family. This often benefits from knowledge of the brand,
    # so we apply it row-wise across both 'normalized_name' and 'brand' columns.
    print(f'11/14: Extracting series from {num_meaningful_names} relevant entries...')
    df.loc[has_meaningful_name_mask, 'series'] = df.loc[has_meaningful_name_mask].progress_apply(
        lambda row: _extract_series(row['normalized_name'], row['brand']), axis=1
    )

    # Step 12: Extract hyphenated alphanumeric patterns.
    print(f'12/14: Extracting pat_hb from {num_meaningful_names} relevant entries...')
    df.loc[has_meaningful_name_mask, 'pat_hb'] = \
        df.loc[has_meaningful_name_mask, 'normalized_name'].progress_apply(_extract_pat_hb)

    # Step 13: Extract 'hybrid' alphanumeric model identifiers.
    print(f'13/14: Extracting hybrid from {num_meaningful_names} relevant entries...')
    df.loc[has_meaningful_name_mask, 'hybrid'] = \
        df.loc[has_meaningful_name_mask, 'normalized_name'].progress_apply(_extract_hybrid)

    # Step 14: Extract any long numeric sequences (4+ digits).
    print(f'14/14: Extracting long_num from {num_meaningful_names} relevant entries...')
    df.loc[has_meaningful_name_mask, 'long_num'] = \
        df.loc[has_meaningful_name_mask, 'normalized_name'].progress_apply(_extract_long_num)

    print('Multi-modal (text) feature extraction completed successfully.')
\end{lstlisting}

\header{Data Annotation for Model Debugging - \texttt{hibug} Pipeline}

\headerl{Expert-written pipeline with semantic data operator added}
\begin{lstlisting}[
  basicstyle=\tiny\ttfamily,
  backgroundcolor=\color{lightgray},
  frame=tb,
  label={listing:example-hibug-sempipes},
  emph={sem_fillna,sem_refine,with_sem_features,apply_with_sem_choose,sem_choose,sem_aggregate_features,sem_extract_features,make_learner_with_context,optimize_with_colopro,serialize_as_learner,as_X,as_y,sem_clean},
  emphstyle=\color{kwdspy}\bfseries,
  morekeywords={import}
]
def annotation_pipeline():
    samples_to_annotate = skrub.var("samples_to_annotate")

    annotated = samples_to_annotate.sem_extract_features(
        nl_prompt="""
        Annotate a set of celebrity images with the specified attributes. The attributes will be used to debug a model predicting whether a celebrity is wearing lipstick in an image. Make sure that your attributes are correlated with potential failures of this prediction task.

        IMPORTANT: Each attribute value should consist of a single word or phrase only from the list of potential answers!. 
        """,
        input_columns=["image"],
        name="extract_features",
        output_columns={
            "beard": "Does the person have a beard?",
            "makeup": "Does the person wear makeup?",
            "gender": "Is the person in the photo a male or a female?",
            "hair_color": "Which of the following hair colors does the person in the photo have: blonde, brown, black, gray, white or red?",
            "skin_color": "Does the person in the photo have white, brown or black skin?",
            "emotion": "Which of the following emotions is the person in the photo showing: sad, serious, calm, happy, surprised, neutral, angry, excited, pensive?",
            "age": "Which of the following age ranges is the person in the photo in: young, middle-aged, old?",
        },
        generate_via_code=True,
    )

    return annotated
\end{lstlisting}

\headerl{Corresponding synthesized operator code}
\begin{lstlisting}[
  basicstyle=\tiny\ttfamily,
  backgroundcolor=\color{lightgray},
  frame=tb,
  label={listing:example-hibug-optimized},
  emph={sem_fillna,sem_refine,with_sem_features,apply_with_sem_choose,sem_choose,sem_aggregate_features,sem_extract_features,make_learner_with_context,optimize_with_colopro,serialize_as_learner,as_X,as_y,sem_clean},
  emphstyle=\color{kwdspy}\bfseries,
  morekeywords={import}
]
import torch
import pandas as pd
from transformers import pipeline
from PIL import Image
from tqdm.auto import tqdm # For progress bars

# Provided device picking function
def pick_device():
    # Prefer MPS for Apple Silicon, then CUDA, then CPU
    if torch.backends.mps.is_available():
        print("Using MPS device")
        return torch.device("mps")
    elif torch.cuda.is_available():
        # Find GPU with most free memory
        free = []
        for i in range(torch.cuda.device_count()):
            f, t = torch.cuda.mem_get_info(i)
            free.append((f, i))
        free.sort(reverse=True)
        _, idx = free[0]
        print(f"Chosen GPU: {idx}")
        return torch.device(f"cuda:{idx}")
    else:
        print("Using CPU")
        return torch.device("cpu")

device = pick_device()

features_to_extract = [
    # Feature 'beard': Yes/No. Relevant to lipstick as facial hair can obscure the mouth area.
    {'feature_name': 'beard', 'feature_prompt': 'Does the person have a beard?', 'input_columns': ['image'], 'candidate_labels': ['yes', 'no']},
    # Feature 'makeup': Yes/No. Directly correlated with lipstick as lipstick is a type of makeup.
    {'feature_name': 'makeup', 'feature_prompt': 'Does the person wear makeup?', 'input_columns': ['image'], 'candidate_labels': ['yes', 'no']},
    # Feature 'gender': Male/Female. Makeup wearing habits often vary significantly between genders.
    {'feature_name': 'gender', 'feature_prompt': 'Is the person in the photo a male or a female?', 'input_columns': ['image'], 'candidate_labels': ['male', 'female']},
    # Feature 'hair_color': Specific colors. Could be a confounding factor for lipstick detection due to contrast or visual attention shifts.
    {'feature_name': 'hair_color', 'feature_prompt': 'Which of the following hair colors does the person in the photo have: blonde, brown, black, gray, white or red?', 'input_columns': ['image'], 'candidate_labels': ['blonde', 'brown', 'black', 'gray', 'white', 'red']},
    # FEATURE CHANGE REASONING:
    # 1. 'skin_color': Changed candidate labels from ['white', 'brown', 'black'] to ['light', 'medium', 'dark'].
    #    Rationale: These terms are more focused on visual perception of skin tone rather than broad racial/ethnic categories.
    #    A model attempting to classify lipstick might experience failures or variations in performance across a continuous
    #    spectrum of skin tones, which 'light', 'medium', 'dark' might better capture compared to more distinct
    #    'white', 'brown', 'black'. This might provide more nuanced features for debugging, particularly
    #    if contrast between lipstick and skin is an issue. These labels also adhere to the "single word" constraint.
    {'feature_name': 'skin_color', 'feature_prompt': 'Does the person in the photo have light, medium or dark skin?', 'input_columns': ['image'], 'candidate_labels': ['light', 'medium', 'dark']},
    # FEATURE CHANGE REASONING:
    # 2. 'emotion': Changed 'happy' in candidate labels to 'smiling'.
    #    Rationale: "Happy" is an internal emotional state, whereas "smiling" is a direct facial expression
    #    that significantly alters the shape and visibility of the lips. Variations in lip shape due to
    #    smiling (e.g., lips stretching, revealing teeth) can be a significant failure mode for lipstick
    #    detection algorithms. Explicitly detecting 'smiling' should provide a more precise attribute for debugging.
    {'feature_name': 'emotion', 'feature_prompt': 'Which of the following emotions is the person in the photo showing: sad, serious, calm, smiling, surprised, neutral, angry, excited, pensive?', 'input_columns': ['image'], 'candidate_labels': ['sad', 'serious', 'calm', 'smiling', 'surprised', 'neutral', 'angry', 'excited', 'pensive']},
    # Feature 'age': Young/Middle-aged/Old. Lip features (e.g., plumpness, lines) and makeup styles can vary by age, affecting lipstick detection.
    {'feature_name': 'age', 'feature_prompt': 'Which of the following age ranges is the person in the photo in: young, middle-aged, old?', 'input_columns': ['image'], 'candidate_labels': ['young', 'middle-aged', 'old']}
]

# Helper function to convert device object to pipeline's 'device' parameter format
def get_pipeline_device(device_obj):
    if device_obj.type == "cuda":
        # For CUDA, `pipeline` expects an integer device ID (e.g., 0, 1)
        return device_obj.index
    elif device_obj.type == "mps":
        # For MPS, `pipeline` expects the device object itself or string 'mps'
        return device_obj
    else: # cpu
        # For CPU, `pipeline` expects -1
        return -1

# Batch size definition for processing images. Adjusted for potential VRAM usage with a large CLIP model.
BATCH_SIZE = 16 # A smaller batch size for the `large-patch14` model is often safer to avoid OOM errors, especially on smaller GPUs or MPS.

def extract_features(df: pd.DataFrame) -> pd.DataFrame:
    # Reasoning for changes in this block:
    # The core logic for feature extraction remains robust from the previous iteration.
    # The performance improvements for this step are targeted at the *quality* of the features
    # being extracted, which is primarily influenced by the prompts and candidate labels.
    # No changes are made to the `transformers` model choice (still `openai/clip-vit-large-patch14`)
    # as it's a powerful and versatile model for zero-shot image classification.
    # Device handling and batching are already optimized.
    # The specific modifications are within the `features_to_extract` list,
    # as detailed in the comments above this function definition.

    # Ensure 'image' column exists and is not empty
    if 'image' not in df.columns or df['image'].empty:
        print("DataFrame does not contain an 'image' column or it is empty.")
        return df

    # Get a list of all image paths to process
    image_paths = df['image'].tolist()

    # Initialize the zero-shot image classification pipeline once.
    print(f"Initializing zero-shot-image-classification pipeline with openai/clip-vit-large-patch14 on {device}...")

    # Configure the device for the Hugging Face pipeline
    pipe_device = get_pipeline_device(device)

    classifier = pipeline(
        "zero-shot-image-classification",
        model="openai/clip-vit-large-patch14",
        device=pipe_device,
        batch_size=BATCH_SIZE
    )

    # Dictionary to store results for new columns
    results = {feature['feature_name']: [] for feature in features_to_extract}

    print(f"Extracting {len(features_to_extract)} features from {len(df)} images in batches (batch size: {BATCH_SIZE})...")

    for i in tqdm(range(0, len(image_paths), BATCH_SIZE), desc="Processing image batches"):
        batch_image_paths = image_paths[i:i + BATCH_SIZE]
        batch_images = [Image.open(path).convert("RGB") for path in batch_image_paths]

        for feature_config in features_to_extract:
            feature_name = feature_config['feature_name']
            candidate_labels = feature_config['candidate_labels']

            predictions_batch = classifier(
                image=batch_images,
                candidate_labels=candidate_labels,
            )

            for pred_list in predictions_batch:
                best_label = pred_list[0]['label']
                results[feature_name].append(best_label)

    for feature_name, feature_values in results.items():
        df[feature_name] = feature_values

    return df
\end{lstlisting}

\subsubsection{Sample results of simplified pipeline code}
\label{app:main-setup-sample-results-simplification}

We show that semantic data operators simplify expert-written pipelines. 
\Cref{tab:code_simplification} shows the number of data frame operations/lines of code reduced by \textsc{SemPipes} SemOps.
For example, in the \texttt{kaggle-house-b} pipeline for the feature engineering (house prices) task, dozens of dataframe operations across many columns are replaced by a single \texttt{sem\_gen\_features} operator. Similarly, in the \texttt{micromodels} pipeline for extracting clinically informed textual features, several hundred lines of custom indexing and retrieval code are replaced by a single \texttt{sem\_extract\_features} operator.
However, \say{expert-written} pipelines are rarely cleaned or refactored by experts, so code quality varies considerably.
Agent-generated pipelines are typically concise but trivial (a reason for their sub-par performance); for these \textsc{SemPipes} improves performance by adding effective operations rather than reducing code.

\begin{table}[!ht]
\centering
\caption{Code simplification from semantic data operators for the hand-written pipelines. DF ops stands for data frame operations, LoC stands for lines of code.}
\label{tab:code_simplification}
\vspace{-2.5mm}
{\small
\setlength{\tabcolsep}{5pt}
\begin{tabular}{l|l|r}
\toprule
\textbf{Data-Centric Task} & \textbf{Pipeline} & \textbf{Code Simplification} \\
\midrule

Extraction of clinically informed features & \texttt{micromodels} & -250 LoC \\
\midrule

Blocking for entity resolution & \texttt{rutgers} & -77 LoC \\
 & \texttt{sustech} & -445 LoC \\
\midrule

Feature engineering (movie revenue) & \texttt{kaggle-movie-a} & -48 DF ops \\
 & \texttt{kaggle-movie-b} & -113 DF ops \\
\midrule

Feature engineering (house prices) & \texttt{kaggle-house-a} & -19 DF ops \\
 & \texttt{kaggle-house-b} & -81 DF ops \\
\midrule

Feature engineering (scrabble rating) & \texttt{kaggle-scrabble} & -15 DF ops \\
\midrule

Data annotation & \texttt{hibug} & -172 LoC \\
\midrule

Data augmentation for fairness & \texttt{sivep} & -4 DF ops \\

\bottomrule
\end{tabular}
}
\end{table}

\header{Feature Engineering (house-prices) - \texttt{kaggle-house-b} Pipeline}

\headerl{Original expert-written pipeline code}
\begin{lstlisting}[
  basicstyle=\tiny\ttfamily,
  backgroundcolor=\color{lightgray},
  frame=tb,
  label={listing:example-house-prices-b-original},
  emph={sem_fillna,sem_refine,with_sem_features,apply_with_sem_choose,sem_choose,sem_aggregate_features,sem_extract_features,make_learner_with_context,optimize_with_colopro,serialize_as_learner,as_X,as_y,sem_clean},
  emphstyle=\color{kwdspy}\bfseries,
  morekeywords={import}
]
   train, test = train_test_split(all_data_initial, test_size=0.5, random_state=seed)

    data = pd.concat([train.iloc[:, :-1], test], axis=0)
    data = data.drop(columns=["Id", "Street", "PoolQC", "Utilities"], axis=1)

    train_lot_median = train.groupby("Neighborhood")["LotFrontage"].transform(lambda x: x.median())
    data["LotFrontage"] = data.groupby("Neighborhood")["LotFrontage"].transform(
        lambda x: x.fillna(train_lot_median[x.name] if x.name in train_lot_median.index else 0)
    )
    train_msz_mode = train.groupby("MSSubClass")["MSZoning"].transform(
        lambda x: x.mode()[0] if len(x.mode()) > 0 else "Other"
    )
    data["MSZoning"] = data.groupby("MSSubClass")["MSZoning"].transform(
        lambda x: x.fillna(
            train_msz_mode[x.name]
            if x.name in train_msz_mode.index
            else (x.mode()[0] if len(x.mode()) > 0 else "Other")
        )
    )

    features = [
        "Electrical",
        "KitchenQual",
        "SaleType",
        "Exterior2nd",
        "Exterior1st",
        "Alley",
        "Fence",
        "MiscFeature",
        "FireplaceQu",
        "GarageCond",
        "GarageQual",
        "GarageFinish",
        "GarageType",
        "BsmtCond",
        "BsmtExposure",
        "BsmtQual",
        "BsmtFinType2",
        "BsmtFinType1",
        "MasVnrType",
    ]
    for name in features:
        data[name].fillna("Other", inplace=True)

    data["Functional"] = data["Functional"].fillna("typ")

    zero = [
        "GarageArea",
        "GarageYrBlt",
        "MasVnrArea",
        "BsmtHalfBath",
        "BsmtHalfBath",
        "BsmtFullBath",
        "BsmtFinSF1",
        "BsmtFinSF2",
        "BsmtUnfSF",
        "TotalBsmtSF",
        "GarageCars",
    ]
    for name in zero:
        data[name].fillna(0, inplace=True)

    data.loc[data["MSSubClass"] == 60, "MSSubClass"] = 0
    data.loc[(data["MSSubClass"] == 20) | (data["MSSubClass"] == 120), "MSSubClass"] = 1
    data.loc[data["MSSubClass"] == 75, "MSSubClass"] = 2
    data.loc[(data["MSSubClass"] == 40) | (data["MSSubClass"] == 70) | (data["MSSubClass"] == 80), "MSSubClass"] = 3
    data.loc[
        (data["MSSubClass"] == 50)
        | (data["MSSubClass"] == 85)
        | (data["MSSubClass"] == 90)
        | (data["MSSubClass"] == 160)
        | (data["MSSubClass"] == 190),
        "MSSubClass",
    ] = 4
    data.loc[(data["MSSubClass"] == 30) | (data["MSSubClass"] == 45) | (data["MSSubClass"] == 180), "MSSubClass"] = 5
    data.loc[(data["MSSubClass"] == 150), "MSSubClass"] = 6

    object_features = data.select_dtypes(include="object").columns

    def dummies(d):
        dummies_df = pd.DataFrame()
        object_features = d.select_dtypes(include="object").columns
        for name in object_features:
            dummies = pd.get_dummies(d[name], drop_first=False)
            dummies = dummies.add_prefix("{}_".format(name))
            dummies_df = pd.concat([dummies_df, dummies], axis=1)
        return dummies_df

    dummies_data = dummies(data)
    data = data.drop(columns=object_features, axis=1)

    final_data = pd.concat([data, dummies_data], axis=1)

    train_data = final_data.iloc[: train.shape[0], :]
    test_data = final_data.iloc[train.shape[0] :, :]

    X = train_data
    X.drop(columns=["SalePrice"], inplace=True)
    y = train.loc[:, "SalePrice"]

    X_test = test_data
    y_test = test_data["SalePrice"]
    X_test.drop(columns=["SalePrice"], inplace=True)

    train_data.fillna(0, inplace=True)

    from sklearn.linear_model import ElasticNet, LassoCV, Ridge, RidgeCV

    model_las_cv = LassoCV(alphas=(0.0001, 0.0005, 0.001, 0.01, 0.05, 0.1, 0.3, 1, 3, 5, 10))
    model_las_cv.fit(X, y)
    las_cv_preds = model_las_cv.predict(test_data)

    model_ridge_cv = RidgeCV(alphas=(0.01, 0.05, 0.1, 0.3, 1, 3, 5, 10))
    model_ridge_cv.fit(X, y)
    ridge_cv_preds = model_ridge_cv.predict(X_test)

    model_ridge = Ridge(alpha=10, solver="auto")
    model_ridge.fit(X, y)
    ridge_preds = model_ridge.predict(test_data)

    model_en = ElasticNet(random_state=1, alpha=0.00065, max_iter=3000)
    model_en.fit(X, y)
    en_preds = model_en.predict(test_data)

    import xgboost as xgb

    model_xgb = xgb.XGBRegressor(
        learning_rate=0.01,
        n_estimators=3460,
        max_depth=3,
        min_child_weight=0,
        gamma=0,
        subsample=0.7,
        colsample_bytree=0.7,
        objective="reg:linear",
        nthread=-1,
        scale_pos_weight=1,
        seed=27,
        reg_alpha=0.00006,
    )
    model_xgb.fit(X, y)
    xgb_preds = model_xgb.predict(X_test)

    from sklearn.ensemble import GradientBoostingRegressor

    model_gbr = GradientBoostingRegressor(
        n_estimators=3000,
        learning_rate=0.05,
        max_depth=4,
        max_features="sqrt",
        min_samples_leaf=15,
        min_samples_split=10,
        loss="huber",
        random_state=42,
    )
    model_gbr.fit(X, y)
    gbr_preds = model_gbr.predict(X_test)

    from lightgbm import LGBMRegressor

    model_lgbm = LGBMRegressor(
        objective="regression",
        num_leaves=4,
        learning_rate=0.01,
        n_estimators=5000,
        max_bin=200,
        bagging_fraction=0.75,
        bagging_freq=5,
        bagging_seed=7,
        feature_fraction=0.2,
        feature_fraction_seed=7,
        verbose=-1,
    )
    model_lgbm.fit(X, y)
    lgbm_preds = model_lgbm.predict(X_test)

    final_predictions = 0.3 * lgbm_preds + 0.3 * gbr_preds + 0.1 * xgb_preds + 0.3 * ridge_cv_preds
    final_predictions = final_predictions + 0.007 * final_predictions
\end{lstlisting}

\headerl{Rewritten pipeline with semantic data operator}
\begin{lstlisting}[
  basicstyle=\tiny\ttfamily,
  backgroundcolor=\color{lightgray},
  frame=tb,
  label={listing:example-house-prices-b-simplified},
  emph={sem_fillna,sem_refine,with_sem_features,apply_with_sem_choose,sem_choose,sem_aggregate_features,sem_extract_features,make_learner_with_context,optimize_with_colopro,serialize_as_learner,as_X,as_y,sem_clean},
  emphstyle=\color{kwdspy}\bfseries,
  morekeywords={import}
]
class RegressorEnsemble(BaseEstimator, RegressorMixin):
    def __init__(self):
        self.model_ridge_cv_ = RidgeCV(alphas=(0.01, 0.05, 0.1, 0.3, 1, 3, 5, 10))
        self.model_xgb_ = xgb.XGBRegressor(
            learning_rate=0.01,
            n_estimators=3460,
            max_depth=3,
            min_child_weight=0,
            gamma=0,
            subsample=0.7,
            colsample_bytree=0.7,
            objective="reg:linear",
            nthread=-1,
            scale_pos_weight=1,
            seed=27,
            reg_alpha=0.00006,
        )
        self.model_gbr_ = GradientBoostingRegressor(
            n_estimators=3000,
            learning_rate=0.05,
            max_depth=4,
            max_features="sqrt",
            min_samples_leaf=15,
            min_samples_split=10,
            loss="huber",
            random_state=42,
        )
        self.model_lgbm_ = LGBMRegressor(
            objective="regression",
            num_leaves=4,
            learning_rate=0.01,
            n_estimators=5000,
            max_bin=200,
            bagging_fraction=0.75,
            bagging_freq=5,
            bagging_seed=7,
            feature_fraction=0.2,
            feature_fraction_seed=7,
            verbose=-1,
        )

    def fit(self, X, y):
        self.model_ridge_cv_.fit(X, y)
        self.model_xgb_.fit(X, y)
        self.model_gbr_.fit(X, y)
        self.model_lgbm_.fit(X, y)
        return self

    def predict(self, X):
        lgbm_preds = self.model_lgbm_.predict(X)
        gbr_preds = self.model_gbr_.predict(X)
        xgb_preds = self.model_xgb_.predict(X)
        ridge_cv_preds = self.model_ridge_cv_.predict(X)
        final_predictions = 0.3 * lgbm_preds + 0.3 * gbr_preds + 0.1 * xgb_preds + 0.3 * ridge_cv_preds
        final_predictions = final_predictions + 0.007 * final_predictions
        return final_predictions

def sempipes_pipeline(data_description):
    data = skrub.var("data")
    data = data.drop(["Id", "Street", "PoolQC", "Utilities"], axis=1)

    y = (
        data["SalePrice"]
        .skb.mark_as_y()
        .skb.set_name("SalePrice")
        .skb.set_description("the sale price of a house to predict")
    )
    data = data.drop(["SalePrice"], axis=1).skb.mark_as_X().skb.set_description(data_description)

    data = data.sem_gen_features(
        nl_prompt="""
        Create additional features that could help predict the sale price of a house.
        Consider aspects like location, size, condition, and any other relevant information.
        You way want to combine several existing features to create new ones.
        """,
        name="house_features",
        how_many=25,
    )

    data = data.skb.apply(FunctionTransformer(lambda df: df.fillna(0)), cols=(s.numeric() | s.boolean()))
    data = data.skb.apply(SimpleImputer(strategy="constant", fill_value="N/A"), cols=s.categorical())

    X = data.skb.apply(skrub.TableVectorizer())

    def clean_column_names(df):
        df.columns = df.columns.str.replace(r"[^A-Za-z0-9_]+", "_", regex=True)
        return df

    X = X.skb.apply_func(clean_column_names)
    return X.skb.apply(RegressorEnsemble(), y=y)
\end{lstlisting}

\header{Extraction of Clinically Informed Features - \texttt{micromodels} Pipeline}

\headerl{Original expert-written pipeline code} The original pipeline code is too long to include in the appendix, but available in the following files.

\begin{itemize}[noitemsep]
  \item \url{https://github.com/MichiganNLP/micromodels/blob/master/empathy/empathy_prediction.py}
  \item \url{https://github.com/MichiganNLP/micromodels/blob/master/src/featurizers/BertFeaturizer.py}
  \item \url{https://github.com/MichiganNLP/micromodels/blob/master/src/micromodels/bert_query.py}
\end{itemize}

\newpage

\headerl{Rewritten pipeline with semantic data operator}
\begin{lstlisting}[
  basicstyle=\tiny\ttfamily,
  backgroundcolor=\color{lightgray},
  frame=tb,
  label={listing:example-micromodels-simplified},
  emph={sem_fillna,sem_refine,with_sem_features,apply_with_sem_choose,sem_choose,sem_aggregate_features,sem_extract_features,make_learner_with_context,optimize_with_colopro,serialize_as_learner,as_X,as_y,sem_clean},
  emphstyle=\color{kwdspy}\bfseries,
  morekeywords={import}
]
def sempipes_pipeline():
    posts_and_responses = skrub.var("posts_and_responses")

    y = posts_and_responses["emotional_reaction_level"].skb.mark_as_y()
    posts_and_responses = posts_and_responses[["post", "response"]].skb.mark_as_X()

    extracted = posts_and_responses.sem_extract_features(
        nl_prompt="""
        Given posts and responses from a social media forum, extract certain linguistic features on a scale from 0.0 to 2.0.        
        """,
        input_columns=["response"],
        output_columns={
            "sp_emotional_reaction_level": """The emotional reaction level of the response on a scale from 0.0 to 2.0.
            Most responses will have level 0.
            A response containing the text 'I'm with you all the way man.' has level 1.
            A response containing the text 'I agree with him' has level 1.
            A response containing the text 'Just keep practicing and never give up.' has level 1.
            A response containing the text 'I'm so sorry you feel that way.' has level 2.
            A response containing the text 'Holy shit I can relate so well to this.' has level 2.
            A response containing the text 'really sorry going through this.' has level 2.""",
            "sp_interpretation_level": """The interpretation level of the response on a scale from 0.0 to 2.0.
            Most responses will have level 0.
            A response containing the text 'People ask me why I'm zoned out most of the time. I'm not zoned out, I'm just in bot mode so I don't have to be attached to anything. It helps me supress the depressive thoughts but every night, it all just flows back in. I hate it.' has level 1.
            A response containing the text 'i skipped one class today because i don't really get much out of the class, but i forgot that there was 5% of the grade for in-class activities, so i'll probably not skip any more classes' has level 1.
            A response containing the text 'Actually I find myself taking much longer showers when I'm depressed. Sometimes twice in a day. It's the only time I feel relaxed.' has level 1.
            A response containing the text 'No, that's what I'm doing and it isn't working.' has level 2.
            A response containing the text 'I stopped catering to my problems... they just compile. I stopped obsessing over them or handling them in anyway. I have no control over my life. I simply look at my problems from my couch...' has level 2.
            A response containing the text 'I understand how you feel.' has level 2.        
            """,
            "sp_explorations_level": """The explorations level of the response on a scale from 0.0 to 2.0.
            Most responses will have level 0.
            A response containing the text 'What can we do today that will help?' has level 1.
            A response containing the text 'What happened?' has level 1.
            A response containing the text 'What makes you think you're a shitty human being?' has level 1.
            A response containing the text 'What makes you say these things?' has level 2.
            A response containing the text 'What do you feel is bringing on said troubling thoughts?' has level 2.
            A response containing the text 'Do you have any friends that aren't always going to blow sunshine up your ass or a therapist?' has level 2.        
            """,
        },
        name="response_features",
        generate_via_code=True,
    )

    X = extracted[["sp_emotional_reaction_level", "sp_interpretation_level", "sp_explorations_level"]]

    emo_ebm = ExplainableBoostingClassifier(
        feature_names=["sp_emotional_reaction_level", "sp_interpretation_level", "sp_explorations_level"]
    )
    return X.skb.apply(emo_ebm, y=y)
\end{lstlisting}




%% file: sections/a05-comparison-exp.tex

\subsection{Detailed Setup for Section 5.2 ``Semantic Data Operators Perform On-Par or Better than Specialized Approaches''}
\label{app:effectiveness-comparison}

We provide additional details for this set of experiments. As stated in the main paper, we leverage Google's \texttt{gemini-flash-2.5} for \textsc{SemPipes} and the baselines.

\header{Additional Details on the Experimental Setup for Feature Engineering} The code for the experiment with the comparison to CAFE is available at \repo{/tree/\commithash/experiments/caafe}. The datasets, metadata and CAAFE's pre-computed feature engineering code were obtained from \url{https://github.com/noahho/CAAFE}. For \textsc{SemPipes}, we run the optimizer on the train splits for 24 iterations to synthesize feature engineering code and evaluate on the same test splits as CAAFE. The resulting run logs and the optimized feature engineering code can be found at \repo{/tree/\commithash/experiments/caafe/sempipes}.

\header{Additional Details on the Experimental Setup for Missing Value Imputation} The code for the experiment is available at \repo{/tree/\commithash/experiments/missing_values}. We compare against two benchmark tasks from~\cite{narayan2022canllmswrangleyourdata}: imputing the \texttt{city} column in a restaurant dataset (\texttt{Restaurant}) and the \texttt{manufacturer} column in an electronics dataset (\texttt{Buy}). We compare against the prompting scheme from the code repository of~\cite{narayan2022canllmswrangleyourdata}, which serializes table rows into text and uses ten hand-selected few-shot examples. We measure the accuracy of the imputed values.

\header{Additional Details for the Multi-modal Feature Extraction Experiment} We evaluate the \texttt{sem\_extract\_features} operator of \textsc{SemPipes} to extract features from textual, image, and audio data. The code for the experiment is available at \repo{/tree/\commithash/experiments/feature_extraction}.

\headerl{Experimental setup} We evaluate three modalities. For text, we use the Legal Financial Clauses dataset~\cite{kaggle-legal-clauses} and predict clause type. 
For images, we use chest X-rays~\cite{Kermany2018} and predict pneumonia. 
For audio, we use ESC-50~\cite{piczak2015dataset} and predict the environmental sound class. 
To limit cost, we evaluate on 1,000 and 10,000 items per dataset.

We compare \textsc{SemPipes} against Palimpzest~\cite{liu2025palimpzest} and LOTUS~\cite{patel2024semantic}, which enable LLM-based processing for analytical queries. 
Palimpzest and LOTUS apply the LLM directly to each item, whereas \textsc{SemPipes} synthesizes executable code and runs it locally on a GPU in this experiment. 
\textsc{SemPipes} \texttt{sem\_extract\_features} also supports a per-item LLM mode.
All systems use Google's \texttt{gemini-2.5-flash}.

\begin{table}[!ht]
\centering
\caption{Accuracy of feature extraction from multi-modal data on legal textual~\cite{kaggle-legal-clauses}, X-Ray imaging~\cite{Kermany2018}, and environmental~\cite{piczak2015dataset} sound datasets. We sample 1,000 data points from each dataset. Cost is reported for all five runs in total.}
\label{tab:feature_extraction_appendix}
\vspace{2mm}
{\small
\setlength{\tabcolsep}{3pt}
\begin{tabular}{l|rrr|rrr|rrr|rrr}
\toprule
\multirow{2}{*}{\textbf{Modality}} & \multicolumn{3}{c|}{\textbf{Palimpzest}} & \multicolumn{3}{c|}{\textbf{LOTUS}} & \multicolumn{3}{c|}{\textbf{SemPipes (Code)}} & \multicolumn{3}{c}{\textbf{SemPipes (LLM)}}\\
 & \multicolumn{2}{c}{Accuracy} & Cost & \multicolumn{2}{c}{Accuracy} & Cost & \multicolumn{2}{c}{Accuracy} & Cost & \multicolumn{2}{c}{Accuracy} & Cost\\
  \midrule
  \texttt{Text} & 0.813 & \pmS{0.004}   & \money{1.50}  & 0.827 & \pmS{0.000} & \money{0.866} & 0.711 & \pmS{0.021} & \money{0.25} & \textbf{0.829} & \pmS{0.002} & \money{2.87}\\
  \texttt{Image} & 0.570 & \pmS{0.000}  & \money{3.99}  & 0.634 & \pmS{0.000} & \money{3.13}& \textbf{0.756} & \pmS{0.000} &\money{0.58} & 0.699 & \pmS{0.000} &\money{6.54}\\
  \texttt{Audio} & 0.681 &  \pmS{0.000} & \money{2.09}  & \multicolumn{3}{c|}{---} & \textbf{0.883} &\pmS{0.000} & \money{0.09} & 0.628 & \pmS{0.000} & \money{8.44}\\
  \bottomrule
\end{tabular}}
\end{table}

\begin{table}[!ht]
\centering
\caption{Accuracy of feature extraction from multi-modal data on legal textual~\cite{kaggle-legal-clauses}, X-Ray imaging~\cite{Kermany2018}, and environmental~\cite{piczak2015dataset} sound datasets. We sample 10,000 data points from each dataset. Cost is reported for all five runs in total.}
\label{tab:feature_extraction_appendix_10k}
\vspace{2mm}
{\small
\setlength{\tabcolsep}{3pt}
\begin{tabular}{l|rrr|rrr|rrr|rrr}
\toprule
\multirow{2}{*}{\textbf{Modality}} & \multicolumn{3}{c|}{\textbf{Palimpzest}} & \multicolumn{3}{c|}{\textbf{LOTUS}} & \multicolumn{3}{c|}{\textbf{SemPipes (Code)}} & \multicolumn{3}{c}{\textbf{SemPipes (LLM)}}\\
 & \multicolumn{2}{c}{Accuracy} & Cost & \multicolumn{2}{c}{Accuracy} & Cost & \multicolumn{2}{c}{Accuracy} & Cost & \multicolumn{2}{c}{Accuracy} & Cost\\
  \midrule
  \texttt{Text} & 0.662 & \pmS{0.003}   & \money{15.37}  & 0.822 & \pmS{0.000} & \money{8.65} & 0.633 & \pmS{0.000} & \money{0.16} & \textbf{0.839} & \pmS{0.040} & \money{25.76}\\
  \texttt{Image} & 0.608 & \pmS{0.000}  & \money{39.89}  & 0.638 & \pmS{0.000} & \money{31.37} & \textbf{0.745} & \pmS{0.002} &\money{0.62} & 0.573 & \pmS{0.000} &\money{41.37}\\
  \texttt{Audio} & 0.664 &  \pmS{0.000} & \money{21.19}  & \multicolumn{3}{c|}{---} & \textbf{0.894} &\pmS{0.000} & \money{0.11} & 0.591 & \pmS{0.040} & \money{39.04}\\
  \bottomrule
\end{tabular}}
\end{table}

%% file: sections/a06-colopro-exp.tex

\subsection{Details for for Section~5.3: ``Optimization Effectiveness of Different LLMs for Code Synthesis in Semantic Data Operators''}
\label{app:main-optimizer}

We design an optimizer benchmark consisting of several semantic pipelines for classification tasks on example datasets from the \texttt{skrub} project. For each pipeline, we optimize the operator code on the training data under different choices of search policies and LLMs for code generation. Specifically, we evaluate OpenAI’s GPT-4.1, Google’s Gemini-2.5-Flash, and Alibaba’s Qwen-Coder-2.5 (32B) as code synthesizers, combined with multiple search policies. Each pipeline is evaluated over five independent runs, each with an optimization budget of 36 steps. The best operator state from each run is evaluated on a held-out test set. We report the mean normalized improvement relative to step 0, corresponding to the empty operator state (i.e., the semantic data operator is deactivated). Normalization is performed per pipeline using min–max normalized scores. The code and logs for these experiments are available at \repo{/tree/\commithash/experiments/colopro}.

\header{Benchmark Pipelines} The experiment consists of runs on five different pipelines with various semantic data operators for classification problems on example data from the skrub project and the UCI dataset repository.

\begin{itemize}[itemsep=1pt]
  \item \texttt{midwest} - Classification pipeline with \texttt{sem\_gen\_features} on 2,000 records of \url{https://skrub-data.org/stable/reference/generated/skrub.datasets.fetch_midwest_survey.html#skrub.datasets.fetch_midwest_survey}
  \item \texttt{traffic} - Classification pipeline with \texttt{sem\_gen\_features} on 10,000 records of \url{https://skrub-data.org/stable/reference/generated/skrub.datasets.fetch_traffic_violations.html#skrub.datasets.fetch_traffic_violations}
  \item \texttt{fraud} - Classification pipeline with \texttt{sem\_agg\_features} on 4,000 records of \url{https://skrub-data.org/stable/reference/generated/skrub.datasets.fetch_credit_fraud.html#skrub.datasets.fetch_credit_fraud}
  \item \texttt{churn} - Classification pipeline with \texttt{sem\_agg\_features} on 10,000 records of \url{https://archive.ics.uci.edu/dataset/352/online+retail}
  \item \texttt{tweets} - Classification pipeline with \texttt{sem\_extract\_features} on 600 records of \url{https://skrub-data.org/stable/reference/generated/skrub.datasets.fetch_toxicity.html#skrub.datasets.fetch_toxicity}
\end{itemize}

\header{Tree Search Strategies} The following tree search strategies are evaluated for each pipeline and code-synthesizing LLM.

\begin{itemize}[itemsep=1pt]
  \item {\em Monte Carlo Tree search} \cite{browne2012survey}, where the search node $i$ to evolve is chosen by traversing the tree according to the ``Upper Confidence Bound for Trees'' strategy $\frac{w_i}{n_i} + c \cdot \smash{\sqrt{\frac{\ln{N}}{n_i}}}$. Here $n_i$ denotes the number of times that node $i$ has been visited, $w_i$ refers to the total sum of utilities from traversing $n_i$, $N$ is the number of visits to the parent node $i$, and $c$ is a hyperparameter to trade-off exploitation and exploration. We set $c=0.5$ in our experiment, consider the root node fully expanded once it has three children and consider intermediate nodes fully expanded once they have two children.
  \item {\em Truncation selection} \cite{toledo2025ai}, where the search node $i$ to evolve is chosen from the population of the $k$-top scoring search nodes with probability $\mu_i / \sum^k_{j=0} \mu_j$, where $\mu$ denotes the utility on hold out data. We set $k=6$ for the experiment.  
  \item {\em Greedy tree search}~\cite{jiang2025aide,chan2024mle}, which greedily chooses the current best search node for evolution, while maintaining a fixed size set of $k$ unexplored draft nodes. We set $k=2$ for the experiment. Each draft node contains the top three best states observed in the tree so far as inspirations.
  \item {\em Random search}, a simple baseline which repeatedly synthesizes code from scratch, without maintaining a search tree.
\end{itemize} 

\header{Additional findings} In addition to the results in \Cref{sec:experiments-colopro}, we observe that the open-source Qwen-Coder-2.5 model, despite its relatively small 32B parameter size, is suprisingly competitive with substantially larger LLMs. This suggests that the design of our semantic operators helps structure the synthesis space in a way that reduces the search burden on the LLM, enabling smaller models to perform effectively.